\pdfoutput=1

\documentclass[11pt,twoside]{article}

\usepackage{amsmath}
\usepackage{amsthm}
\usepackage{amssymb}
\usepackage{amsfonts}
\usepackage[top = 1 in, bottom = 1.2 in, left = 1 in, right = 1 in]{geometry}
\usepackage{color}
\usepackage{mathrsfs}
\usepackage[normalem]{ulem}
\usepackage{longtable}
\usepackage{hyperref}
\usepackage{cleveref}
\usepackage{dsfont}
\usepackage{graphicx}
\usepackage{caption}
\usepackage{subcaption}
\usepackage{float}
\usepackage{mathtools}
\usepackage{etoolbox}
\usepackage{thmtools}
\usepackage{enumitem}
\usepackage[textsize=tiny]{todonotes}
\usepackage{multirow}

\usepackage{esvect}

\usepackage[utf8]{inputenc}

\newtheorem{theorem}{Theorem}
\newtheorem{lemma}{Lemma}

\newtheorem{proposition}{Proposition}
\newtheorem{corollary}{Corollary}

\declaretheoremstyle[headfont=\bf,bodyfont=\normalfont]{ex}
\declaretheorem[style=ex]{example}
\declaretheoremstyle[bodyfont=\normalfont]{rm}

\DeclareMathOperator*{\argmax}{arg\,max}

\newcommand{\normmm}{{\vert\kern-0.25ex\vert\kern-0.25ex\vert}}
\newcommand{\bignormmm}{{\big\vert\kern-0.25ex\big\vert\kern-0.25ex\big\vert}}
\newcommand{\Bignormmm}{{\Big\vert\kern-0.25ex\Big\vert\kern-0.25ex\Big\vert}}

\newcommand{\defn}{\ensuremath{:\,=}}
\newcommand{\proj}{\ensuremath{\Pi}}

\newcommand{\yd}[1]{#1}

\makeatletter
\long\def\@makecaption#1#2{
\vskip 0.8ex
\setbox\@tempboxa\hbox{\small {\bf #1:} #2}
\parindent 1.5em  
\dimen0=\hsize
\advance\dimen0 by -3em
\ifdim \wd\@tempboxa >\dimen0
\hbox to \hsize{
\parindent 0em
\hfil 
\parbox{\dimen0}{\def\baselinestretch{0.96}\small
    {\bf #1.} {#2}
  } 
\hfil}
\else \hbox to \hsize{\hfil \box\@tempboxa \hfil}
\fi
}
\makeatother

\usepackage{hyperref}

\newcommand{\errprob}{\ensuremath{\delta}}

\long\def\comment#1{}

\newcommand{\CurveNorm}[2]{\ensuremath{\|#1\|}}

\newcommand{\yaqidone}{}
\newcommand{\ydsout}[1]{}

\usepackage{stackrel}

\newcommand{\Prob}{\ensuremath{\mathds{P}}}
\newcommand{\Exp}{\ensuremath{\mathds{E}}}

\newcommand{\Var}{\ensuremath{{\rm Var}}}
\newcommand{\Real}{\ensuremath{\mathds{R}}}
\newcommand{\real}{\Real}

\newcommand{\plaincon}{\ensuremath{c}}
\newcommand{\PlainCon}{\ensuremath{c}}
\newcommand{\const}[1]{\ensuremath{\plaincon_{#1}}}
\newcommand{\Const}[1]{\ensuremath{\PlainCon_{#1}}}

\newcommand{\MyCurve}[1]{ \ensuremath{C_{#1}}}
\newcommand{\MyCurveTwo}[2]{\MyCurve{#1}(#2)}
\newcommand{\MyCurveTwosq}[2]{\MyCurve{#1}^2(#2)}

\newcommand{\Curvstarfun}[2]{\ensuremath{\boldsymbol{\kappa}_{#1, #2}(\Bell^\star)}}
\newcommand{\Curvstar}[1]{\ensuremath{\kappa^{\star}_{#1}}}

\newcommand{\occupstar}{\ensuremath{\policy^\star}}

\newcommand{\Curvhtoh}[2]{\ensuremath{\boldsymbol{\kappa}_{#1,#2}(\policystar)}}

\newcommand{\Dim}{\ensuremath{d}}

\newcommand{\usedim}{\ensuremath{d}}

\newcommand{\DiffRKHS}{\ensuremath{\partial \RKHS}}

\newcommand{\Value}{\ensuremath{V}}
\newcommand{\Valueh}[1]{\ensuremath{\Value_{#1}}}
\newcommand{\Valuepolicyh}[2]{\ensuremath{\Valueh{#1}^{#2}}}

\newcommand{\thetastar}{\ensuremath{\theta^*}}

\newcommand{\Deltafun}{\ensuremath{\boldsymbol{\Delta}}}
\newcommand{\Deltafunh}[1]{\ensuremath{\Delta_{#1}}}

\newcommand{\ssum}[2]{\ensuremath{{\textstyle \sum_{#1}^{#2}}\,}}
\newcommand{\myssum}[2]{\ensuremath{{\sum_{#1}^{#2}}\,}}

\newcommand{\DiffOp}{\ensuremath{\beta}}
\newcommand{\DiffOphat}{\ensuremath{\widehat{\DiffOp}}}

\newcommand{\RKHS}{\ensuremath{\mathscr{F}}}

\newcommand{\reward}{\ensuremath{r}}

\newcommand{\rewardh}[1]{\ensuremath{r_{#1}}}

\newcommand{\discount}{\ensuremath{\gamma}}

\newcommand{\TransOp}{\ensuremath{\boldsymbol{\mathcal{P}}}}

\newcommand{\TransOph}[1]{\ensuremath{\mathcal{P}_{#1}}}
\newcommand{\TransOppi}[1]{\ensuremath{\TransOp^{#1}}}
\newcommand{\TransOppih}[2]{\ensuremath{\mathcal{P}_{#1}^{#2}}}

\newcommand{\StateSp}{\ensuremath{\mathcal{S}}}

\newcommand{\StateSpgoodh}[1]{\ensuremath{}}

\newcommand{\StateSpbadh}[1]{\ensuremath{}}
\newcommand{\Bell}{\ensuremath{\mathcal{T}}}
\newcommand{\BellOp}[1]{\ensuremath{\boldsymbol{\Bell}^{#1}}}
\newcommand{\BellOph}[2]{\ensuremath{\Bell_{#1}^{#2}}}

\newcommand{\BellOpstar}{\ensuremath{\BellOp{\star}}}

\newcommand{\BellOpstarh}[1]{\ensuremath{\BellOph{#1}{\star}\,}}

\newcommand{\weight}[1]{\ensuremath{w_{#1}}}
\newcommand{\bweight}{\ensuremath{{\boldsymbol{w}}}}
\newcommand{\bweighttil}{\ensuremath{\widetilde{\bweight}}}
\newcommand{\bweighthat}{\ensuremath{\widehat{\bweight}}}
\newcommand{\bweightstar}{\ensuremath{\bweight^{\star}}}

\newcommand{\Deltaweight}{\ensuremath{\Delta\bweight}}

\newcommand{\CovOp}[1]{\ensuremath{\boldsymbol{\Sigma}_{#1}}}

\newcommand{\CovOpdata}[1]{\ensuremath{\widehat{\boldsymbol{\Sigma}}_{#1, \Data}}}

\newcommand{\by}{\ensuremath{\boldsymbol{y}}}

\newcommand{\IdMt}{\ensuremath{\boldsymbol{I}}}

\newcommand{\Data}{\ensuremath{\mathcal{D}}}

\newcommand{\widgraph}[2]{\includegraphics[keepaspectratio,width=#1]{#2}}

\newcommand{\nfed}{\ensuremath{= \, :}}

\newcommand{\numobs}{\ensuremath{n}}
\newcommand{\Tsafe}{\ensuremath{T_0}}

\newcommand{\plaindelta}{\ensuremath{\nu}}

\newcommand{\delcrith}[1]{\ensuremath{\plaindelta_{#1}}}

\newcommand{\delcrittildeh}[1]{\ensuremath{\widetilde{\plaindelta}_{#1}}}

\newcommand{\eighat}{\ensuremath{\widehat{\mu}}}

\usepackage{upgreek} \newcommand{\distr}{\ensuremath{\upmu}}

\newcommand{\ridge}{\ensuremath{\lambda_\numobs}}
\newcommand{\ridgeh}[1]{\ensuremath{\lambda_{#1}}}
\newcommand{\distrdata}{\ensuremath{\bar{\distr}}}
\newcommand{\distrdatah}[1]{\ensuremath{\bar{\distr}_{#1}}}

\newcommand{\sdistrdata}{\ensuremath{\sdistr_{\Data}}}

\newcommand{\sdistr}{\ensuremath{\xi}}

\newcommand{\sdistrinit}{\ensuremath{\sdistr_1}}

\newcommand{\distrstarh}[1]{\ensuremath{\distr^{\star}_{#1}}}

\newcommand{\Fclass}{\ensuremath{\mathscr{F}}}

\newcommand{\newradhath}[1]{\ensuremath{\widehat{R}_{#1}}}

\newcommand{\Term}{\ensuremath{T}}

\newcommand{\state}{\ensuremath{s}}

\newcommand{\stateh}[1]{\ensuremath{\state_{#1}}}

\newcommand{\Stateh}[1]{\ensuremath{\State_{#1}}}

\newcommand{\statenew}{\ensuremath{\state'}}
\newcommand{\statenewh}[1]{\ensuremath{\state'_{#1}}}

\newcommand{\State}{\ensuremath{S}}
\newcommand{\Statenew}{\ensuremath{\State'}}
\newcommand{\diff}{\ensuremath{d}}

\newcommand{\bx}{\ensuremath{\boldsymbol{x}}}

\DeclarePairedDelimiterX{\anglep}[1]{(}{)}{#1}
\makeatletter
\newcommand{\@spanstar}[1]{{\rm span}\anglep*{#1}}
\newcommand{\@spannostar}[2][]{{\rm span}\anglep[#1]{#2}}
\newcommand{\Span}{\@ifstar\@spanstar\@spannostar}
\makeatother

\DeclarePairedDelimiterX{\dfun}[2]{(}{)}{#1 \;\delimsize\|\; #2}
\makeatletter
\newcommand{\@trunstar}[2]{\chi^2\dfun*{#1}{#2}}
\newcommand{\@trunnostar}[3][]{\chi^2\dfun[#1]{#2}{#3}}
\newcommand{\chisq}{\@ifstar\@trunstar\@trunnostar}
\makeatother

\DeclarePairedDelimiterX{\inprod}[2]{\langle}{\rangle}{#1, \, #2}

\DeclarePairedDelimiterX{\kulldiv}[2]{(}{)}{#1\;\delimsize\|\;#2}
\makeatletter
\newcommand{\@kullstar}[2]{D_{\text{KL}}\kulldiv*{#1}{#2}}
\newcommand{\@kullnostar}[3][]{D_{\text{KL}}\kulldiv[#1]{#2}{#3}}
\newcommand{\kull}{\@ifstar\@kullstar\@kullnostar}
\makeatother

\makeatletter
\newcommand{\@hilinstar}[2]{\inprod*{#1}{#2}_{\RKHS}}
\newcommand{\@hilinnostar}[3][]{\inprod[#1]{#2}{#3}_{\RKHS}}
\newcommand{\hilin}{\@ifstar\@hilinstar\@hilinnostar}
\makeatother

\makeatletter
\newcommand{\@mudatainstar}[2]{\inprod*{#1}{#2}_{\distrdata}}
\newcommand{\@mudatainnostar}[3][]{\inprod[#1]{#2}{#3}_{\distrdata}}
\newcommand{\mudatain}{\@ifstar\@mudatainstar\@mudatainnostar}
\makeatother

\makeatletter
\newcommand{\@mudatahinstar}[3]{\inprod*{#2}{#3}_{\distrdata}}
\newcommand{\@mudatahinnostar}[4][]{\inprod[#1]{#3}{#4}_{\distrdata}}
\newcommand{\mudatahin}{\@ifstar\@mudatahinstar\@mudatahinnostar}
\makeatother

\DeclarePairedDelimiterX{\defabs}[1]{|}{|}{#1}
\makeatletter
\newcommand{\@absstar}[1]{\defabs*{#1}}
\newcommand{\@absnostar}[2][]{\defabs[#1]{#2}}
\newcommand{\abs}{\@ifstar\@absstar\@absnostar}
\makeatother

\DeclarePairedDelimiterX{\norm}[1]{\|}{\|}{#1}
\makeatletter
\newcommand{\@normstar}[1]{\norm*{#1}_{\RKHS}}
\newcommand{\@normnostar}[2][]{\norm[#1]{#2}_{\RKHS}}
\newcommand{\hilnorm}{\@ifstar\@normstar\@normnostar}
\makeatother

\DeclareFontFamily{U}{matha}{\hyphenchar\font45}
\DeclareFontShape{U}{matha}{m}{n}{
	<-6> matha5 <6-7> matha6 <7-8> matha7
	<8-9> matha8 <9-10> matha9
	<10-12> matha10 <12-> matha12
}{}
\DeclareSymbolFont{matha}{U}{matha}{m}{n}

\DeclareFontFamily{U}{mathx}{\hyphenchar\font45}
\DeclareFontShape{U}{mathx}{m}{n}{
	<-6> mathx5 <6-7> mathx6 <7-8> mathx7
	<8-9> mathx8 <9-10> mathx9
	<10-12> mathx10 <12-> mathx12
}{}
\DeclareSymbolFont{mathx}{U}{mathx}{m}{n}

\DeclareMathDelimiter{\vvvert} {0}{matha}{"7E}{mathx}{"17}%

\DeclarePairedDelimiterX{\opnorm}[1]{\vvvert}{\vvvert}{#1}

\makeatletter
\newcommand{\@hilopnormstar}[1]{\opnorm*{#1}_{\RKHS}}
\newcommand{\@hilopnormnostar}[2][]{\opnorm[#1]{#2}_{\RKHS}}
\newcommand{\hilopnorm}{\@ifstar\@hilopnormstar\@hilopnormnostar}
\makeatother

\makeatletter
\newcommand{\@muopnormstar}[1]{\opnorm*{#1}_{\distr}}
\newcommand{\@muopnormnostar}[2][]{\opnorm[#1]{#2}_{\distr}}
\newcommand{\muopnorm}{\@ifstar\@muopnormstar\@muopnormnostar}
\makeatother

\makeatletter
\newcommand{\@mudataopnormstar}[1]{\opnorm*{#1}_{\distrdata}}
\newcommand{\@mudataopnormnostar}[2][]{\opnorm[#1]{#2}_{\distrdata}}
\newcommand{\mudataopnorm}{\@ifstar\@mudataopnormstar\@mudataopnormnostar}
\makeatother

\makeatletter
\newcommand{\@supnormstar}[1]{\norm*{#1}_{\infty}}
\newcommand{\@supnormnostar}[2][]{\norm[#1]{#2}_{\infty}}
\newcommand{\supnorm}{\@ifstar\@supnormstar\@supnormnostar}
\makeatother

\makeatletter
\newcommand{\@munormstar}[1]{\norm*{#1}_{\distr}}
\newcommand{\@munormnostar}[2][]{\norm[#1]{#2}_{\distr}}
\newcommand{\munorm}{\@ifstar\@munormstar\@munormnostar}
\makeatother

\makeatletter
\newcommand{\@mudatanormstar}[1]{\norm*{#1}_{\distrdata}}
\newcommand{\@mudatanormnostar}[2][]{\norm[#1]{#2}_{\distrdata}}
\newcommand{\mudatanorm}{\@ifstar\@mudatanormstar\@mudatanormnostar}
\makeatother

\makeatletter
\newcommand{\@xidatanormstar}[1]{\norm*{#1}_{\sdistrdata}}
\newcommand{\@xidatanormnostar}[2][]{\norm[#1]{#2}_{\sdistrdata}}
\newcommand{\xidatanorm}{\@ifstar\@xidatanormstar\@xidatanormnostar}
\makeatother

\makeatletter
\newcommand{\@distrnormstar}[2]{\norm*{#1}_{#2}}
\newcommand{\@distrnormnostar}[3][]{\norm[#1]{#2}_{#3}}
\newcommand{\distrnorm}{\@ifstar\@distrnormstar\@distrnormnostar}
\makeatother

\makeatletter
\newcommand{\@psinormstar}[2]{\norm*{#2}_{\psi_{#1}}}
\newcommand{\@psinormnostar}[3][]{\norm[#1]{#3}_{\psi_{#2}}}
\newcommand{\psinorm}{\@ifstar\@psinormstar\@psinormnostar}
\makeatother

\newcommand{\Feature}{\ensuremath{\boldsymbol{\phi}}}
\newcommand{\Featurenew}{\ensuremath{\boldsymbol{\psi}}}
\newcommand{\featureh}[1]{\ensuremath{\phi}}
\newcommand{\featurebarh}[1]{\ensuremath{\overline{\boldsymbol{\phi}}_{#1}}}

\newcommand{\FeatureSet}{\Featureset}
\newcommand{\Featureset}{\ensuremath{\Upphi}}

\newcommand{\radius}{\ensuremath{\rho}}

\newcommand{\metric}{\ensuremath{d}}
\newcommand{\Lip}{\ensuremath{L}}

\newcommand{\Diff}{\ensuremath{\zeta}}

\newcommand{\supZ}{\ensuremath{Z_{\numobs}}}

\newcommand{\policy}{\ensuremath{{\boldsymbol{\pi}}}}

\newcommand{\policyhat}{\ensuremath{\widehat{\policy}}}
\newcommand{\policystar}{\ensuremath{\policy^{\star}}}
\newcommand{\policystarh}[1]{\ensuremath{\pi^{\star}_{#1}}}

\newcommand{\policyh}[1]{\ensuremath{\pi_{#1}}}

\newcommand{\policyrefh}[1]{\ensuremath{\policyh{#1}^{\dagger}}}
\newcommand{\policyhath}[1]{\ensuremath{\widehat{\pi}_{#1}}}

\newcommand{\Ho}{\ensuremath{H}}

\newcommand{\ho}{\ensuremath{h}}

\newcommand{\honew}{\ensuremath{\ho'}}
\newcommand{\honewnew}{\ensuremath{j}}

\newcommand{\Qfun}{\ensuremath{{\boldsymbol{\qfun}}}}
\newcommand{\Qfunhat}{\ensuremath{\boldsymbol{\widehat{\qfun}}}}
\newcommand{\Qfunstar}{\ensuremath{\Qfun^{\star}}}
\newcommand{\qfun}{\ensuremath{f}}

\newcommand{\qfunh}[1]{\ensuremath{\qfun_{#1}}}

\newcommand{\qfunhat}{\ensuremath{\boldsymbol{\widehat{f}}\!\,}}
\newcommand{\qfunhath}[1]{\ensuremath{\widehat{f}_{#1}}}
\newcommand{\qfunstar}{\ensuremath{\qfun^{\star}}}
\newcommand{\qfunstarh}[1]{\ensuremath{\qfunstar_{#1}}}

\newcommand{\action}{\ensuremath{a}}
\newcommand{\Action}{\ensuremath{A}}

\newcommand{\actionh}[1]{\ensuremath{\action_{#1}}}
\newcommand{\Actionh}[1]{\ensuremath{\Action_{#1}}}

\newcommand{\actionnew}{\ensuremath{\action'}}

\newcommand{\ActionSp}{\ensuremath{\mathcal{A}}}

\newcommand{\ValueErr}[1]{\ensuremath{ \valuescalar(\policystarh{})-
    \valuescalar(#1)}}

\newcommand{\BellErr}{\ensuremath{\varepsilon}}

\newcommand{\BellErrh}[1]{\ensuremath{\BellErr_{#1}}}

\newcommand{\RadBase}{\ensuremath{b}}
\newcommand{\Radphi}{\ensuremath{\RadBase_{\RKHS}}}

\newcommand{\Radphistar}{\ensuremath{\Radphi}}

\newcommand{\valuescalar}{\ensuremath{J}}

\newcommand{\stderrdatah}[1]{\ensuremath{\sigma_{#1, \Data}}}
\newcommand{\stderrhatdatah}[1]{\ensuremath{\widehat{\sigma}_{#1, \Data}}}
\newcommand{\stderrh}[1]{\ensuremath{\sigma_{#1}}}

\newcommand{\zerovec}{\ensuremath{\boldsymbol{0}}}

\newcommand{\fun}{\ensuremath{\boldsymbol{f}}}
\newcommand{\funh}[1]{\ensuremath{f_{#1}}}
\newcommand{\funnew}{\ensuremath{\boldsymbol{g}}}
\newcommand{\funnewh}[1]{\ensuremath{g_{#1}}}

\newcommand{\gfunh}[1]{\funnewh{#1}}

\newcommand{\Diffun}{\ensuremath{D}}
\newcommand{\Diffunstar}{\ensuremath{D^{\star}}}

\newcommand{\MOp}{\ensuremath{\mathcal{P}}}

\newcommand{\MOph}[1]{\ensuremath{\MOp_{#1}}}

\newcommand{\MOpstarh}[1]{\ensuremath{\MOp^{\star}_{#1}}}
\newcommand{\MOppih}[2]{\ensuremath{\MOph{#1}^{#2}}}

\newcommand{\MOppihtoh}[3]{\ensuremath{\MOph{#1, #2}^{#3}}}

\newcommand{\MOpstarhtoh}[2]{\ensuremath{\MOpstarh{#1, #2}}}

\newcommand{\DDiffOp}{\ensuremath{\Delta \DiffOp}}
\newcommand{\DDiffOpone}{\ensuremath{\nu_2}}
\newcommand{\DDiffOptwo}{\ensuremath{\nu_1}}
\newcommand{\DDiffOpthree}{\ensuremath{\nu_3}}
\newcommand{\DDiffOpfour}{\ensuremath{\nu_4}}

\newcommand{\metrich}[1]{\ensuremath{\metric_{#1}}}

\newcommand{\Lipf}[1]{\ensuremath{\Lip_{f}}}

\newcommand{\constrain}{\ensuremath{\boldsymbol{g}}}
\newcommand{\constraini}[1]{\ensuremath{g_{#1}}}
\newcommand{\blambda}{\ensuremath{\boldsymbol{\lambda}}}
\newcommand{\blambdatil}{\ensuremath{\widetilde{\blambda}\!\,}}
\newcommand{\blambdastar}{\ensuremath{\blambda^{\star}\!\,}}

\newcommand{\lambdastari}[1]{\ensuremath{\lambda^{\star}_{#1}\!\,}}

\newcommand{\numcon}{\ensuremath{m}}

\newcommand{\Hess}{\ensuremath{\boldsymbol{H}}}
\newcommand{\subspace}{\ensuremath{\mathds{G}}}
\newcommand{\projG}{\ensuremath{\proj_{\subspace}}}

\newcommand{\Lag}{\ensuremath{\mathcal{L}}}

\newcommand{\WMt}{\ensuremath{\boldsymbol{W}}}

\newcommand{\constraintil}{\ensuremath{\widetilde{g}}}

\newcommand{\regret}{\ensuremath{\mbox{Regret}}}

\newcommand{\smallo}{\ensuremath{o}}
\newcommand{\bigO}{\ensuremath{\mathcal{O}}}

\newenvironment{carlist}
{\begin{list}{$\bullet$}
		{\setlength{\topsep}{0.1in} \setlength{\partopsep}{0in}
			\setlength{\parsep}{0.1in} \setlength{\itemsep}{\parskip}
			\setlength{\leftmargin}{0.15in} \setlength{\rightmargin}{0.08in}
			\setlength{\listparindent}{0in} \setlength{\labelwidth}{0.08in}
			\setlength{\labelsep}{0.1in} \setlength{\itemindent}{0in}}}
	{\end{list}}

\newcommand{\bcar}{\begin{carlist}}
	\newcommand{\ecar}{\end{carlist}}

\newcommand{\regular}{\ensuremath{\Lambda}}
\newcommand{\regularh}[1]{\ensuremath{\regular_{#1}}}

\newcommand{\pos}{\ensuremath{p}}
\newcommand{\posmin}{\ensuremath{\pos_{\min}}}
\newcommand{\posmax}{\ensuremath{\pos_{\max}}}
\newcommand{\posgoal}{\ensuremath{\pos_{\rm goal}}}
\newcommand{\vel}{\ensuremath{v}}
\newcommand{\velmin}{\ensuremath{\vel_{\min}}}
\newcommand{\velmax}{\ensuremath{\vel_{\max}}}
\newcommand{\for}{\ensuremath{f}}
\newcommand{\formin}{\ensuremath{\for_{\min}}}
\newcommand{\formax}{\ensuremath{\for_{\max}}}
\newcommand{\mou}{\ensuremath{m}}
\newcommand{\moudiff}{\ensuremath{\mou'}}
\newcommand{\noisepos}{\ensuremath{\sigma_{\pos}}}
\newcommand{\noisevel}{\ensuremath{\sigma_{\vel}}}

\newcommand{\featurepos}{\ensuremath{\boldsymbol{\phi}_{\pos}}}
\newcommand{\featureposj}[1]{\ensuremath{\phi_{\pos, #1}}}
\newcommand{\featurevel}{\ensuremath{\boldsymbol{\phi}_{\vel}}}
\newcommand{\featurevelj}[1]{\ensuremath{\phi_{\vel, #1}}}
\newcommand{\featurefor}{\ensuremath{\boldsymbol{\phi}_{\for}}}
\newcommand{\vectorize}{\ensuremath{{\rm vec}}}

\newcommand{\truncate}{\ensuremath{\Psi}}

\newcommand{\Featurestar}{\ensuremath{\Feature^{\star}}}

\newcommand{\Loss}{\ensuremath{\mathcal{L}}}

\newcommand{\bradius}{\ensuremath{\boldsymbol{\rho}}}
\newcommand{\bbellerr}{\ensuremath{\boldsymbol{\BellErrh{}}}}

\newcommand{\PlainNeigh}{\ensuremath{\mathcal{N}}}
\newcommand{\Neigh}{\ensuremath{\PlainNeigh(\bradius)}}

\newcommand{\exprad}{\ensuremath{\varrho}}

\newcommand{\PhaseOne}{{Phase 1}}
\newcommand{\PhaseTwo}{{Phase 2}}

\newcommand\xqed[1]{%
\leavevmode\unskip\penalty9999 \hbox{}\nobreak\hfill
\quad\hbox{#1}}
\newcommand{\myexample}[3]{\begin{example}[#1]
\label{#2}{#3}\xqed{$\triangle$}\vspace{1em}
\end{example}}

\begin{document}
  
\begin{center}
  {\bf \LARGE Taming ``data-hungry'' reinforcement learning?
    \\ Stability in continuous state-action spaces} \\
  
	\vspace{1em}
    {\large{
    		\begin{tabular}{ccc}
    			Yaqi Duan$^\diamond$ && Martin
                        J. Wainwright$^{\dagger}$
    		\end{tabular}
    		
    		\medskip

    		\begin{tabular}{c}
                  Laboratory for Information \& Decision Systems \\
                  Statistics and Data Science Center \\
                        EECS \& Mathematics \\
    			Massachusetts Institute of Technology$^\dagger$
    		\end{tabular}
    		
    		\medskip 
    		\begin{tabular}{c}
                  Stern School of Business,  New York University$^{\diamond}$
    		\end{tabular}
    		
    }}
  \vspace{.6em}
  \today
\end{center}

\begin{center}
    {\bf Abstract} \\ \vspace{.6em}
  \begin{minipage}{0.9\linewidth}
 {\small ~~~~ We introduce a novel framework for analyzing
   reinforcement learning (RL) in continuous state-action spaces, and
   use it to prove fast rates of convergence in both off-line and
   on-line settings.  Our analysis highlights two key stability
   properties, relating to how changes in value functions and/or
   policies affect the Bellman operator and occupation measures.  We
   argue that these properties are satisfied in many continuous
   state-action Markov decision processes, and demonstrate how they
   arise naturally when using linear function approximation methods.
   Our analysis offers fresh perspectives on the roles of pessimism
   and optimism in off-line and on-line RL, and highlights the
   connection between off-line RL and transfer learning.  }
  \end{minipage}
\end{center}



\section{Introduction}
\label{sec:intro}

Many domains of science and engineering involve making a sequence of
decisions over time, with previous decisions influencing the future in
uncertain ways.  The central challenge is \yd{choosing} a
\emph{decision-making policy} that leads to desirable outcomes over a
longer period. For example, in the treatment of chronic diseases such
as diabetes~\cite{yu2021reinforcement}, a clinician can choose from a
range of treatments depending on the patient's history, and any such
policy can have uncertain effects on the patient's status at future
times.  In a rather different domain, the design of tokamak systems
for nuclear fusion requires learning policies for plasma control and
shaping~\cite{Deg22_Fusion}; here the actions or decisions are
effected via coils that are magnetically coupled to the plasma. Other
applications include inventory and pricing systems for
businesses~\cite{gijsbrechts2022can}; navigation systems in robotics
and autonomous driving~\cite{tai2017virtual, kiran2021deep}; resource
deployment for wildfire prevention and management~\cite{Alt22_Fire};
and optimization and control of industrial processes~\cite{Spi19}.

Markov decision processes provide a flexible framework for describing
such sequential problems, and reinforcement learning (RL) refers to a
broad class of data-driven methods for estimating policies.  Some
applications are data-rich, meaning that it is relatively inexpensive
to collect samples of states, actions and rewards from the underlying
process.  When given access to large sample sizes, RL methods have
proven to be very successful, with especially prominent examples in
competitive game-playing (e.g., AlphaGo and its
extensions~\cite{Sil17}).  However, many applications have far more
limited sample sizes---sometimes referred to as the ``small data''
setting---which renders deployment of RL more challenging. For
example, in healthcare applications, there is limited data available
for certain types of disease, or certain types of
patients~\cite{yu2021reinforcement}. Similarly, for portfolio
optimization in finance (e.g.,~\cite{RaoJel22}), effective data sizes
are often very limited due to lack of history, or underlying
non-stationarity. With limited data, characterizing and improving the
\emph{sample complexity} of RL methods---meaning the amount of data
required to learn near-optimal policies---becomes critical.

Considerable research effort has been devoted to studying RL sample
complexity in many settings, including generative models/simulators,
off-line observational studies, and on-line interactive
learning. Existing studies for either the generative or the off-line
settings
(e.g.,~\cite{jin2021pessimism,zanette2021provable,yin2022near}) give
procedures that, when applied to an dataset of size $\numobs$, yield a
value gap that decays at the rate $1/\sqrt{\numobs}$.  In the on-line
setting, there are various procedures that yield cumulative regret
that grows at the rate $\sqrt{T}$
(e.g.,~\cite{jin2018q,jin2020provably,jin2021bellman,du2021bilinear}).
In contrast, the main result of this paper is to formalize conditions,
suitable for RL in continuous domains, under which \emph{much faster
rates can be obtained using the same dataset.}  In particular, in
either the generative or off-line settings, our theory provides
conditions under which the value gap decays as quickly as $1/\numobs$.
So as a concrete example, obtaining a policy with value gap at most
$\epsilon = 1/100$ requires on the order of $\numobs = 100$ samples,
as opposed to the much larger sample size $\numobs = (100)^2 = 10^4$
required by the classical ``slow rate''.  Similarly, in the regret
setting, we reduce the classical $\sqrt{T}$ growth to a much better
$\log T$ rate.

As revealed by our analysis, these accelerated rates depend on certain
\emph{stability properties}, ones that---as we argue---are naturally
satisfied in many control problems with continuous state-action
spaces.  Roughly speaking, these conditions ensure that perturbing the
policy changes future outcomes by at most a quantity proportional to
the magnitude of the perturbation. In other words, the evolution of
the dynamic system depends in a ``smooth'' way on the influence of
decision policy.  Such notions of stability should be expected in
various controlled systems with continuous state-action spaces.  In
robotics, for example, a minor torque or motion perturbation that
occurs during a single step should not cause a notable deviation from
the intended trajectory. Similarly, in clinical treatment, slight
deviations in medication dosage should not significantly compromise
effectiveness or safety. In inventory management, a properly designed
supply chain should be able to handle minor variations due to supplier
delays or demand changes while approximately maintaining expected
inventory levels in the future. \vspace{1em}

\subsection{A simple illustrative example: Mountain Car}

\begin{figure}[!ht]
  \centering
  \begin{tabular}{ccc}
  ~~~
  \begin{minipage}{.32\textwidth}   
    \vspace{-17.5em} \widgraph{\textwidth}{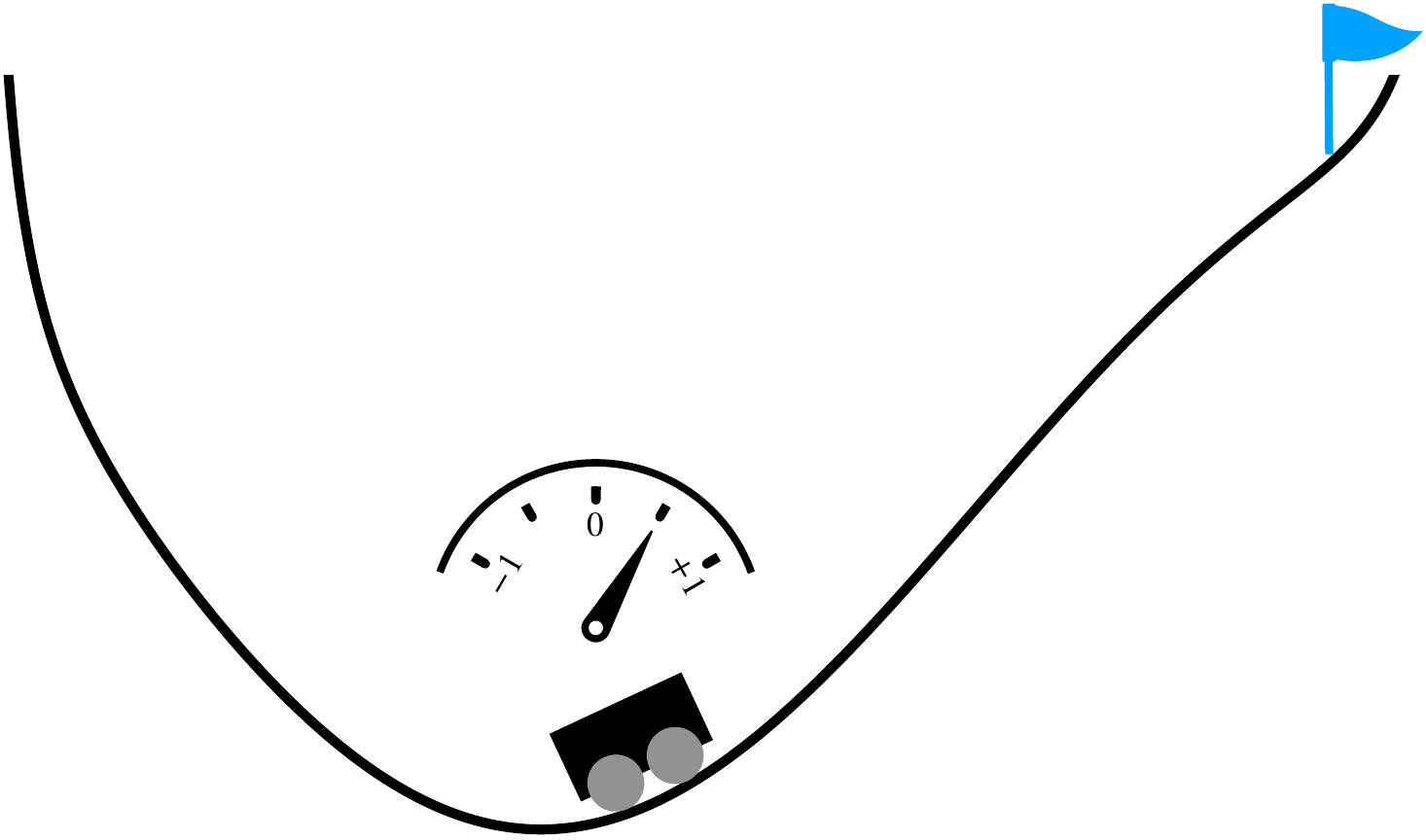}
  \end{minipage}
  & ~~ &\widgraph{0.55\textwidth}{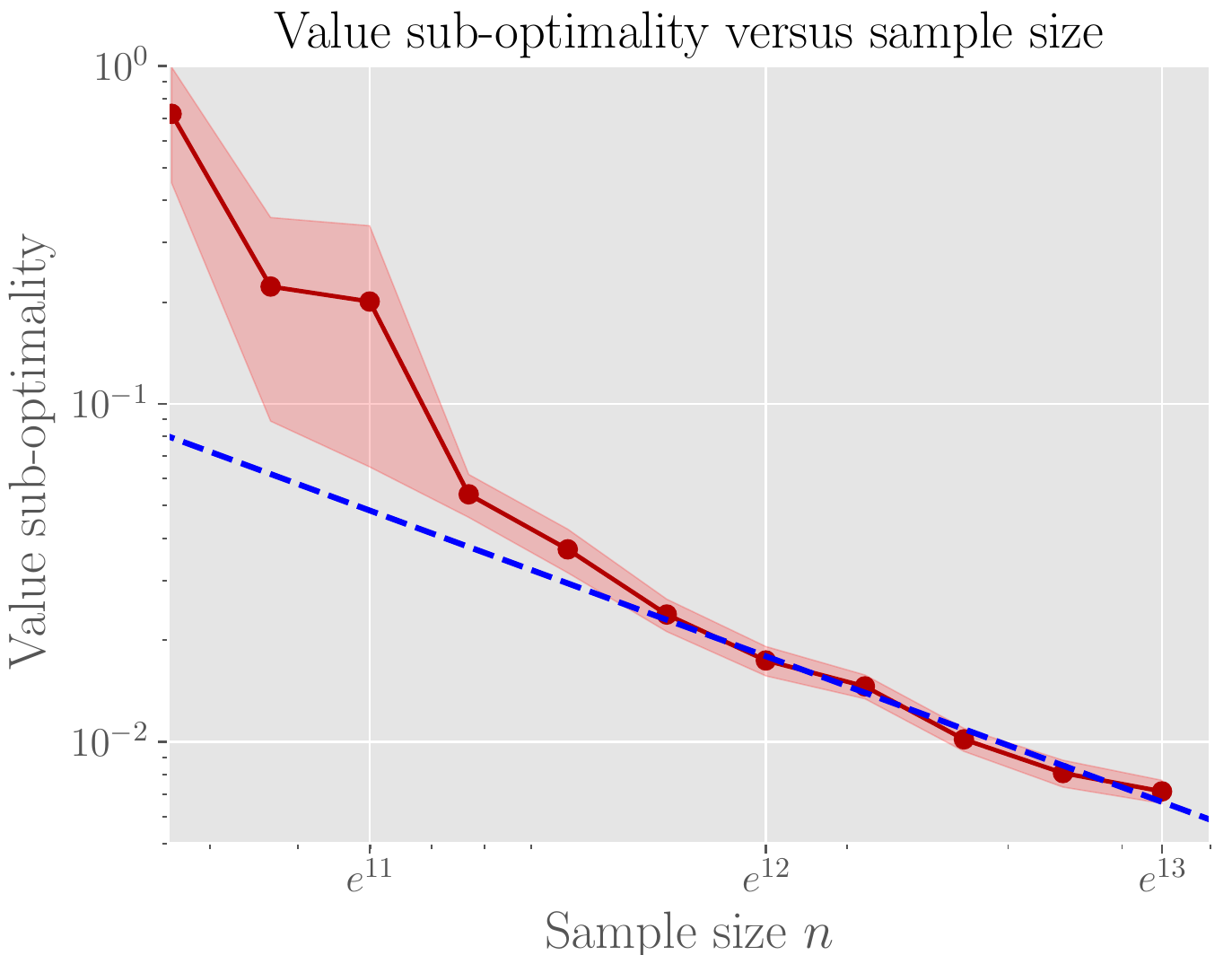}
  \\ (a) & & (b)\\ & &
  \end{tabular}
  \caption{Illustration of the ``fast rate'' phenomenon for fitted
    $Q$-iteration (FQI) applied to the Mountain Car problem.  (a) The
    Mountain Car problem is a canonical continuous state-action space
    control problem, in which the goal is to drive the car to the
    flag.  See~\Cref{sec:mountain_car} for further details.  (b) We
    used off-line FQI with linear function approximation to learn
    approximately optimal policies $\policyhath{\numobs}$ over a range
    of sample sizes $\numobs$.  Log-log plot of the value
    sub-optimality $\ValueErr{\policyhath{\numobs}}$ over sample sizes
    \mbox{$\numobs \in \big\{ \lfloor e^{k} \rfloor \bigm| k = 10.5,
      10.75, 11, \ldots, 13 \big\} = \{ 36315 , 46630, 59874, \ldots,
      442413 \}$.}  In the plot, each {\bf red point} represents the
    average value sub-optimality $\ValueErr{\policyhath{\numobs}}$
    estimated from $T = 80$ Monte Carlo trials.  The shaded area
    represents twice the standard errors.  The \mbox{\bf blue dashed
      line} represents the least-squares fit to the last $6$ data
    points.  This regression leads to the $95\%$ confidence interval
    $(-1.084, -0.905)$ for the underlying slope, indicative of a decay
    rate much faster than the typical $-0.5$ ``slow rate''.}
  \label{fig:mountaincar}
\end{figure}

The acceleration phenomenon---as well as the underlying
stability---can be observed in a simple instance of a continuous
control problem.  The so-called ``Mountain Car'' problem is a
benchmark example of a continuous control task.  As illustrated
in~\Cref{fig:mountaincar}(a), it involves a car positioned between two
hills, where the ultimate goal is to maneuver the car so as to reach
the top of the right-side hill by adjusting its acceleration. Due to
limited force, the car must learn a decision policy that causes it to
oscillate back and forth, using the potential energy to overcome the
hill. In our study, we employ offline reinforcement learning using
observations collected from the mountain car system. The physical
system itself is continuous and subject to noise, with nonlinear
dynamics governing the transitions. The control variable, acceleration
or force, is represented as a real number within interval $[-1,1]$.

In order to investigate the acceleration phenomenon, we learned
near-optimal policies for this problem in the off-line setting, using
linear methods with well-chosen basis functions to approximate the
value.  (See~\Cref{sec:mountain_car} for the full details of
experiments that produce the numerical results shown here.)  As
demonstrated in~\Cref{fig:mountaincar}(b), the value sub-optimality
behaves in an interesting way as a function of the sample size
$\numobs$.  Instead of decaying at the classical $1/\sqrt{\numobs}$
rate, we see that its rate is very well-approximated\footnote{The
approximation holds when disregarding transient behavior for small
sample sizes.} by the $1/\numobs$ rate, corresponding to a slope of
$-1$ on the log-log scale.  To the best of our knowledge, this
phenomenon has not been addressed in past work, possibly due to the
following two properties:
(i) the continuous state-action space that renders inapplicable
fast-rate analysis that depends on gaps or margins
(e.g.,~\cite{hu2021fast,nguyen2022instance}); and
(ii) the nonlinear dynamics, as contrasted with related
work~\cite{mania2019certainty} on the linear quadratic regulator
(LQR).

The theoretical analysis given in this paper sheds light on this
intriguing phenomenon.  In the specific setting of the ``Mountain
Car'' problem, we observe that small perturbations in the driving
policy $\policyhath{}$ results in only a modest deviation in future
trajectories of the car, with the magnitude of the deviation being
proportional to the perturbation size. Our theory shows that fast
rates can be guaranteed in the off-line setting whenever this property
holds, and we exhibit a broad family of continuous control tasks for
which it holds.


\subsection{Contributions of this paper}

With this high-level perspective in mind, let us summarize the key
contributions of this paper, which can be divided into three parts.

\paragraph{Fast rate of convergence:}

We develop a framework for analyzing RL in continuous state-action
spaces, and use it to prove a general result
(\Cref{thm:deterministic_star}) under which fast rates can be
obtained.  The key insight is that stability conditions lead to upper
bounds on the value sub-optimality that are proportional to the
\emph{squared} norm of Bellman residuals. This quadratic scaling
results in accelerated convergence compared to the standard linear
scaling obtained via arguments that isolate only a single copy of
Bellman residuals. In the off-line setting, this framework improves
convergence from a rate of $\numobs^{-\frac{1}{2}}$ to $\numobs^{-1}$,
while in on-line learning, it enhances the regret bound from
$\sqrt{T}$ to $\log T$.

\paragraph{Reconsidering pessimism and optimism principles:}
  
Our framework provides a novel perspective on the roles of pessimism
and optimism in off-line and on-line RL.  In the off-line setting, a
line of past work~\cite{jin2021pessimism,dean2020sample} has
established the utility of pessimistic or risk-averse approaches to
policy evaluation and optimization.  Pessimism serves to protect
against uncertainty associated with a fixed off-line data set.  On the
other hand, in the on-line setting, optimism drives exploration and
embraces uncertainty, making it fundamental in online
RL~\cite{jin2018q,jin2020provably,jin2021bellman,du2021bilinear,foster2021statistical},
where learning occurs through trial and error.

Our theory reveals that there are settings in which \emph{neither
pessimism nor optimism} are required for effective policy
optimization---in particular, they are not required as long as one has
a sufficiently accurate pilot estimate $\policyhath{}$.  Thus, while
the use of pessimism or optimism can be useful in obtaining such a
pilot estimate, they are not needed in later stages of training.
Moreover, our analysis shows that some procedures based on certainty
equivalence can achieve fast-rate convergence, showing that the
benefits gained from incorporating additional pessimism or optimism
measures may be limited in this context.

\paragraph{Connecting off-line RL with transfer learning:}

Our theory relates value sub-optimality to the Bellman residual as
measured under a problem-specific norm.  For instance, in the case of
linear function approximation (as discussed
in~\Cref{sec:lin_fun_star}), the norm is induced by the occupation
measures of the optimal policy $\policystarh{}$. In other examples
such as the linear quadratic regulator (LQR), the norms can take on a
more complex form; in certain cases, it can even capture the Bellman
variance\footnote{See the papers~\cite{duan2021optimal,duan2022policy}
for more details on Bellman variances.}  associated with the optimal
policy $\policystarh{}$.  We explore this phenomenon in a forthcoming
paper \cite{X}.  In this regard, off-line RL can be seen as a form of
transfer learning, where the goal is to minimize the loss under
covariate shifts from the distribution of historical data to a
distribution related to the optimal policy.

While the bulk of our theory is of a general nature, we explore in
depth the special case of RL methods based on linear approximation to
value functions.  In this concrete setting, we can give an intuitive
and geometric interpretation of the stability conditions that underlie
our analysis, along with the connection to covariate shift.  Stability
is related to the curvature of the set of feature vectors achievable
by varying the action (with the state fixed), and covariate shift is
reflected through a comparison of covariance matrices.


\subsection{Related work}

In this section, we discuss related work having to do with fast rates
in optimization and statistics.

\paragraph{Fast rates in stochastic optimization and risk minimization:}
Many statistical estimators (e.g., likelihood methods, empirical risk
minimization) are based on minimizing a data-dependent objective
function.  It is now \yd{well-understood that} the local geometry around the
optimum determines whether fast rates can be obtained.  For instance,
when the loss function exhibits some form of strong convexity (such as
exp-concave loss) or strict saddle properties, it can lead to
significant reductions in additive regret from $\bigO(\sqrt{T})$ to
just $\bigO(\log T)$ in stochastic approximation
(e.g.,~\cite{hazan2007logarithmic}), or a decrease in the error rate
from $\numobs^{-\frac{1}{2}}$ to $\numobs^{-1}$ in empirical risk
minimization~\cite{koren2015fast,gonen2017fast}.  The theory of
localization is also instrumental in characterizing this
phenomenon. To achieve the sharpest analysis, it is essential to
determine an appropriate radius in the measurement of function class
complexity that accurately reflects the curvature of the loss function
near the optimum~\cite{bartlett2005local,liang2015learning}.  These
fast rate phenomena rely \yd{on} a form of stability, one which relates
the similarity of functions to the closeness of their optima.

Our work shares a similar spirit, in that we isolate certain stability
conditions that ensure fast rate convergence in RL.  To the best of
our knowledge, there is currently no literature systematically
discussing the relationship between our stability conditions and fast
rate convergence in the context of RL, or how the inherent curvature
affects the convergence rate. Our work, in making this connection
explicit and rigorous, provides a framework for analysis of
value-based RL methods, akin to the role played by stability analysis
in statistical learning.

\paragraph{Fast rates in reinforcement learning:}

In the RL literature, there are various lines of work related to fast
rates, but the underlying mechanisms are typically different from
those considered here.  For problems with discrete state-action
spaces, there is a line of recent
work~\cite{hu2021fast,he2021logarithmic, wang2022gap,
  nguyen2022instance} that performs gap/marginal-dependent analyses of
RL algorithms.  These papers focus on action spaces with finite
cardinality, for which it is reasonable to assume a strictly positive
gap between the value of an optimal action relative to a sub-optimal
one. However, such separation assumptions are not helpful for
continuous action spaces, since (under mild Lipschitz conditions) we
can find sub-optimal actions with value\yd{s} arbitrarily close to that of
an optimal one.  Other work for discrete state-action
spaces~\cite{rashidinejad2021bridging} has shown convergence rates in
off-line RL are influenced by data quality, with a nearly-expert
dataset enabling faster rate. In contrast, our analysis reveals that
for off-line RL in continuous domains, fast convergence can occur
whether or not the dataset has good coverage properties.

An important sub-class of continuous state-action problems are those
with linear dynamics and quadratic reward functions (LQR for short).
For such problems, it has been
shown~\cite{mania2019certainty,recht2019tour} that value
sub-optimality can be connected with the squared error in system
identification.  Our general theory can also be used to derive
guarantees for LQR problems, as we explore in more detail in a
follow-up paper~\cite{X}.  Stability also arises in the analysis of
(deterministic) policy optimization and Newton-type
algorithms~\cite{puterman1979convergence,bertsekas2022lessons}, where
it is possible to show superlinear convergence in a local
neighborhood.  This accelerated rate stems from the smoothness of the
on-policy transition operator $\TransOppih{}{\policyh{\funh{}}}$ with
respect to changes in the value function $\funh{}$; for instance, see
condition (10) in Puterman and
Brumelle~\cite{puterman1979convergence}.  Our framework exploits
related notions of smoothness, but is tailored to the stochastic
setting of reinforcement learning, in which understanding the effect
of function approximation and finite sample sizes is essential.


\vspace{1em}


\section{Fast rates for value-based reinforcement learning}

Let us now set up and state the main result of this paper.  We begin
in~\Cref{SecMDP} with background on Markov decision processes (MDPs)
and value-based methods, before turning to the statement of our main
result in~\Cref{SecStable}.  In~\Cref{SecIntuition}, we provide
intuition for why stability leads to faster rates, and discuss
consequences for both the off-line and on-line settings of RL.

\subsection{Markov decision processes and value-based methods}
\label{SecMDP}

Here we provide a brief description of Markov decision processes,
along with the idea of a value-based method for approximating an
optimal policy.  We refer the reader to some standard
references~\cite{bertsekas1996neuro,sutton2018reinforcement} for more
detailed background.

\subsubsection{Basic set-up}
We consider decision-making over $\Ho$ stages, as described by a
Markov decision process (MDP) with state space~$\StateSp$ and action
space $\ActionSp$.  The evolution of the state over time is specified
by a family of transition kernels \mbox{$\TransOp =\{
  \TransOph{\ho}\}_{\ho =1}^{\Ho-1}$,} where the transition kernel
$\TransOph{\ho}$ maps each state-action pair $(\state, \action) \in
\StateSp \times \ActionSp$ to a distribution $\TransOph{\ho}(\cdot
\mid \state, \action)$ over the state space $\StateSp$.  Given an
initial state~$\state_1$ and a sequence of actions $(\action_1,
\action_1, \ldots, \action_{\Ho-1}, \action_\Ho)$, these transition
dynamics generate a state sequence $(\state_1, \state_2, \ldots,
\state_{\Ho})$ via $\stateh{\ho+1} \sim \TransOph{\ho}(\cdot \mid
\stateh{\ho}, \actionh{\ho})$ \mbox{for $\ho = 1, 2, \ldots, \Ho-1$.}
An additional ingredient is the family of reward functions
$\{\rewardh{\ho}\}_{\ho =1}^\Ho$. At time $\ho$, the mapping
$(\state_\ho, \action_\ho) \mapsto \rewardh{\ho}(\state_\ho,
\action_\ho) \in \real$ specifies the reward received when in
state-action pair $(\state_\ho, \action_\ho)$.  In this paper, we
assume that the rewards~\mbox{$\rewardh{\ho}: \StateSp \times
  \ActionSp \rightarrow \Real$} are known; however, this condition can
be relaxed.

A policy~$\policyh{\ho}$ at time $\ho$ is a mapping from any
state~$\state$ to a distribution $\policyh{\ho}(\cdot \mid \state)$
over the action space~$\ActionSp$.  If the support of
$\policyh{\ho}(\cdot \mid \state)$ is a singleton, we also let
\mbox{$\policyh{\ho}(\state) \in \ActionSp$} denote the single action to
be chosen at state $\state$.  Given an initial distribution
$\sdistrinit$ over the states at time $\ho = 1$, the \emph{expected
reward} obtained by choosing actions according to a policy sequence
$\policy = (\policyh{1}, \ldots, \policyh{\Ho})$ is given by
\begin{align}
\label{eq:def_valuescalar}
  \valuescalar(\policy) & \equiv \valuescalar(\policy; \sdistrinit)
  \defn \Exp_{\sdistrinit, \policy} \bigg[ \sum_{\ho=1}^{\Ho}
    \rewardh{\ho}(\Stateh{\ho}, \Actionh{\ho}) \bigg],
\end{align}
where $\Stateh{1} \sim \sdistrinit$, $\Stateh{\ho+1} \sim
\TransOph{\ho}(\cdot \mid \Stateh{\ho}, \Actionh{\ho})$ and
$\Actionh{\ho} \sim \policyh{\ho}(\cdot \mid \Stateh{\ho})$ for
\mbox{$\ho = 1, 2, \ldots, \Ho$.}  Our goal is to estimate an \emph{optimal
policy} $\policystar \in \arg \max_{\policy } \valuescalar(\policy)$.


\subsubsection{Value functions and Bellman operators}

We now describe the connection between the expected return
$\valuescalar(\policy)$ and value functions.  Starting from a given
state-action pair $(\state, \action)$ at stage $\ho$, the expected
return over subsequent stages defines the \emph{state-action value
function}
\begin{align}
\label{EqnDefnQfunPol}
  \qfunh{\ho}^{\policy}(\state, \action) \defn \Exp_{\policy{}} \bigg[
    \, \myssum{\honew=\ho}{\Ho} \, \reward(\State_{\honew},
    \Action_{\honew}) \biggm| \Stateh{\ho} = \state, \Actionh{\ho} =
    \action \, \bigg].
\end{align}
Here the expectation is taken over a sequence $(\State_{\ho} = \state,
\Action_{\ho} = \action, \State_{\ho+1}, \Action_{\ho+1},
\State_{\ho+2}, \Action_{\ho+2}, \ldots, \State_{\Ho}, \Action_{\Ho})$
governed by the transition kernels
$\{\TransOph{\honew}\}_{\honew=\ho}^{\Ho-1}$ and the decision policy
$\policy = (\policyh{1}, \policyh{2}, \ldots, \policyh{\Ho})$.  The
sequence of functions $\Qfun^\policy = (\qfunh{1}^\policy, \ldots,
\qfunh{\Ho}^\policy)$ defined by equation~\eqref{EqnDefnQfunPol} \yd{is}
known as the \emph{$Q$-functions} associated with $\policy$.

The $Q$-functions $\Qfun^\policy$ have an important connection with
the \emph{Bellman evaluation operator} for~$\policy$.  For any policy
$\policy$ and any function $\funh{} \in \Real^{\StateSp \times
  \ActionSp}$, we introduce the shorthand
\begin{subequations}
\begin{align}
f(\state, \policyh{}(\state)) \defn \int_{\ActionSp} \funh{}(\state,
\action) \; \policyh{}(\diff \action \mid \state).
\end{align}
At stage $\ho$, we extend the transition functions to a linear
operator that maps a function $f$ on the the state-action space
$\StateSp \times \ActionSp$ to a new function on the state-action
space as follows:
\begin{align}
  (\TransOppih{\ho}{\policy} \funh{})(\state, \action) \defn
  \int_{\StateSp \times \ActionSp} \funh{} \big(\statenew,
  \policyh{\ho+1}(\statenew) \big) \; \TransOph{\ho}(\diff
  \statenew \mid \state, \action) \qquad & \mbox{for any function
    $\funh{} \in \Real^{\StateSp \times \ActionSp}$}\,.
\end{align}
With this notation, the \emph{Bellman evaluation operator} at stage
$\ho$ takes the form
\begin{align}
\label{EqnDefnBellmanEval}  
  (\BellOph{\ho}{\policy}f)(\state, \action) & \defn
\rewardh{\ho}(\state, \action) + (\TransOppih{\ho}{\policy}
\funh{})(\state, \action).
\end{align}
\end{subequations}
From classical dynamic programming, the $Q$-functions $\Qfun^\policy$
must satisfy the Bellman relations
\begin{align}
\qfunh{\ho}^\policy(\state, \action) & = (\BellOph{\ho}{\policy}
\qfunh{\ho+1}^\policy)(\state, \action) \qquad \mbox{for $\ho = 1,
  \ldots, \Ho -1$.}
\end{align}
Furthermore, these $Q$-value functions are connected to the expected
returns $\valuescalar$ under $\policy$; in particular, recalling the
definition~\eqref{eq:def_valuescalar}, for any initial distribution
$\sdistrinit$ over the states, we can use the value function
$\qfunh{1}^\policy$ to compute the expected return as
\mbox{$\valuescalar(\policy; \sdistrinit) \; = \; \Exp_{\State \sim
    \sdistrinit} \Big[ \qfunh{1}^{\policy{}} \big(\State,
    \policyh{1}(\State) \big) \Big]$.}


\paragraph{Bellman principle for optimal policies:}

Under mild regularity conditions, there is at least one policy
$\policystar$ such that, for any other policy $\policy$, we have
$\qfunh{\ho}^{\policystar}\!(\state, \action) \geq
\qfunh{\ho}^\policy(\state, \action)$, for any $\ho \in [\Ho]$, and
uniformly over all state-action pairs $(\state, \action)$.  Any
optimal policy $\policystar$ must be greedy with respect to the
optimal $Q$-function $\Qfunstar$. This function $\Qfunstar$ is
determined by the Bellman optimality operator, defined as
\begin{align}
  \label{eq:def_BellOpstar}
  (\BellOpstarh{\ho} f)(\state, \action) \defn \rewardh{\ho}(\state,
  \action) + \Exp_{\ho} \big[ \max_{\actionnew \in \ActionSp}
    f(\Statenew, \actionnew) \bigm| \state, \action \big], \qquad
  \text{where $\Statenew \sim \TransOph{\ho}(\cdot \mid \state,
    \action)$ .}
\end{align}
By classical dynamic programming, the optimal $Q$-function $\Qfunstar$
is obtained by setting $\qfunstar_\Ho = \reward_\Ho$, and then
recursively computing \mbox{$\qfunstar_{\ho} = \BellOpstarh{\ho}
  \qfunstar_{\ho+1}$} for \mbox{$\ho = \Ho - 1, \ldots, 2, 1$.}

\subsubsection{Value-based RL methods}

The main result of this paper applies to a broad class of methods for
reinforcement learning.  They are known as \emph{value-based}, due to
their reliance on the following two step approach for approximating
an optimal policy $\policystar$:
\begin{enumerate}
\item[(1)] Construct an estimate \mbox{$\Qfunhat = (\qfunhath{1},
  \ldots, \qfunhath{\Ho})$} of the optimal \mbox{value function}
  \mbox{$\Qfunstar = (\qfunstarh{1}, \ldots, \qfunstarh{\Ho})$.}
\item[(2)] Use $\Qfunhat$ to compute the greedy-optimal policy
\begin{align}
  \label{EqnGreedyHat}
  \policyhath{\ho}(\state) & \in \arg \max_{\action}
  \qfunhath{\ho}(\state, \action) \qquad \mbox{for $\ho = 1, 2,
    \ldots, \Ho$.}
\end{align}
\end{enumerate}
It should be noted that there is considerable freedom in the design of
a value-based method, since different methods can be used to
approximate value functions in Step 1.  Rather than applying to a
single method, our main result applies to a very broad class of these
methods.

Underlying any value-based method is a class $\RKHS$ of functions
$(\state, \action) \mapsto f(\state, \action)$ used to approximate the
state-action value functions.  In general, different function classes
may be selected at each stage $\ho = 1, 2, \ldots, \Ho$; here, so as
to reduce notational clutter, we assume that the same function class
$\RKHS$ is used for each stage.  Moreover, we assume that the function
class $\RKHS$ is rich enough---relative to the Bellman evaluation
operators~\eqref{EqnDefnBellmanEval}---to ensure that for any greedy
policy $\policy$ induced by some $\fun = (\funh{1}, \ldots,
\funh{\Ho}) \in \RKHS^{\Ho}$, we have the inclusion
\begin{align}
\label{EqnCompleteness}  
 \BellOph{\ho}{\policy} \, \RKHS \subseteq \RKHS \qquad \mbox{for
   $\ho=1,\ldots,\Ho\!-\!1$.}
\end{align}
From the definition~\eqref{EqnDefnBellmanEval}, we see that this
condition depends on the structure of the transition distributions
$\TransOph{\ho}(\cdot \mid \state, \action)$.  In many practical
examples, the reward function itself has some number of derivatives,
and these transition distributions perform some type of smoothing, so
that we expect that the output of the Bellman update, given a suitably
differentiable function, will remain suitably differentiable.

\subsection{Stable problems have fast rates}
\label{SecStable}

We now turn the central question in understanding the behavior of any
value-based method: \emph{how to translate ``closeness'' of the
$Q$-function estimate $\Qfunhat$ to a bound on the value gap
$\valuescalar(\policystar) - \valuescalar(\policyhat)$?} At a high
level, existing theory provides guarantees of the following type: if
the $Q$-function estimates are $\varepsilon$-accurate for some
$\varepsilon \in (0, 1)$, then the value gap is bounded by a quantity
proportional to $\varepsilon$.  In contrast, our main result shows
that when the MDP is stable in a suitable sense, the value gap can be
upper bounded by a quantity proportional to $\varepsilon^2$.  This
\emph{quadratic as opposed to linear scaling} encapsulates the ``fast
rate'' phenomenon of this paper.

Our analysis isolates two key stability properties required for faster
rates; both are Lipschitz conditions with respect to a certain norm.
Here we define them with respect to the $L^2$-norm induced by the
state-action occupation measure induced by the optimal policy---namely
\begin{align}
\label{EqnDefnSAOcc}  
  \distrnorm{\funh{\,}}{\ho} \defn \sqrt{\Exp_{\occupstar} \big[
      \funh{}^2(\Stateh{\ho}, \Actionh{\ho}) \big]} \qquad \mbox{for
    any $\funh{} \in \DiffRKHS$\yd{\footnotemark},}
\end{align}
and over a neighborhood $\PlainNeigh$ of the optimal $Q$-value
function $\Qfunstar$.  
See~\Cref{AppGeneral} for the precise
definition of the neighborhood $\PlainNeigh$, as well as more general
definitions of stability that allow for different norms.

\newcommand{\ftil}{\ensuremath{\widetilde{f}}}
\footnotetext{\yd{We let $\DiffRKHS$ be the set of all difference functions of
the form $g = f - \ftil$ for some $f, \ftil$ in our base function
class $\Fclass$.}}

\paragraph{Bellman stability:}
The first condition measures the stability of the Bellman optimality
operator~\eqref{eq:def_BellOpstar}: in particular, we require that
there is a scalar $\Curvstar{\ho}$ such that
\begin{align}
  \label{eq:def_smooth_BellOpstar}
  \distrnorm[\big]{\BellOpstarh{\ho} \funh{\ho+1} - \BellOpstarh{\ho}
    \qfunstarh{\ho+1}}{\ho} \; \leq \; \Curvstar{\ho} \;
  \distrnorm[\big]{\funh{\ho+1} - \qfunstarh{\ho+1}}{\ho+1} \qquad
  \tag*{{\bf (\small{Stb}(\BellOph{}{}))}}
\end{align}
\mbox{for any $\fun \in \PlainNeigh$.}  Moreover, for any pair $(\ho,
\honew)$ of indices such that $1 \leq \ho < \honew \leq \Ho-1$, we
define
\begin{align*}
  \Curvstarfun{\ho}{\honew} \defn \Curvstar{\ho} \, \Curvstar{\ho+1}
  \ldots \Curvstar{\honew-1} \, .
\end{align*}

Condition~\ref{eq:def_smooth_BellOpstar} is directly linked to the
stability of estimating the $Q$-function $\Qfunstar$. In typical
estimation procedures, such as approximate dynamic programming, the
estimation is carried out iteratively in a backward manner, so that it
is important to control the propagation of estimation errors across
the iterations.  Condition~\ref{eq:def_smooth_BellOpstar} captures
this property, since it implies that
\begin{align*}
  \distrnorm[\big]{\BellOpstarh{\ho} \BellOpstarh{\ho+1} \ldots
    \BellOpstarh{\honew-1} \funh{\honew} - \BellOpstarh{\ho}
    \BellOpstarh{\ho+1} \ldots \BellOpstarh{\honew-1}
    \qfunstarh{\honew}}{\ho} \; \leq \; \Curvstarfun{\ho}{\honew}
  \cdot \distrnorm[\big]{\funh{\honew} - \qfunstarh{\honew}}{\honew}
  \, ,
\end{align*}
which shows how the estimation error $\big(\funh{\honew} -
\qfunstarh{\honew}\big)$ at step $\honew$ can be controlled in terms
of estimation error at an earlier time step $\ho \leq \honew$. \\

\vspace*{0.02in}

\paragraph{Occupation measure stability:}
Our second condition is more subtle, and is key in our argument.  Let
us begin with some intuition.  Consider two sequences of policies
\begin{align*}
\big( \policystarh{1}, \ldots, \policystarh{\ho-1}, \policystarh{\ho},
\policystarh{\ho+1}, \ldots, \policystarh{\honew} \big) \qquad
\mbox{and} \qquad \big( \policystarh{1}, \ldots, \policystarh{\ho-1},
\policyh{\ho}, \policystarh{\ho+1}, \ldots, \policystarh{\honew}
\big),
\end{align*}
that only differ at the $\ho$-th step, where $\policystarh{\ho}$ has
been replaced by $\policyh{\ho}$.  These two policy sequences induce
Markov chains whose distributions differ from stage $\ho$ onwards, and
our second condition controls this difference in terms of the
difference $\distrnorm{\funh{\ho} - \qfunstarh{\ho}}{\ho}$ between the
two $Q$-functions $\funh{\ho}$~and~$\qfunstarh{\ho}$ that induce
$\policyh{\ho}$ and $\policystarh{\ho}$, respectively.

We adopt $\MOpstarh{\ho}$ as a convenient shorthand for the transition
operator~$\TransOph{\ho}^{\policystar}$, and define the multi-step
transition operator \mbox{$\MOpstarhtoh{\ho}{\honew} \defn
  \MOpstarh{\ho} \, \MOpstarh{\ho+1} \cdots \MOpstarh{\honew-1}$.}
\ydsout{Moreover, we let $\DiffRKHS$ be the set of all difference functions of
the form $g = f - \ftil$ for some $f, \ftil$ in our base function
class $\Fclass$.}  Using this notation, for any $\honew \geq \ho + 1$,
we require that there is a scalar $\Curvhtoh{\ho}{\honew}$ such that
\begin{align}
  \label{eq:smooth_main}
    \sup_{ \substack{\gfunh{} \in \DiffRKHS
        \\ \distrnorm{\gfunh{}}{\honew} > 0}} \frac{ \abs[\big]{ \,
        \Exp_{\occupstar} \big[ \big(\MOpstarhtoh{\ho}{\honew} \,
          \gfunh{} \, \big) (\Stateh{\ho},
          \policystarh{\ho}(\Stateh{\ho})) -
          \big(\MOpstarhtoh{\ho}{\honew} \, \gfunh{} \, \big) (
          \Stateh{\ho}, \policyh{\ho}(\Stateh{\ho}))\big] } }
        {\distrnorm{\gfunh{}}{\honew}} & \; \leq \;
        \Curvhtoh{\ho}{\honew} \, \frac{\distrnorm[\big]{\funh{\ho} -
            \qfunstarh{\ho}}{\ho}}{\distrnorm[\big]{\qfunstarh{\ho}}{\ho}}
        \tag*{{\bf (\small{Stb}(\sdistr))}}
\end{align}
for any $\fun \in \PlainNeigh$.  The renormalization in this
definition serves to enforce a natural scale invariance; we show how
it arises naturally in~\Cref{sec:lin_fun_star}. \\

With these notions of stability in hand, we are now equipped to state
our main result.  Taking as input a value function estimate
$\Qfunhat$, it relates the induced value gap to the \emph{Bellman
residuals} $\BellOpstarh{\ho} \qfunhath{\ho+1} - \qfunhath{\ho}$.
Note that these residuals are a way of quantifying proximity to
the optimal value function $\Qfunstar$, which has Bellman residual
zero by definition.
We assume that $\Qfunhat$ has Bellman residuals bounded as
\begin{subequations}
\begin{align} 
\label{eq:def_BellErrh}
\distrnorm[\big]{ \, \BellOpstarh{\ho} \qfunhath{\ho+1} -
  \qfunhath{\ho} \, }{\ho} \; \leq \; \BellErrh{\ho} \qquad \mbox{for
  $\ho = 1, 2, \ldots, \Ho - 1$}
\end{align}
for some sequence $\bbellerr = (\BellErrh{1}, \ldots,
\BellErrh{\Ho-1}, \BellErrh{\Ho} = 0)$ that satisfies the constraint
\begin{align}
\label{EqnRegular}
\BellErrh{\ho} \geq \frac{1}{\Ho-\ho} \sum_{\honew=\ho+1}^{\Ho}
\BellErrh{\honew} \qquad \mbox{for $\ho = 1, 2, \ldots, \Ho-1$.}
\end{align}
\end{subequations}
This last condition means that the Bellman residual $\BellErrh{\ho}$
is larger than or equal to the average of the bounds established after
step $\ho + 1$.  It is natural because estimating at step $\ho$ is
at least as challenging as a stage $\honew > \ho$; indeed, any such state
$\honew$ occurs earlier in the dynamic programming backward iteration
process. As a special case, the bound~\eqref{EqnRegular} holds when
$\BellErrh{\ho} = \varepsilon$ for all stages. \\

\noindent With this set-up, we have the following guarantee in terms
of the stability coefficients $\Curvhtoh{\ho}{\honew}$ and
$\Curvstarfun{\ho}{\honew}$ from conditions~\ref{eq:smooth_main}
and~\ref{eq:def_smooth_BellOpstar}.
\begin{theorem}
\label{thm:deterministic_star}
There is a neighborhood of $\Qfunstar$ such that for any value
function estimate $\qfunhat$ with $\bbellerr$-bounded Bellman
residuals~\eqref{eq:def_BellErrh}, the induced greedy policy
$\policyhat$ has value gap bounded as
\begin{align}
\label{EqnFastRate}
\valuescalar(\policystar) - \valuescalar(\policyhat) & \leq 2 \,
\sum_{\ho=1}^{\Ho-1} \; \frac{1}{\distrnorm{\qfunstarh{\ho}}{\ho}} \,
\bigg\{ \sum_{\honew=\ho}^{\Ho-1} \, \Curvhtoh{\ho}{\honew} \;
\BellErrh{\honew} \bigg\} \bigg\{ \sum_{\honew = \ho}^{\Ho-1}
\Curvstarfun{\ho}{\honew} \; \BellErrh{\honew} \bigg\}.
\end{align}
\end{theorem} 
\vspace*{0.1in}
\noindent See~\Cref{sec:proof:thm} for the proof. \\

Treating dependence on the stability coefficients as constant, the
main take-away is that value sub-optimality is bounded above by a
quantity proportional to the \emph{squared} norm of the Bellman
residuals.  Concretely, if the Bellman residuals are uniformly upper
bounded by some $\varepsilon$, then equation~\eqref{EqnFastRate} leads
to an upper bound of the form
\begin{align}
\label{EqnQuadratic}
\valuescalar(\policystar) - \valuescalar(\policyhat) \leq c \; \Ho^3 \;
\varepsilon^2,
\end{align}
where $c$ is a universal constant.  Due to the quadratic scaling in
the Bellman residual error $\varepsilon$, this bound is substantially
tighter than the linear in $\varepsilon$ rates afforded by a
conventional analysis.  We discuss this difference in more detail in
the sequel.


\subsection{Intuition for fast rates}
\label{SecIntuition}

Why does ``fast rate'' phenomenon formalized
in~\Cref{thm:deterministic_star} arise?  In order to provide
intuition, we begin by stating a standard telescope inequality for the
value gap between two policies, and then describe the novel part of
our analysis that leads to the sharper
$\varepsilon^2$-bound~\eqref{EqnQuadratic}.  We also discuss
connections to the pessimism principle (for the off-line setting of
RL), as well as the optimism principle (for the on-line setting).

\subsubsection{Smoothness and cancelling terms in the telescope bound}
\label{SecFastIntuition}

The fast rates proved in this paper are established by a novel
argument, starting from a known telescope bound, which we begin by
stating. It controls the value gap between a given policy, and an
arbitrary comparator $\policy$.  In particular, given a $Q$-function
estimate $\qfunhat = \big(\qfunhath{1}, \ldots, \qfunhath{\Ho}\big)$,
let $\policyhat$ denote the induced greedy policy.  Then the value gap
of $\policyhat$ with respect to an arbitrary comparator policy
$\policy$ is bounded as
\begin{align}
  \label{eq:subopt_ub_old}
  \valuescalar(\policy) - \valuescalar(\policyhat) & \; \leq \;
  \sum_{\ho=1}^{\Ho-1} \big( \Exp_{\policy} - \Exp_{\policyhat} \big)
  \big[ \big( \BellOpstarh{\ho} \qfunhath{\ho+1} - \qfunhath{\ho}
    \big)(\Stateh{\ho}, \Actionh{\ho}) \big] \, .
\end{align}
This result follows by a ``telescope'' relation induced by the
structure of the Bellman updates.  Results of this type are known; for
example, analogous results can be found in past work (e.g., Theorem 2
of the paper~\cite{xie2020q}; or Lemma 3.2 in the
paper~\cite{duan2021risk}).  For completeness, we provide a proof of
the telescope bound in~\Cref{proof:subopt_ub_old}. \\

A key feature of inequality~\eqref{eq:subopt_ub_old} is the difference
of two expectations $\Exp_{\policy} - \Exp_{\policyhat}$,
corresponding to the occupation measures under $\policy$ versus
$\policyhat$.  In standard uses of this inequality, an initial
argument is used to guarantee that one of these expectations is
negative, and so can be dropped.  We describe two forms of this
argument below, either based on pessimism (\Cref{SecPess}) or optimism
(\Cref{SecOpt}). \\

In contrast, the proof of our~\Cref{thm:deterministic_star} exploits a
more refined approach, one that handles the difference of expectations
directly.  Doing so can be beneficial---and lead to ``fast rates''---
because various terms in this difference can cancel each other out.
Specifically, under the smoothness conditions that
underlie~\Cref{thm:deterministic_star}, when applying the telescope
inequality~\eqref{eq:subopt_ub_old} with comparator $\policy =
\policystar$, we show that the discrepancy between the occupation
measures associated with $\policystar$ and $\policyhat$ is of the
\emph{same order} as the Bellman residual associated with $\qfunhat$.
Note that the Bellman residuals of $\qfunhat$ already appear on the
right-hand side of inequality~\eqref{eq:subopt_ub_old}, so that this
fortuitous cancellation can be exploited---along with a number of
auxiliary results laid out in the proof---so as to upper bound the
value gap by a quantity proportional to the squared Bellman residual
$\varepsilon^2$. \\

It is worthwhile making an explicit comparison of our cancellation
approach with the more standard uses of the telescope relation, which
typically consider only one portion of the Bellman residuals
(e.g.,~\cite{jin2018q,jin2020provably,jin2021pessimism,jin2021bellman,du2021bilinear,foster2021statistical,yin2022near}).
We do so in the following two subsections.

\subsubsection{Pessimism for off-line RL}
\label{SecPess}

In the off-line instantiation of RL, the goal is to learn a ``good''
policy based on a pre-collected dataset~$\Data$.  Note that no further
interaction with the environment is permitted, hence the notion of the
learning being off-line.  More precisely, an \emph{off-line dataset}
$\Data$ of size $\numobs$ consists of quadruples
\begin{align*}
  \Data = \Big \{\big(\stateh{\ho,\, i}, \actionh{\ho, \, i},
  \statenew_{\ho, \, i}, \rewardh{\ho, \,i} \big) \Big\}_{i=1}^\numobs
  \, ,
\end{align*}
where $\stateh{\ho,\, i}$ and $\actionh{\ho, \,i}$ represent the
$i$-th state and action at the $\ho$-th step in the MDP;
$\statenewh{\ho, \, i}$ is the successive state; and $\rewardh{\ho,
  \,i} = \rewardh{\ho} (\stateh{\ho,\,i}, \actionh{\ho,\,i})$ denotes
the scalar reward.  Note that while the successive states are defined
by transition dynamics, and the rewards by the reward function, there
are no restrictions on how the state-action pairs $(\stateh{\ho, \,i},
\actionh{\ho, \, i})$ are collected.  That is, they need not have been
generated by any fixed policy, but may have collected from some
ensemble of behavioral policies, or even adaptively by human experts.
The goal of off-line reinforcement learning is to use the
$\numobs$-sample dataset $\Data$ so as to estimate a policy
$\policyhat \equiv \policyhat_\numobs$ that (approximately) maximizes
the expected return $\valuescalar(\policyhat_\numobs)$.  We expect
that---at least for a sensible method for estimating
$\policyhat_\numobs$---the value gap $\valuescalar(\policystar) \; -
\; \valuescalar(\policyhat_\numobs)$ should decay to zero as $\numobs$
increases to infinity, and we are interested in understanding this
rate of decay.

The use of pessimism is standard in off-line RL algorithms.  Its
purpose is to mitigate risks associated with ``poor coverage'' of the
off-line dataset.  For instance, the naive approach of simply
maximizing $Q$-function estimates based on an off-line dataset can
behave poorly when certain portions of the state-action space are not
well covered by the given dataset.  The pessimism principle suggests
to form a \emph{conservative estimate} of the value function---say
with
\begin{subequations}
\begin{align}
\label{EqnPessValue}  
  \qfunhath{\ho}(\state, \action) \leq \BellOpstarh{\ho}
  \qfunhath{\ho+1}(\state, \action)
\end{align}
with high probability over state-action pairs $(\state, \action)$.
Thus, the estimated value $\qfunhath{\ho}(\state, \action)$ is an
under-estimate of the Bellman update, a form of conservatism that
protects against unrealistically high estimates due to poor coverage.
Doing so in the appropriate way ensures that
\begin{align}
  \label{EqnPess}
  - \Exp_{\policyhat} \big[ \big( \BellOpstarh{\ho} \qfunhath{\ho+1} -
    \qfunhath{\ho} \big)(\Stateh{\ho}, \Actionh{\ho}) \big] & \leq 0.
\end{align}
\end{subequations}
Applying this upper bound to the inequality~\eqref{eq:subopt_ub_old}
yields the sub-optimality bound
  \begin{align*}
    \valuescalar(\policy) - \valuescalar(\policyhat) & \; \leq \;
    \sum_{\ho=1}^{\Ho-1} \Exp_{\policy} \big[ \big(
      \BellOpstarh{\ho} \qfunhath{\ho+1} - \qfunhath{\ho}
      \big)(\Stateh{\ho}, \Actionh{\ho}) \big] \, .
  \end{align*}
Upper bounds derived in this manner only contain one portion of the
Bellman residual.  When the value functions are approximated in a
parametric way (e.g., tabular problems, linear function
approximation), this line of analysis leads to value sub-optimality
decaying at a ``slow'' $1/\sqrt{\numobs}$ rate in terms of the sample
size $\numobs$ (e.g.,~\cite{jin2021pessimism}).  In contrast, an
application of~\Cref{thm:deterministic_star} can lead to value
gaps bounded by $1/\numobs$;  see~\Cref{sec:thm_imp_offline} for
details in the linear setting.


\subsubsection{Optimism in on-line RL}
\label{SecOpt}

In the setting of on-line RL, a learning agent interacts with the
environment in a sequential manner, receiving feedback in the form of
rewards based on its actions.  At the beginning, the learner possesses
no prior knowledge of the system's dynamics.  In the $t$-th episode,
the agent learns an optimal policy $\policyhat^{(t)}$ using existing
observations, implements the policy and collects data $\big\{
\big(\state_{\ho}^{(t)}, \, \action _{\ho}^{(t)}, \,
\reward_{\ho}^{(t)} \big) \big\}_{\ho=1}^{\Ho}$ from the new episode.
In each round, the system starts at an initial state $\state_1^{(t)}$
independently drawn from a fixed distribution $\sdistrinit$.

In this on-line setting, it is common to measure the performance of an
algorithm by comparing it, over the $T$ rounds of learning, with an
oracle that knows and implements an optimal policy.  At each round
$t$, we incur the \emph{instantaneous regret}
$\valuescalar(\policystar) - \valuescalar(\policyhat^{(t)})$, where
$\policystar$ is any optimal policy.  Over $T$ rounds, we measure
performance in terms of the \emph{cumulative regret}
  \begin{align}
    \label{eq:def_regret}
    \regret \big( \{\policyhat^{(t)} \}_{t=1}^T \big) & \defn
    \max_{\text{policy $\policy$}} \; \; \sum_{t=1}^T \Big \{
    \valuescalar(\policy) - \valuescalar(\policyhat^{(t)}) \Big \} \;
    = \; \sum_{t=1}^T \underbrace{\Big \{ \valuescalar(\policystar) -
      \valuescalar(\policyhat^{(t)})\Big \}}_{\mbox{Regret at round
        $t$}}.
  \end{align}
In a realistic problem, the cumulative regret of any procedure grows
with $T$, and our goal is to obtain algorithms whose regret grows as
slowly as possible.

In contrast to off-line RL, the on-line setting allows for exploring
state-action pairs that have been rarely encountered; doing so makes
sense since they might be associated with high rewards.  Principled
exploration of this type can be effected via the \emph{optimism
principle}: one constructs function estimates such that
\begin{subequations}
\begin{align}
\label{EqnOptValue}
\qfunhath{\ho}(\state, \action) \geq \BellOpstarh{\ho}
\qfunhath{\ho+1}(\state, \action)
\end{align}
with high probability over state-action pairs.\footnote{Please refer
to, for example, Lemma~B.3 in the paper~\cite{jin2020provably} for
further details.}  Note that $\qfunhath{\ho}(\state, \action)$ is
optimistic in the sense that it is an over-estimate of the Bellman
update $\BellOpstarh{\ho} \qfunhath{\ho+1}(\state, \action)$.  In this
way, we can ensure that
\begin{align}
  \label{EqnOpt}
\Exp_{\policy} \big[ \big( \BellOpstarh{\ho} \qfunhath{\ho+1} -
  \qfunhath{\ho} \big)(\Stateh{\ho}, \Actionh{\ho}) \big] & \leq 0.
\end{align}
\end{subequations}
Combining this inequality with the telescope
bound~\eqref{eq:subopt_ub_old} allows one to upper bound the regret as
\begin{align*}
\regret \big( \{\policyhat^{(t)} \}_{t=1}^T \big) \; = \; \sum_{t=1}^T
\big\{ \valuescalar(\policystar) - \valuescalar(\policyhat^{(t)})
\big\} & \; \leq \; \sum_{t=1}^T \sum_{\ho=1}^{\Ho-1}
\Exp_{\policyhat^{(t)}} \big[ \big( \qfunhath{\ho} - \BellOpstarh{\ho}
  \qfunhath{\ho+1} \big)(\Stateh{\ho}, \Actionh{\ho}) \big] \, .
\end{align*}
which only includes a single portion of the Bellman residual. In the
case of tabular or linear representations of the $Q$-functions, it
results in a regret rate of $\sqrt{T}$ (e.g., see the
papers~\cite{jin2018q,jin2020provably}).  In contrast, an appropriate
use of~\Cref{thm:deterministic_star} leads to regret growing only as
$\log(T)$, which corresponds to a much better guarantee.
See~\Cref{sec:online_thm} for details in the case of linear function
approximation. \\

In summary, then, the fast rates obtained in this paper are based on a
different approach than the standard pessimism or optimism principles.
Since we deal directly with the difference of expectations in the
bound~\eqref{eq:subopt_ub_old}, there is no need to nullify either of
them through the use of these principles.  However, it should be noted
that we are assuming smoothness conditions that allow us to control
this difference.  As we discuss in the sequel, such smoothness
conditions rule out certain ``hard instances'' used in past work on
lower bounds (e.g.~\cite{
  jin2018q,jin2020provably,jin2021pessimism,zanette2021provable}).



\section{Consequences for linear function approximation \yaqidone}
\label{sec:lin_fun_star}

In this section, we explore some consequences of our general theory
when applied to value-based methods using (finite-dimensional) linear
function approximation.  Notably, the geometry of the problem---having
to do with curvature conditions---plays a key role in verifying the
general stability conditions in this particular setting.

We consider a method that approximates value functions based on a
weighted linear combination of base features.  More concretely, let
$\Feature: \StateSp \times \ActionSp \rightarrow \real^\Dim$ be a
given feature map on the state-action space, and consider linear
expansions of the form
\begin{align*}
f_\bweight(\state, \action) = \inprod{\Feature(\state,
  \action)}{\bweight} \equiv \textstyle \sum_{j=1}^\Dim \weight{j}
\Feature_j(\state, \action)
\end{align*}
where $\bweight \in \real^\Dim$ is a weight vector.  We adopt the
conventional assumption that the feature mapping~$\Feature$ is
uniformly bounded, meaning that $\norm{\Feature(\state, \action)}_2 \,
\leq \, 1$ for all state-action pairs.  Defining the linear function
class \mbox{$\RKHS \defn \big \{ f_\bweight \mid \bweight \in
  \real^\Dim \big \}$,} we note that the Minkowski difference class
$\DiffRKHS$ is equal to~$\RKHS$, since we have not imposed any
constraints on $\bweight$.

In our analysis of linear approximation, we make use of the norm
$\distrnorm{\funh{\,}}{\ho} \defn \sqrt{\Exp_{\policystar} \big[
    \funh{}^2(\Stateh{\ho}, \Actionh{\ho}) \big]}$, corresponding to
$L^2$-norm under the occupation measure induced by the optimal policy
$\policystar$.  Given the linear structure, this norm has the explicit
representation
\begin{align}
  \distrnorm{\funh{\bweight}}{\ho} & \equiv \distrnorm{
    \bweight}{\CovOp{\ho}} \; \defn \; \sqrt{\bweight^{\top}
    \CovOp{\ho} \, \bweight}
\end{align}
where we have defined the covariance matrix \mbox{$\CovOp{\ho} \defn
  \Exp_{\occupstar} \big[ \, \Feature(\Stateh{\ho}, \Actionh{\ho}) \,
    \Feature(\Stateh{\ho}, \Actionh{\ho})^{\top} \big] \in \Real^{\Dim
    \times \Dim}$.}

\subsection{Curvature for linear approximation}

In analyzing methods based on linear approximation, it is natural to
consider curvature conditions of the following type.  At a given stage
$\ho$, let $\funh{\ho}$ and $\qfunstarh{\ho}$ be (respectively) a
value function estimate, and the optimal value function.  Letting
$\policyh{\ho}$ and $\policystarh{\ho}$ denote the corresponding
greedy-optimal policies, our analysis is based on curvature conditions
of the form
\begin{subequations}
  \begin{align}
    \distrnorm[\big]{\Feature(\state, \policyh {\ho}(\state)) -
      \Feature(\state, \policystarh{\ho}(\state))}{\CovOp{\ho}^{-1}}
    \ & \leq \ \MyCurveTwo{\ho}{\state} \, \sqrt{\Dim} \, \cdot \,
    \frac{\distrnorm[\big]{\funh{\ho} - \qfunstarh{\ho}}{\ho}}
         {\distrnorm{\qfunstarh{\ho}}{\ho}} \, , \tag*{{\bf (Curv1)}}
    \label{eq:curv_1} \\
    \funh{\ho}(\state, \policyh {\ho}(\state)) - \funh{\ho}(\state,
    \policystarh{\ho}(\state)) \ & \leq \ \MyCurveTwo{\ho}{\state} \,
    \sqrt{\Dim} \, \cdot \, \distrnorm{\qfunstarh{\ho}}{\ho} \, \cdot
    \, \bigg\{ \frac{\distrnorm[\big]{\funh{\ho} -
        \qfunstarh{\ho}}{\ho}} {\distrnorm{\qfunstarh{\ho}}{\ho}}
    \bigg\}^2 \, , \tag*{{\bf (Curv2)}}
    \label{eq:curv_2}
  \end{align}
\end{subequations}
where $\MyCurveTwo{\ho}{\state}$ is a state-dependent \emph{curvature
parameter}.  As shown in our analysis, by introducing the
$\sqrt{\usedim}$-factor on the right-hand side, the quantity
$\MyCurveTwo{\ho}{\state}$ can typically be chosen independent of
dimension.

As our analysis shows, these two inequalities arise naturally from a
sensitivity analysis of maximizing a linear objective function over a
constraint set defined by the feature mapping.  In particular, given a
value function estimate of the form $\funh{\ho}(\state, \action) =
\inprod{\bweight_f}{\Feature(\state, \action)}$, the induced greedy
policy $\policyh{\ho}(\state)$ satisfies the relation
\begin{align}
\inprod{\bweight_f}{\Feature(\state, \policyh{\ho}(\state))} & =
\max_{\action \in \ActionSp} \inprod{\bweight_f}{\Feature(\state,
  \action)} \; = \; \max_{u \in \Featureset(\state)}
\inprod{\bweight_f}{u},
\end{align}
where we have defined the constraint set $\Featureset(\state) = \{
\Feature(\state, \action) \mid \action \in \ActionSp \}$.  When this
constraint set exhibits sufficient curvature, as captured by the
quantity $\MyCurveTwo{\ho}{\state}$, the change in optimizers grows
linearly with the perturbation (inequality~\ref{eq:curv_1}), while the
corresponding function values grow quadratically
(inequality~\ref{eq:curv_2}).
\noindent In~\Cref{SecGenLinear}, we state and prove a general result
(\Cref{PropCurve}) that makes this intuition very precise. \\

\noindent Here let us consider a simple but concrete example that
illustrates the linear-quadratic behavior.
\myexample{\emph{An illustration of curvature
  property}}{ExaCurve}{Consider the state and action spaces
  \begin{align*}
    \StateSp \defn \big\{ (\state_1, \state_2) \in \Real^2 \bigm|
    \state_1^2 + \state_2^2 \leq \tfrac{1}{4} \big\} \qquad \mbox{and}
    \qquad \ActionSp \defn \big\{ (\action_1, \action_2) \in \Real^2
    \bigm| \action_1^2 + \action_2^2 \leq \exprad^2 \big\} \mbox{~with
      $\exprad \leq \tfrac{1}{2}$},
  \end{align*}
along with the feature mapping $\Feature(\state, \action) \defn \state
+ \action \in \Real^2$.  With these definitions, the constraint set
$\Featureset(\state) = \{ \state + \action \in \Real^d \mid \action
\in \ActionSp \}$ forms a disk in $\Real^2$ centered at $\state$ with
a radius of $\exprad$.  Note that by definition of the feature
mapping, we have
\begin{align}
\label{EqnBasic}  
\distrnorm[\big]{\Feature(\state, \policyh{}(\state)) -
  \Feature(\state, \policystarh{}(\state))}{2} & = \distrnorm[\big]{
  \policyh {}(\state) - \policystarh{}(\state) }{2}
\end{align}
for any pair of policies.

\begin{figure}[!ht]
  \centering
  \begin{tabular}{ccc}
    \widgraph{0.48\linewidth}{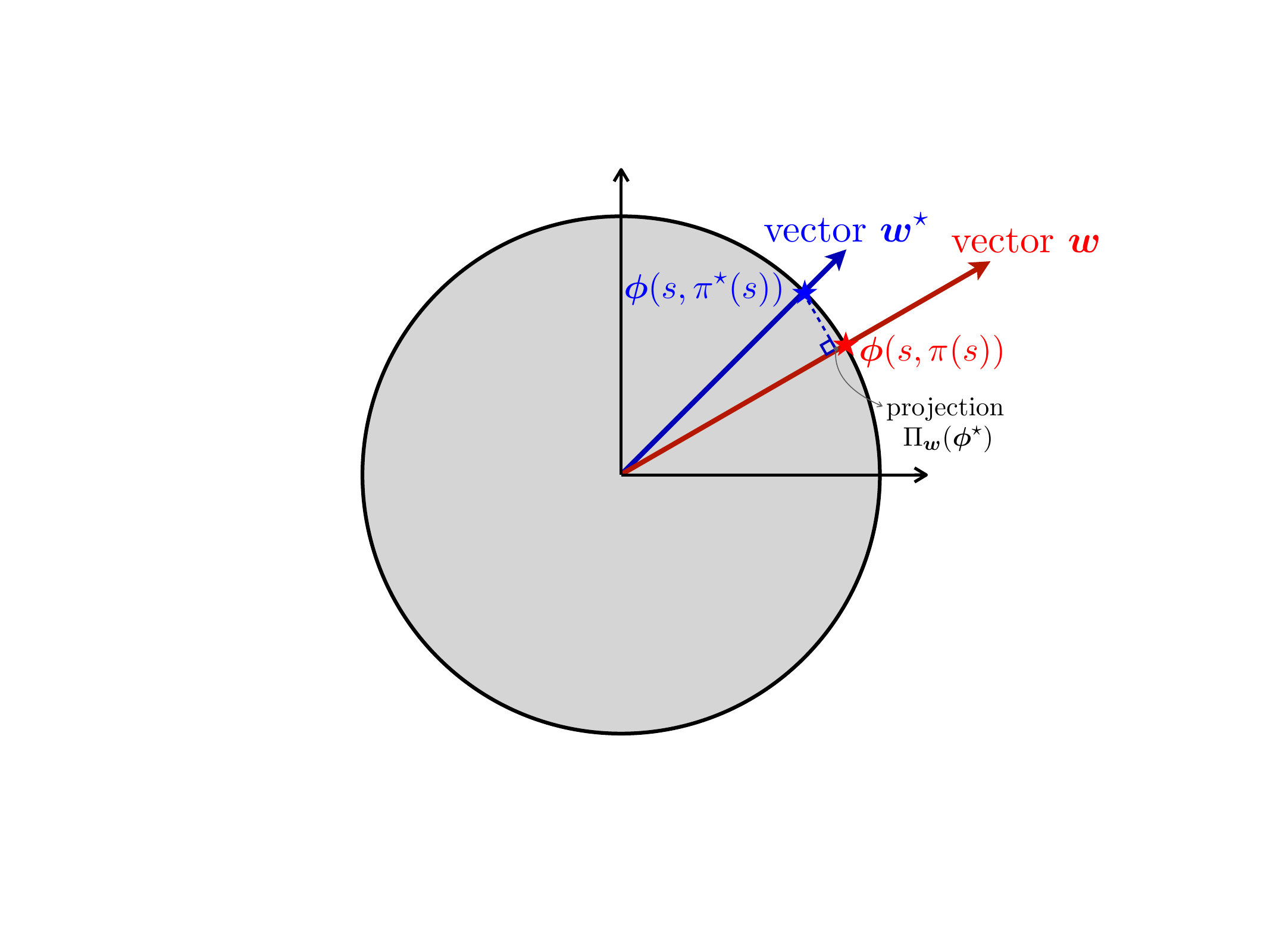}
    \hspace{-4em} & \hspace{1em} &
    \widgraph{0.36\linewidth}{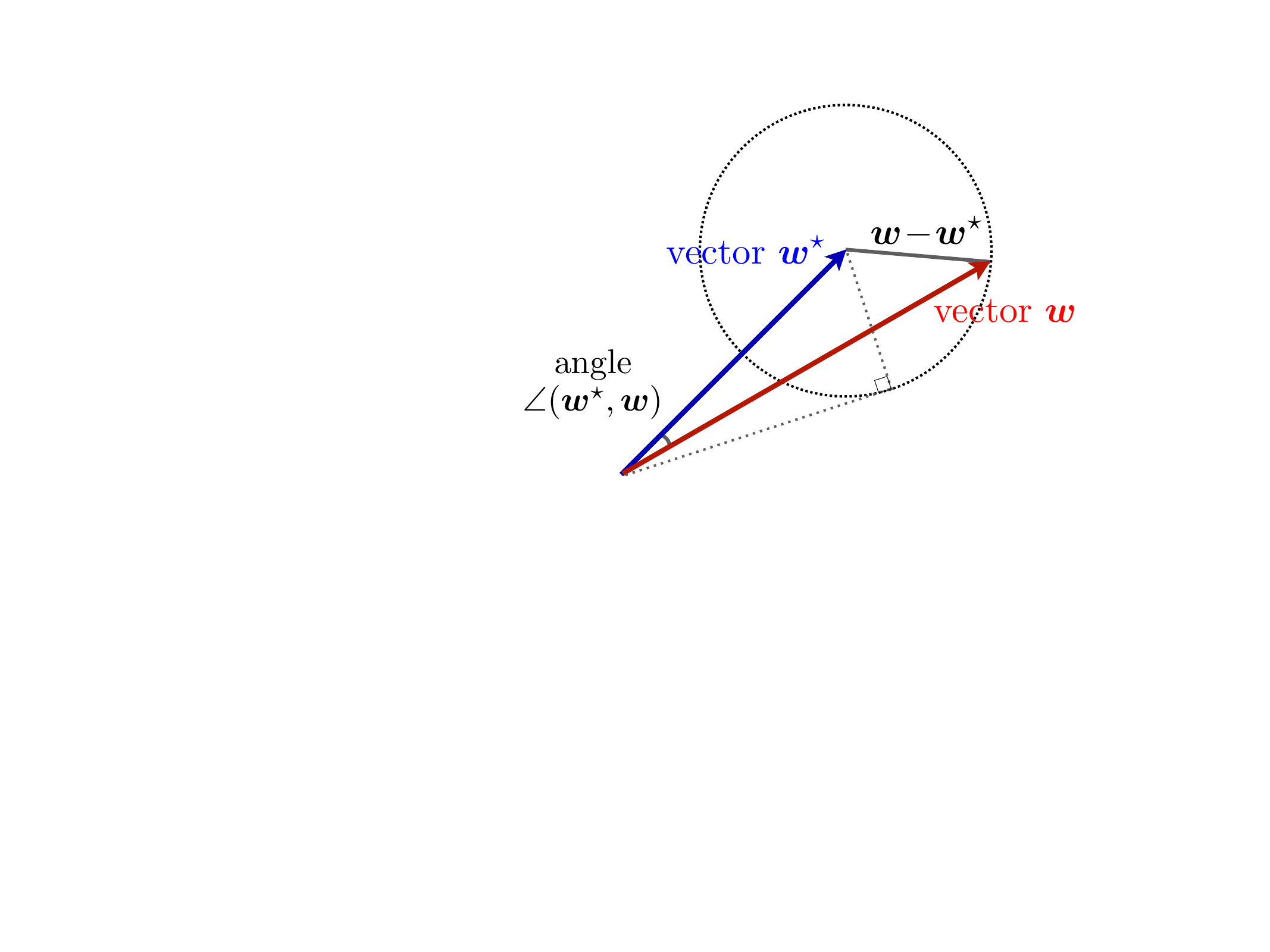}
    \\ (a) \hspace{2em} && \hspace{2.52em} (b)
  \end{tabular}
  \caption{An example with feature mapping $\Feature$ defined in
    $\Real^2$. {\bf (a)} The relation between $\Feature -
    \Featurestar$ and $\bweight - \bweightstar$.  The feature vectors
    $\Featurestar \equiv \Feature(\state, \policystarh{}(\state))$ and $\Feature \equiv \Feature(\state, \policyh{}(\state))$ at the greedy policies
    $\policystarh{}$ and $\policyh{}$ are marked by stars.  The figure
    shows that the Euclidean norm of the deviation $\Feature -
    \Featurestar$ is approximately $\exprad \; \angle(\bweightstar,
    \bweight)$.  Furthermore, when measured along the direction of
    $\bweight$, the deviation $\proj_{\bweight}(\Feature -
    \Featurestar)$ is rather small and, in fact, is of second order
    with respect to the angle $\angle(\bweightstar, \bweight)$.  {\bf
      (b)}~The~relation between the difference in vectors $\bweight -
    \bweightstar$ and the angle~$\angle( \bweightstar, \bweight)$. A
    key observation is that $\angle(\bweightstar, \bweight) \leq
    \arcsin \{\norm{\bweight - \bweightstar}_2 /
    \norm{\bweightstar}_2\}$.}
    \label{fig:lin_fun}
  \end{figure}

Suppose that the optimal $Q$-function is given by
$\qfunstarh{}(\state, \action) =
\inprod{\bweightstar}{\Feature(\state, \action)}$ for some weight
vector $\bweightstar \in \Real^2$.  Writing this weight vector as
$\bweightstar = \distrnorm{\bweightstar}{2} \, (\cos\thetastar, \,
\sin\thetastar)$ for some angle $\thetastar \in [0, 2\pi)$, the
  optimal policy takes the form $\policystarh{}(\state) =
  \arg\max_{\action \in \ActionSp} \inprod{\bweightstar}{\state +
    \action} = \arg\max_{\action \in \ActionSp}
  \inprod{\bweightstar}{\action} = \exprad \, (\cos\thetastar, \,
  \sin\thetastar)$.  Similarly, for a value function estimate
  $\funh{}$ defined by the $\bweight = \distrnorm{\bweight}{2} \,
  (\cos \theta, \, \sin\theta)$, we can write
  \mbox{$\policyh{}(\state) = \exprad \, (\cos\theta, \,
    \sin\theta)$.}  Combining with the
  representation~\eqref{EqnBasic}, we find that
\begin{subequations}
  \begin{align}
    \distrnorm[\big]{\Feature(\state, \policyh{}(\state)) -
      \Feature(\state, \policystarh{}(\state))}{2} & = \exprad \,
    \distrnorm[\big]{( \cos\theta - \cos\thetastar, \, \sin\theta -
      \sin\thetastar)}{2} \leq \exprad \; \angle(\bweight, \,
    \bweightstar) \, ,
    \label{eq:exp_curv0_1}
  \end{align}
  where the angle $\angle(\bweight, \, \bweightstar)$ is defined as
  $\angle(\bweight, \, \bweightstar) = \abs{ \theta - \thetastar}$.
  Furthermore, the difference in function values satisfies
  \begin{align}
    \abs[\big]{\funh{}(\state, \policyh {}(\state)) - \funh{}(\state,
      \policystarh{}(\state))} & = \abs[\big]{ \big\{ \Feature(\state,
      \policyh {}(\state)) - \Feature(\state, \policystarh{}(\state))
      \big\}\!\,^{\top} \bweight \, } \notag \\
    & = \abs[\big]{ \{ \policyh {}(\state) - \policystarh{}(\state)
      \}^{\top} \bweight \, } \notag \\
    & = \distrnorm{\bweight}{2} \; \distrnorm[\big]{
      \policyh{}(\state) - \proj_{\bweight} ( \policystarh{}(\state)
      )}{2} \notag \\
 & = \distrnorm{\bweight}{2} \, \cdot \, \exprad \, \big\{ 1 - \cos
    \angle(\bweight, \, \bweightstar) \big\} \leq \frac{1}{2} \,
    \exprad \, \distrnorm{\bweight}{2} \; \{ \angle(\bweight, \,
    \bweightstar) \}^2 \, .
    \label{eq:exp_curv0_2}
  \end{align}
\end{subequations}
See~\Cref{fig:lin_fun}(a) for an illustration of
inequalities~\eqref{eq:exp_curv0_1} and~\eqref{eq:exp_curv0_2}.

In order to establish curvature conditions, we need to relate the
angle $\angle (\bweight, \, \bweightstar)$ to the difference in
vectors $\bweight - \bweightstar$.  As shown in~\Cref{fig:lin_fun}(b),
when $\distrnorm{\bweight - \bweightstar} {2} \leq
\distrnorm{\bweightstar}{2}$, we have the bounds $\angle(\bweight, \,
\bweightstar) \; \leq \; \arcsin \frac{\distrnorm{\bweight -
    \bweightstar}{2}} {\distrnorm{\bweightstar}{2}} \; \leq \; \frac{2
  \, \distrnorm{\bweight - \bweightstar}{2}}
            {\distrnorm{\bweightstar}{2}}$.  These facts can be used
            to show that conditions~\ref{eq:curv_1}
            and~\ref{eq:curv_2} hold with parameter
            $\MyCurveTwo{\ho}{\state} \defn 16\sqrt{2} \; \exprad$.
            See~\Cref{AppExample} for details.}

Example~\ref{ExaCurve} captures the geometric intuition that underlies
a much broader class of examples for which the curvature conditions
hold.  In particular, suppose that the constraint set
\mbox{$\FeatureSet(\state) = \{ \Feature(\state, \action) \mid \action
  \in \ActionSp \}$} can be defined by inequalities of the form
\begin{align*}
  g_j(\Feature(\state, \action)) \leq 0 \qquad \mbox{for $j = 1,
    \ldots, M$}
\end{align*}
where each $g_j: \Real^\Dim \rightarrow \real$ is a strongly convex
function.  Note that Example~\ref{ExaCurve} provides an instance of
this set-up with a single constraint ($M = 1$), namely
$g_1(\Feature(\state, \action)) \defn \|\Feature(\state, \action) -
\state\|_2^2 - \rho^2$.  In general, the strong convexity conditions
on the constraint functions $\{g_j\}_{j=1}^M$ allow one to prove that
the curvature conditions~\ref{eq:curv_1}~and~\ref{eq:curv_2} hold.  We
refer the reader to~\Cref{PropCurve} in~\Cref{SecPropCurve} for a
complete justification.


\subsection{From curvature to fast rates}
\label{SecCurv2Fast}

Thus far, we have defined some curvature properties, and argued that
they are satisfied when the feature set has a suitable geometry.  We
now turn to the consequences of these curvature conditions for fast
rates.  Our result applies to a value function estimate $\qfunhat$,
based on $\Dim$-dimensional linear approximation over $\Ho$ stages,
whose residuals can be controlled in terms of a \emph{regular} sequence
$\bbellerr = (\BellErrh{1}, \ldots, \BellErrh{\Ho-1},
\BellErrh{\Ho}=0)$ such that
\begin{align}
\label{eq:cor_n_cond}
    \BellErrh{\ho} \, \leq \,
    \frac{\distrnorm{\qfunstarh{\ho+1}}{\ho+1}} {6 \sqrt{\Dim} \;
      \CurveNorm{\MyCurve{\ho+1}}{} \, (\Ho-\ho)^2 (1 + \log \Ho)}
    \qquad \mbox{for $\ho \in [\Ho-1]$},
\end{align}
where $\CurveNorm{\MyCurve{\ho}}{} \defn \sqrt{\Exp_{\sdistrinit,
    \policystar}\big[\MyCurveTwosq{\ho}{\Stateh{\ho}} \big]}$.

\begin{proposition}
\label{PropLin}
Consider a value function estimate $\qfunhat$ that has
$\bbellerr$-bounded Bellman residuals~\eqref{eq:def_BellErrh} for a
regular sequence $\bbellerr$ satisfying
condition~\eqref{eq:cor_n_cond}.  Then the value sub-optimality is at
most
\begin{align}
\label{EqnPropLin}
\valuescalar(\policystar) - \valuescalar(\policyhat) & \; \leq \; 6
\sqrt{\Dim} \, \sum_{\ho=1}^{\Ho-1}
\frac{\CurveNorm{\MyCurve{\ho}}{}}{\distrnorm{\qfunstarh{\ho}}{\ho}}
\, \bigg\{ \sum_{\honew=\ho}^{\Ho-1} \, \BellErrh{\honew} \bigg\}^2.
\end{align}
\end{proposition}
\noindent The proof of this result involves a number of steps.  We
provide all the details in~\Cref{sec:proof:prop:lin}, but let us
isolate a key auxiliary result that underlies the argument. \\

We require the general set-up for stability given
in~\Cref{AppGeneral}.  It allows for a pseudo-metric~$\metrich{\ho}$
that is compatible with the norm
$\|\cdot\|_\ho$ in the sense of definition~\eqref{EqnPseudoCondition}.
In the linear setting under consideration here, the nature of the
curvature condition~\ref{eq:curv_1} suggests a natural choice for this
metric, namely
\begin{align}
  \label{eq:def_metric_lin}
  \metrich{\ho}(\funh{}, \funnewh{}) \ \defn \ \frac{\sqrt{\Dim} \;
    \CurveNorm{\MyCurve{\ho}}{\ho}} {\distrnorm{\qfunstarh{\ho}}{\ho}}
  \, \cdot \, \distrnorm{\funh{} - \funnewh{}}{\ho} \qquad \mbox{for
    any $\funh{}, \funnewh{} \in \RKHS$}.
\end{align}
As stated below in~\Cref{lemma:lin_cond}, this metric satisfies the
condition~\eqref{EqnPseudoCondition} required in our analysis.
Moreover, we can use it to connect the curvature properties to the
stability conditions required for
applying~\Cref{thm:deterministic_star}.

More precisely, the following auxiliary result plays a key role in the
proof of~\Cref{PropLin}:
\begin{lemma}
  \label{lemma:lin_cond}
  \begin{enumerate}
  \item[(a)] The metric $\metrich{\ho}$ given in
    equation~\eqref{eq:def_metric_lin} is well-defined and satisfies
    the bound~\eqref{EqnPseudoCondition}.
\item[(b)] Consider any neighborhood~\eqref{EqnRhoGood} $\Neigh$ with
  radius parameters bounded as
\begin{align*}
  \radius_{\ho} \leq \frac{1}{2} \, (\Ho-\ho+1)^{-1} (1 + \log
  \Ho)^{-1} \qquad \mbox{for $\ho = 2, 3, \ldots, \Ho-1$.}
\end{align*}
Then the stability condition~\ref{eq:def_smooth_BellOpstar} holds with
$\Curvstarfun{\ho}{\honew} \leq 3$ \mbox{for all pairs $\ho \leq
  \honew$.}
\item[(c)] The stability condition~\ref{eq:smooth_main} holds with
  $\Curvhtoh{\ho}{\honew} \leq \sqrt{\Dim} \;
  \CurveNorm{\MyCurve{\ho}}{\ho}$ for all pairs $\ho \leq \honew$.
  \end{enumerate}
\end{lemma}
\noindent See~\Cref{sec:proof:lemma:lin_cond} for the proof
of~\Cref{lemma:lin_cond}, and see~\Cref{sec:proof:prop:lin} for the
proof of~\Cref{PropLin}.


\subsection{Consequences for off-line RL \yaqidone}
\label{sec:thm_imp_offline}

We now turn to some implications of~\Cref{PropLin} for off-line
reinforcement learning.  Let us recall the off-line setting: for each
$\ho = 1, \ldots, \Ho-1$, we are given a dataset $\Data_\ho =
\{(\stateh{\ho, i}, \actionh{\ho, i}, \statenew_{\ho, i},
\rewardh{\ho, i})\}_{i=1}^\numobs$ of quadruples, from which we can
compute estimates $\qfunhat = (\qfunhath{\ho})_{\ho=1}^\Ho$ with
certain Bellman residuals~$\{\BellErrh{\ho}\}_{\ho=1}^{\Ho-1}$, which
then appear in the bound~\eqref{EqnPropLin}.  The remaining factors on
the right-hand side of inequality~\eqref{EqnPropLin}, including the
term $\CurveNorm{\MyCurve{\ho}}{\ho} \, / \,
\distrnorm{\qfunstarh{\ho}}{\ho}$ along with the dimension $\Dim$, do
not depend on the dataset itself (but rather on structural properties
of the MDP).  Consequently, in terms of statistical understanding, the
main challenge is to establish high-probability bounds on the Bellman
residuals $\{\BellErrh{\ho}\}_{\ho=1}^{\Ho-1}$ for a particular
estimator.

\subsubsection{Fitted Q-iteration (FQI)}

As an illustration, let us analyze the use of \emph{fitted
Q-iteration} (FQI) for computing estimates of the $Q$-function.  At a
given stage $\ho = 1, \ldots, \Ho -1$, we can use the associated data
$\Data_\ho$ to define a regularized objective function
\begin{align}
  \label{eq:ridge}
  \Loss_\ho\big(f, \, g\big) \defn \frac{1}{\abs{\Data_{\ho}}} \left [
    \sum_{(\stateh{\ho, i}, \actionh{\ho,i}, \statenewh{\ho,i},
      \reward_{\ho, i}) \in \Data_{\ho}} \big\{ f(\stateh{\ho, \, i},
    \, \actionh{\ho, \, i}) - \big (\reward_{\ho, \, i} +
    \max_{\action \in \ActionSp} g(\statenewh{\ho, \, i}, \action)
    \big) \big\}^2 \, \right]\ + \; \regularh{\ho}^2(f) \, .
\end{align}
Here $g$ represents the target function from stage $\ho + 1$, and it
defines the targeted responses $y_{\ho, i}(g) \defn \reward_{\ho,i} +
\max_{\action \in \ActionSp} g(\statenewh{\ho, i}, \action)$.  For a
given target $g$, we obtain a $Q$-function estimate for stage~$h$ by
minimizing the functional $f \mapsto \Loss_\ho(f, g)$.  Given that our
objective is defined with a quadratic cost, doing so can be understood
as a regression method for estimating the conditional expectation that
underlies the Bellman update---viz.
\begin{align}
\BellOpstarh{\ho} g(\state, \action) = \Exp[ \, y_{\ho, \, i}(g) \mid
  (\stateh{\ho, \, i}, \, \actionh{\ho, \, i}) = (\state, \action) \;
].
\end{align}
The additional quantity $\regularh{\ho}^2(f)$ in our
definition~\eqref{eq:ridge} is a regularizer.  Given this set-up, we
can generate a $Q$-function estimate $\qfunhat = (\qfunhath{1},
\ldots, \qfunhath{\Ho})$ by first initializing $\qfunhath{\Ho} =
\rewardh{\Ho}$, and then recursively computing
\begin{align}
\qfunhath{\ho} = \arg \min_{f \in \Fclass} \Loss_\ho\big(f, \,
\qfunhath{\ho+1} \big), \qquad \mbox{for $\ho = \Ho-1, \Ho-2, \ldots,
  2, 1$.}
\end{align}

\paragraph{Ridge penalty:}
It remains to define the choice of regularizer.  For the linear
functions \mbox{$\funh{} = \inprod{\Feature(\cdot)}{\bweight}$} under
consideration, a standard choice is the ridge penalty
\mbox{$\regularh{\ho}^2(\funh{}) \defn \ridgeh{\ho} \,
  \distrnorm{\bweight}{2}^2$}\,, where $\ridgeh{\ho} \geq 0$ is the
regularization weight.  In the analysis here, we assume that the
dataset consists of i.i.d. tuples (but this can be relaxed as needed).
Concretely, the dataset $\Data_{\ho}$ at stage $\ho$ consists of
$\numobs$ quadruples $\big\{(\stateh{\ho, \, i}, \actionh{\ho, \, i},
\rewardh{\ho, \, i}, \statenewh{\ho, \, i}) \big\}_{i=1}^{\numobs}$,
where the state-action pairs $\{ (\stateh{\ho, \, i}, \, \actionh{\ho,
  \, i}) \}_{i=1}^{\numobs}$ are drawn i.i.d. from a behavioral
distribution \mbox{$\distrdatah{\ho}$} over state-action pairs
$\StateSp \times \ActionSp$.  This data-generating distribution
$\distrdatah{\ho}$ may differ from the occupation measure associated
with the optimal policy $\policystar$---that is, $\distrstarh{\ho} =
\Prob_{\sdistrinit, \policystar}[ \, (\Stateh{\ho}, \Actionh{\ho}) \in
  \cdot \, ]$.  This discrepancy leads to form of \emph{covariate
shift} in the regression steps that underlie the FQI procedure.

\subsubsection{Fast rates for FQI-based estimates}
\label{SecFQIFast}
We now state a corollary of~\Cref{PropLin}, applicable to value
function estimates based on FQI with ridge regression.  Our result
involves the $\Dim$-dimensional empirical covariance matrices
\begin{align}
  \label{eq:def_CovOpdata}
  \CovOpdata{\ho} & \defn \frac{1}{\abs{\Data_{\ho}}}
  \sum_{\Data_{\ho}} \, \Feature(\stateh{ \ho, \, i}, \, \actionh{\ho,
    \, i}) \, \Feature(\stateh{\ho, \, i}, \, \actionh{\ho, \,
    i})^{\top} \in \Real^{\Dim \times \Dim} \, ,
\end{align}
which we assume to be well-conditioned, with a lower bound on the
smallest eigenvalue \mbox{$\eighat_{\min} \geq \const{0} \,
  \Dim^{-1}$}.  Additionally, we define the empirical conditional
variances
\begin{align}
  \label{eq:def_stderrdata}
  \stderrhatdatah{\ho}^2(\funh{\,}) \defn
  \frac{1}{\abs{\Data_{\ho}}}\sum_{\Data_{\ho}} \Var\big[
    \max_{\action \in \ActionSp}\funh{}(\Stateh{\ho+1},
    \action) \bigm| \Stateh{\ho} = \stateh{\ho, \, i}, \, \Actionh{\ho} = \actionh{\ho, \, i} \,
    \big] \qquad \mbox{for any $\funh{} \in \Real^{\StateSp \times
      \ActionSp}$}.
\end{align}

\begin{corollary}[Fast rates for ridge-based FQI]
\label{CorOffRidge}  
For FQI based on ridge regression, with a sufficiently large sample
size $\numobs$ and with suitable choices of the regularization
parameters $ \{ \ridgeh{\ho} \}_{\ho = 1}^{\Ho-1}$, the
bound~\eqref{EqnPropLin} from~\Cref{PropLin} holds with
\begin{align}
\label{EqnOffRidge}
\BellErrh{\ho} & = c \norm[\big]{ \CovOp{\ho} ^{\frac{1}{2}} \, (
  \CovOpdata{\ho} + \ridgeh{\ho} \IdMt )^{-\frac{1}{2}}}_2 \;
\stderrhatdatah{\ho}\big(\qfunhath{\ho+1} \big) \sqrt{\frac{\Dim \;
    \log(\Dim/\errprob)}{\numobs}}
  \end{align}
with probability at least $1 - \errprob$.
\end{corollary}
\noindent See~\Cref{SecProofCorOffRidge} for the proof of this
claim. \\

\paragraph{Fast rates and comparisons to past work:}
So as to be able to compare with results from past work, let us
consider some consequences of the bound~\eqref{EqnOffRidge} under
standard assumptions.  Suppose that the rewards take values in the
unit interval $[0, 1]$, and the covariate shift (discussed at more
length below) is mild, in the sense that we view the term $\big\{
\norm[\big]{ \CovOp{\ho}^{\frac{1}{2}} \, ( \CovOpdata{\ho} +
  \ridgeh{\ho} \IdMt )^{-\frac{1}{2}}}_2 \big\}_{\ho=1}^{\Ho-1}$ as
constant order. Furthermore, we treat the curvature terms $\{
\CurveNorm{\MyCurve{\ho}}{\ho} \}_{\ho=1}^{\Ho-1}$ as constants.  Under
these conditions, it can be shown that the bound
from~\Cref{CorOffRidge} takes the form
\begin{subequations}
\begin{align}
\label{eq:sub_opt_offline_order}
  \valuescalar(\policystar) - \valuescalar(\policyhat) \; \leq \;
  \PlainCon \, \frac{\Dim^{3/2}\; \Ho^3}{\numobs} \, \log (\Dim
  \Ho/\errprob),
\end{align}
and is valid for a sample size $\numobs \geq c \Dim^2 \Ho^3$.
See~\Cref{sec:lin_offline_order_1} for the details of this
calculation.  Alternatively stated, \Cref{CorOffRidge} guarantees that
for FQI using ridge regression with $\Dim$-dimensional function
approximation, the number of samples $\numobs(\epsilon)$ required to
obtain $\epsilon$-optimal policy is at most
\begin{align}
\label{EqnFastLinear}
n_{\mbox{\scriptsize{fast}}} (\epsilon) & \asymp
\frac{\Dim^{\frac{3}{2}} \Ho^3}{\epsilon} + \Dim^2 \Ho^3 \, ,
\end{align}
where we use $\asymp$ to denote a scaling that ignores constants and
logarithmic factors.

Let us compare this guarantee to related work by Zanette et
al.~\cite{zanette2021provable}, who analyzed the use of pessimistic
actor-critic methods for linear function classes.  When translated
into the notation of our paper, their analysis
established\footnote{See~\Cref{sec:lin_offline_order_2} for the
details of this calculation} a sample complexity of the order
\begin{align}
\label{EqnZanette}
n_{\mbox{\scriptsize{Zan}}}(\epsilon) & \asymp \frac{\Dim^2
  \Ho^3}{\epsilon^2}.
\end{align}
\end{subequations}
Consequently, we see that once the target error $\epsilon$ is
relatively small---$\epsilon \in (0, 1)$---then stable MDPs can
exhibit a much smaller $(1/\epsilon)$ sample complexity.

It should be noted that past work
(e.g.,~\cite{jin2021pessimism,zanette2021provable}) has established
$(1/\epsilon^2)$-lower bounds on the sample complexity of estimating
$\epsilon$ policies in the off-line setting.  However, these lower
bounds \emph{do not} contradict our fast rate
guarantee~\eqref{EqnFastLinear}, because the ``hard instances'' used
in these lower bound proofs violate the stability
condition~\ref{eq:smooth_main}.  In particular, even infinitessimally
small perturbations in policy lead to occupation measures that are
significantly different.

\paragraph{Transfer learning and covariate shift:}
It is also worth noting that the bound~\eqref{EqnOffRidge} highlights
an important connection to covariate shift. This phenomenon arises
whenever the data is \emph{not} collected under the occupation measure
induced by the optimal policy.  More precisely, while we measure the
Bellman residual error using the $L^2$-norm under this occupation
measure $\distrstarh{\ho}$, the data are drawn from the
distribution~$\distrdatah{\ho}$, which might differ significantly from
$\distrstarh{\ho}$.  This can be viewed as a form of covariate shift
in the regression problems that underlie the FQI method.

For the linear function classes to which~\Cref{CorOffRidge} applies,
the effect of this covariate shift is measured by the term
$\norm[\big]{ \CovOp{\ho}^{\frac{1}{2}} \, ( \CovOpdata{\ho} +
  \ridgeh{\ho} \IdMt )^{-\frac{1}{2}}}_2$ in
equation~\eqref{EqnOffRidge}.  Here $\CovOp{\ho}$ is the covariance
matrix under the occupation measure, whereas $\CovOpdata{\ho}$ is the
empirical covariance defined by the dataset.  Such a condition is much
milder than any assumption directly posed on density ratios. We note
that related measures of covariate shift in off-line RL have appeared
in past work
(e.g.,~\cite{duan2020minimax,jin2021pessimism,zanette2021provable}),
but without the connections to fast rates given here.

\paragraph{When is pessimism necessary?}

An interesting aspect of the guarantee from~\Cref{CorOffRidge} is that
it provides guarantees for off-policy RL (and with fast rates) using a
method that does \emph{not} incorporate any form of pessimism.  This
is a sharp contrast with many other methods for off-policy RL, such as
pessimistic forms of $Q$-learning and actor-critic methods
(e.g.,~\cite{jin2021pessimism,zanette2021provable}).

To be clear, as noted following the
bound~\eqref{eq:sub_opt_offline_order}, the guarantee
from~\Cref{CorOffRidge} requires the sample size to be lower bounded
as $\numobs \geq c \Dim^2 \Ho^3$.  In contrast, pessimistic schemes
only require a sample size sufficiently large to ensure validity of
the Bellman residual upper bounds that
underlie~\Cref{CorOffRidge}---meaning that $\numobs \gtrsim \Dim$ up
to logarithmic factors.  Thus, the pessimism principle can be useful
for problems with smaller sample sizes.


\subsection{Consequences for on-line RL}
\label{sec:online_thm}

In this section, we explore some consequences of~\Cref{PropLin} for
on-line reinforcement learning.  We begin by describing a two-stage
procedure\footnote{To be clear, the purpose of this scheme is
primarily conceptual, rather than practical in nature.} that allows us
to convert the risk bounds for FQI from off-line RL into regret in
on-line RL:
\begin{description}
\item[\PhaseOne] (\emph{Exploration}) In the initial $\Tsafe$
  episodes, the focus is purely on exploration, resulting in an
  estimate of $Q$-function denoted as $\qfunhat^{(\Tsafe)}$.  {For
    instance, we may apply some fixed exploration policy in each
    episode, designed to ensure reasonable coverage of the data, and
    then use FQI to compute $\qfunhat^{(\Tsafe)}$.}
  \item[\PhaseTwo] (\emph{Fine-tuning}) For $k = 0, 1, \ldots,
    K\!-\!1$ with $K \! \defn \! \big\lceil \! \log_2(T /\Tsafe)
    \big\rceil$, repeat the following process:
    \begin{itemize}
      \item In the $t$-th episode, for each $t = \Tsafe \, 2^k +1,
        \ldots, \Tsafe \, 2^{k+1}$, execute the greedy policy induced
        by function~$\qfunhat^{(\Tsafe \, 2^k)}$.
        \item Update the $Q$-function estimate $\qfunhat^{(\Tsafe \,
          2^{k+1})}$ using FQI based on observations collected from
          episodes \mbox{$\Tsafe \, 2^k + 1, \Tsafe \, 2^k + 2,
            \ldots, \Tsafe \, 2^{k+1}$}.
    \end{itemize}
\end{description}
We assume the burn-in time $\Tsafe$ is large enough so as to ensure
the pilot $Q$-function estimate $\qfunhat^{(\Tsafe)}$ obtained in
\PhaseOne~falls within a certain ``absorbing'' region $\Neigh$
around~$\Qfunstar$, characterized by the following
properties: \vspace{-.5em}
\begin{itemize} \itemsep = -.1em
  \item (\emph{Absorbing property}) For any greedy policy $\policy$
    induced by a function $\Qfun \!\in\! \Neigh$, running it for at
    least $\Tsafe$ episodes and applying FQI to the observed data
    yields an estimated function $\qfunhat$ that belongs to $\Neigh$.
  \item (\emph{Bounded covariate shift}) For any function $\Qfun \in
    \Neigh$, the associated greedy policy $\policy$ is a sufficiently
    accurate approximation to $\policystar$ so as to ensure that the
    covariate shift term $\norm[\big]{ \CovOp{\ho} ^{\frac{1}{2}} \, (
      \CovOpdata{\ho} + \ridgeh{\ho} \IdMt )^{-\frac{1}{2}}}_2 $ is
    upper bounded by a constant.  (Here $\CovOp{\ho}$ is the
    covariance matrix under $\policystar$, whereas $\CovOpdata{\ho}$
    is the empirical covariance when collecting samples under the
    greedy policy $\policy$.)
\end{itemize}

\noindent 
Under these conditions, we have the following bound on the
regret~\eqref{eq:def_regret}, as previously defined.
\begin{corollary}
\label{CorOnRidge}
For FQI based on ridge regression with rewards in $[0,1]$, with a
sufficiently large burn-in time $\Tsafe$ and with suitable choices of
the regularization parameters $ \{ \ridgeh{\ho} \}_{\ho = 1}^{\Ho-1}$,
the two-phase scheme achieves regret bounded as
\begin{align}
  \label{eq:regret}
  \regret(T) \; & \leq \; \PlainCon \, \big\{ \Tsafe \cdot \Ho \; + \;
  \Dim\sqrt{\Dim} \; \Ho^4 \, \log T \, \cdot \, \log (\Dim \Ho
  K/\errprob) \big\}
\end{align}
with probability at least $1 - \errprob$.
\end{corollary}
\noindent See~\Cref{SecProofCorOnRidge} for the proof. \\

\paragraph{Sharper bound on regret:}  
The leading term (as $T$ grows) in the bound~\eqref{eq:regret} grows
as $\log T$, which is much smaller than the typical $\sqrt{T}$-rate
found in past work~\cite{jin2018q,jin2020provably}.  The $\sqrt{T}$
rate has been shown to be unimprovable in general, but the worst-case
instances\cite{jin2018q,jin2020provably} that lead to
$\sqrt{T}$-regret violate the stability conditions used in our
analysis.

\paragraph{When is optimism needed?}
The use of optimism---by adding bonuses to the current value function
estimates so as to encourage exploration---underlies many schemes in
on-line RL.  An interesting take-away from~\Cref{CorOnRidge} is that
under the stability conditions highlighted by our theory, it is
possible to achieve excellent regret bounds without the use of
optimism.  In our two-phase scheme, the only exploration occurs in
\PhaseOne.  All other data is simply collected using the greedy policy
induced by the current $Q$-function estimate.  A well-designed
exploration scheme---one that might incorporate the optimism
principle---is necessary only during the \mbox{burn-in \PhaseOne.}

There are degenerate settings in which additional exploration might be
required.  For example, consider the degenerate situation in which the
optimal policy $\policystar$ leads to a (nearly) rank-deficient
covariance matrix $\CovOp{\ho}$. In such cases, executing greedy
policies in a neighborhood of~$\policystar$ might fail \yd{to} generate
observations that sufficiently represent the underlying dynamics,
thereby hindering efficient estimation.  This observation has
parallels with results on contextual bandits, where exploration-free
algorithms are known to be efficient under a covariate diversity
condition \cite{bastani2021mostly}. Exploration becomes necessary when
this assumption is not satisfied.


\section{Proofs}
\label{SecProofs}

This section is devoted to the proofs of~\Cref{thm:deterministic_star}
and~\Cref{PropLin}.  In both cases, we break down the proofs into a
number of auxiliary claims, and defer the proofs of these more
technical results to the appendices, as indicated.  All of our proofs
make use of the more general stability framework described
in~\Cref{AppGeneral}.


\subsection{Proof of Theorem~\ref{thm:deterministic_star}}
\label{sec:proof:thm}

We begin with the proof of~\Cref{thm:deterministic_star}, which
consists of three main steps.  These steps rely on two auxiliary
lemmas whose proofs are fairly technical, so that they are deferred to
in~\Cref{proof:subopt_ub,sec:proof:lemma:smooth}.

\paragraph{High-level outline:}
Let us outline the three steps of the proof.  In Step 1, we use a
one-step expansion of the difference in the occupation measures to
reformulate the standard telescope
inequality~\eqref{eq:subopt_ub_old}.  Doing so results in a relation
with structure similar to that of the left-hand side of
inequality~\ref{eq:smooth_main}.  In Step 2, we develop a constraint
on the function estimation error $\metrich{\ho}\big(\qfunhath{\ho},
\qfunstarh{\ho}\big)$ that ensures the occupation measure produced by
policy $\policyhat$ remains stable and does not deviate too much from
the occupation measure associated with the optimal policy
$\policystar$.  In Step 3, we use Bellman stability
\ref{eq:def_smooth_BellOpstar} to connect the $Q$-function error
$\qfunhath{\ho} - \qfunstarh{\ho}$ with Bellman residuals.  With this
high-level view in place, we now work through the three steps.


\subsubsection{Step 1: Reformulation of the telescope inequality.}

Recall the standard telescope inequality~\eqref{eq:subopt_ub_old}.
Our proof makes use of an alternative form, which involves the
functions
\begin{align}
\label{eq:def_Deltafun}  
\Deltafunh{\ho}(\policy; \, \state, \action \, ) \; = \;
\myssum{\honew=\ho}{\Ho-1} \; \MOppihtoh{\ho}{\honew}{\policy} \big(
\BellOpstarh{\honew} \qfunhath{\honew+1} - \qfunhath{\honew}
\big)(\state, \action).
\end{align}

\begin{lemma}
\label{lemma:telescope}
Given a $Q$-function estimate $\qfunhat = \big( \qfunhath{1}, \ldots,
\qfunhath{\Ho-1}, \qfunhath{\Ho} = \rewardh{\Ho} \big)$ and the
associated greedy policy $\policyhat$, we have the bound
  \begin{align}
    \label{eq:subopt_ub}
    \valuescalar(\policy) - \valuescalar(\policyhat) & \; \leq \;
    \myssum{\ho=1}{\Ho-1} \Exp_{\policyhat} \big[
      \Deltafunh{\ho}(\policy; \, \stateh{\ho},
      \policyh{\ho}(\stateh{\ho})) - \Deltafunh{\ho}(\policy; \,
      \stateh{\ho}, \policyhath{\ho}(\stateh{\ho})) \big]
  \end{align}
valid for any policy $\policy$.
\end{lemma}
\noindent See~\Cref{proof:subopt_ub} for the proof. \\

We apply the bound~\eqref{eq:subopt_ub} with $\policy =
\policystar$. Following some algebra, we find that
\begin{align*}
  \valuescalar(\policystar) - \valuescalar(\policyhat) & \ \leq
  \ \sum_{\ho=1}^{\Ho-1} \sum_{\honew=\ho}^{\Ho-1} \ \DiffOphat(\ho,
  \, \honew) \, \cdot \, \BellErrh{\honew} \; ,
\end{align*}
where $\BellErrh{\honew}$ is an upper bound on the Bellman residual
$\distrnorm[\big]{ \BellOpstarh{ \honew} \qfunhath{\honew+1} -
  \qfunhath{\honew} }{\honew}$ as given in
equation~\eqref{eq:def_BellErrh}.  The term $\DiffOphat(\ho, \,
\honew)$ is given by
\begin{subequations}
\begin{align}
  \label{eq:def_DiffOphat}
  \DiffOphat(\ho, \, \honew) & \defn \sup_{\funh{} \in \DiffRKHS: \,
    \distrnorm{\funh{}}{\honew} > 0} \; \left \{
  \frac{1}{\distrnorm{\funh{}}{\honew}} \abs[\Big]{\Exp_{\policyhat}
    \Big[ \big( \MOpstarhtoh{\ho}{\honew} \, \funh{\,}
      \big)(\stateh{\ho}, \policystarh{\ho} (\stateh{\ho})) - \big(
      \MOpstarhtoh{\ho}{\honew} \, \funh{\,} \big) (\stateh{\ho},
      \policyhath{\ho}(\stateh{\ho})) \Big]} \right \}.
\end{align}
We note that the left-hand side of inequality~\ref{eq:smooth_main} has
a similar form to the term $\DiffOphat(\ho, \, \honew)$, differing
only in that the expectation is taken over the occupation measure of
running the optimal policy $\policystar$, rather than the estimated
policy $\policyhat$.


\subsubsection{Step 2: Constraint to ensure stability}
\label{sec:proof:thm_outline_StepTwo}

Our next step is to establish an upper bound on the coefficient
$\DiffOphat(\ho, \honew)$ defined by the estimated policy $\policyhat$
in terms of the analogous quantity defined by the optimal policy
$\policystar$---namely, the coefficient
\begin{align}
\label{eq:def_DiffOp}
\DiffOp(\ho, \honew) \defn \!  \sup_{\funh{} \in \DiffRKHS: \,
  \distrnorm{\funh{}}{\honew} > 0} \; \left \{
\frac{1}{\distrnorm{\funh{}}{\honew}} \abs[\Big]{\Exp_{\policystar}
  \Big[ \big( \MOpstarhtoh{\ho}{\honew} \, \funh{\,}
    \big)(\stateh{\ho}, \policystarh{\ho} (\stateh{\ho})) - \big(
    \MOpstarhtoh{\ho}{\honew} \, \funh{\,} \big) (\stateh{\ho},
    \policyhath{\ho}(\stateh{\ho})) \Big]} \right \} .
\end{align}
\end{subequations}
In order to do so, we demonstrate that a sufficiently small function
estimation error $\metrich{\ho}\big( \qfunhath{\ho}, \qfunstarh{\ho}
\big)$
ensures the inequality
\begin{align}
  \label{eq:def_Ttilfast_new}
  \sum_{\ho=1}^{\Ho-1} \sum_{\honew=\ho}^{\Ho-1} \ \DiffOphat(\ho, \,
  \honew) \cdot \BellErrh{\honew} \ \leq \ 2 \; \sum_{\ho=1}^{\Ho-1}
  \sum_{\honew=\ho}^{\Ho-1} \ \DiffOp(\ho, \, \honew) \cdot
  \BellErrh{\honew} \, .
\end{align}
Once we have established this bound, we can replace the term
$\DiffOp(\ho, \honew)$ with \mbox{$\Curvhtoh{\ho}{\honew} \! \cdot \!
  \distrnorm[\big]{ \qfunhath{\ho} \! - \!  \qfunstarh{\ho} }{\ho}
  \big/ \distrnorm { \qfunstarh{\ho} } {\ho}$}, using the
inequality~\ref{eq:smooth_main}.

We summarize the result in the following auxiliary lemma:
\begin{lemma}
\label{lemma:smooth}
Suppose that the function estimation errors satisfy
$\metrich{\ho}\big(\qfunh{\ho}, \, \qfunstarh{\ho} \big) \leq
\frac{1}{2 \, \Radphistar} \, (\Ho-\ho+1)^{-1}$ for \mbox{$\ho = 2, 3,
  \ldots, \Ho-1$} and the sequence $\bbellerr = (\BellErrh{1}, \ldots,
\BellErrh{\Ho-1}, \BellErrh{\Ho} = 0)$ satisfies the regularity
condition~\eqref{EqnRegular}.  Then we have
\begin{align}
\label{eq:stable}
\valuescalar(\policystar) - \valuescalar(\policyhat) \; \leq \; 2 \,
\sum_{\ho=1}^{\Ho-1} \frac{\distrnorm[\big]{\qfunhath{\ho} -
    \qfunstarh{\ho} }{\ho} } { \distrnorm {\qfunstarh{\ho}}{\ho}} \;
\Big \{ \sum_{\honew = \ho}^{\Ho-1} \Curvhtoh{\ho}{\honew} \;
\BellErrh{\honew} \Big \}.
\end{align}
\end{lemma}
\noindent See~\Cref{sec:proof:lemma:smooth} for the proof.


\subsubsection{Step 3: Connecting  $Q$-function error and Bellman residuals}
\label{sec:fun_diff_BellErr}

The remaining piece of the proof is to connect the function difference
$\qfunhath{\ho} - \qfunstarh{\ho}$ with Bellman residuals
$\BellOpstarh{\ho} \qfunhath{\ho+1} - \qfunhath{\ho}$, using the
stability condition~\ref{eq:def_smooth_BellOpstar} on the Bellman
operator $\BellOpstar$.  This is relatively straightforward: indeed,
we claim that
\begin{align}
  \label{eq:fun_diff_BellErr}
  \distrnorm[\big]{\qfunhath{\ho} - \qfunstarh{\ho}}{\ho} & \; \leq \;
  \sum_{\honew = \ho}^{\Ho-1} \; \Curvstarfun{\ho}{\honew} \cdot
  \distrnorm[\big]{\BellOpstarh{\honew} \qfunhath{\honew+1} -
    \qfunhath{\honew} }{\honew} \, .
\end{align}
Recall that $\qfunstarh{\ho} = \BellOpstarh{\ho} \qfunstarh{\ho+1}$
for $\ho = 1, 2, \ldots, \Ho-1$. Therefore, we have
\begin{align*}
  \qfunhath{\ho} - \qfunstarh{\ho} & = \big( \BellOpstarh{\ho}
  \qfunhath{\ho+1} - \BellOpstarh{\ho} \qfunstarh{\ho+1} \big) -
  \big(\BellOpstarh{\ho} \qfunhath{\ho+1} - \qfunhath{\ho}\big) \, .
\end{align*}
By employing the triangle inequality and the Bellman stability given
in equation~\ref{eq:def_smooth_BellOpstar}, we derive that
\begin{align*}
  \distrnorm[\big]{ \qfunhath{\ho} - \qfunstarh{\ho}}{\ho} & \leq
  \distrnorm[\big]{ \BellOpstarh{\ho} \qfunhath{\ho+1} -
    \qfunhath{\ho}}{\ho} + \distrnorm[\big]{ \BellOpstarh{\ho}
    \qfunhath{\ho+1} - \BellOpstarh{\ho} \qfunstarh{\ho+1} }{\ho} \\ &
  \leq \distrnorm[\big]{ \BellOpstarh{\ho} \qfunhath{\ho+1} -
    \qfunhath{\ho} }{\ho} + \Curvstar{\ho} \,
  \distrnorm[\big]{\qfunhath{\ho+1} - \qfunstarh{\ho+1} }{\ho+1} \, .
\end{align*}
Applying this inequality recursively yields the
claim~\eqref{eq:fun_diff_BellErr}. \\

With this piece in place, we can complete the proof
of~\Cref{thm:deterministic_star}.  Indeed, we have
\begin{align*}
\valuescalar(\policystar) - \valuescalar(\policyhat) &
\stackrel{(a)}{\leq} \; 2 \, \sum_{\ho=1}^{\Ho-1}
\frac{\distrnorm[\big]{\qfunhath{\ho} - \qfunstarh{\ho} }{\ho} } {
  \distrnorm {\qfunstarh{\ho}}{\ho}} \; \Big \{ \sum_{\honew =
  \ho}^{\Ho-1} \Curvhtoh{\ho}{\honew} \; \BellErrh{\honew} \Big \} \\
& \stackrel{(b)}{\leq} 2 \, \sum_{\ho=1}^{\Ho-1} \frac{1}{\distrnorm
  {\qfunstarh{\ho}}{\ho}} \; \Big \{ \sum_{\honew = \ho}^{\Ho-1} \;
\Curvstarfun{\ho}{\honew} \; \BellErrh{\honew} \Big \} \; \Big \{
\sum_{\honew = \ho}^{\Ho-1} \Curvhtoh{\ho}{\honew} \;
\BellErrh{\honew} \Big \}.
\end{align*}
Here step (a) is a restatement of the bound~\eqref{eq:stable}
from~\Cref{lemma:smooth}, whereas step (b) follows from
inequality~\eqref{eq:fun_diff_BellErr}.  Thus, we have established the
claim given in~\Cref{thm:deterministic_star}.


\subsection{Proof of Proposition~\ref{PropLin}}
\label{sec:proof:prop:lin}

We now turn to the proof of~\Cref{PropLin}, which provides a guarantee
for value-based methods using linear function approximation.


\subsubsection{High-level overview}

There are two main ingredients in the proof: (i) the auxiliary claims
previously stated as~\Cref{lemma:lin_cond} following the statement of
the proposition; and (ii) verifying that the conditions
of~\Cref{lemma:lin_cond} hold, so that we may apply it, in conjunction
with~\Cref{thm:deterministic_star}, so as to establish the claim.  The
proof of~\Cref{lemma:lin_cond} is given
in~\Cref{sec:proof:lemma:lin_cond}, whereas we prove step (ii) in this
section.

More precisely, our goal is to establish the following auxiliary
claim.  Consider any sequence $\bbellerr = (\BellErrh{1}, \ldots,
\BellErrh{\Ho-1}, \BellErrh{\Ho}=0)$ satisfying the conditions
of~\Cref{PropLin}, and any estimate $\Qfunhat$ such that
$\distrnorm[\big]{ \, \BellOpstarh{\ho} \qfunhath{\ho+1} -
  \qfunhath{\ho} \, }{\ho} \; \leq \; \BellErrh{\ho}$ for $\ho \in
[\Ho-1]$.  We then claim that
\begin{align}
  \label{eq:cor_statement}
  \metrich{\ho} \big( \qfunhath{\ho}, \, \qfunstarh{\ho} \big) \; \leq
  \; \frac{1}{2 \, (\Ho-\ho+1) (1+\log\Ho)} \qquad \mbox{for indices
    $\ho = 1, 2, \ldots, \Ho-1, \Ho$.}
\end{align}
In other words, the estimate $\Qfunhat$ lies with in a neighborhood
$\Neigh$ around $Q$-function $\Qfunstar$ with \mbox{$\radius_{\ho}
  \leq \frac{1}{2} \, (\Ho-\ho+1)^{-1} (1 + \log \Ho)^{-1}$}.  This
auxiliary claim~\eqref{eq:cor_statement} allows us to
invoke~\Cref{lemma:lin_cond}, thereby allowing us to prove the claimed
bound~\eqref{EqnPropLin} as a consequence
of~\Cref{thm:deterministic_star}.  Accordingly, the remainder of our
effort is devoted to the proof of this auxiliary statement.


\subsubsection{Proof of the claim~\eqref{eq:cor_statement}}

We proceed by induction.

\paragraph{Base case:}
Let us first consider the base case with $\ho = \Ho$.  The relation
$\qfunhath{\Ho} = \qfunstarh{\Ho} = \rewardh{\Ho}$ implies
$\metrich{\Ho} \big( \qfunhath{\Ho}, \, \qfunstarh{\Ho} \big) = 0$.
Therefore, inequality~\eqref{eq:cor_statement} naturally holds for
$\ho = \Ho$.

\paragraph{Induction step:}
Suppose that inequality~\eqref{eq:cor_statement} is met for $\ho =
\honew+1, \honew+2, \ldots, \Ho$.  We now establish the
inequality~\eqref{eq:cor_statement} for $\ho = \honew$ based on this
induction hypothesis.

We apply the triangle inequality and derive that
\begin{align}
  \distrnorm[\big]{ \qfunhath{\honew} - \qfunstarh{\honew} }{\honew} &
  \leq \; \distrnorm[\big]{ \BellOpstarh{\honew} \qfunhath{\honew+1} -
    \qfunhath{\honew} }{\honew} +
  \distrnorm[\big]{\BellOpstarh{\honew} \qfunhath{\honew+1} -
    \BellOpstarh{\honew} \qfunstarh{\honew+1}}{\honew} \notag \\ &
  \leq \; \BellErrh{\honew} + \distrnorm[\big]{\BellOpstarh{\honew}
    \qfunhath{\honew+1} -
    \BellOpstarh{\honew}\qfunstarh{\honew+1}}{\honew} \; .
  \label{eq:cor_lin0}
\end{align}
As shown in inequality~\eqref{eq:def_smooth_BellOpstar_app} in the proof of \Cref{lemma:lin_cond}(b) in
\Cref{sec:lin_fun_star_smooth_BellOpstar}, the
bound~\eqref{eq:cor_statement} with $\ho = \honew+1$ implies
\begin{align*}
  \distrnorm[\big]{\BellOpstarh{\honew} \qfunhath{\honew+1} -
    \BellOpstarh{\honew} \qfunstarh{\honew+1}}{\honew} & \; \leq \;
  \Curvstar{\honew} \cdot \distrnorm[\big]{\qfunhath{\honew+1} -
    \qfunstarh{\honew+1} }{\honew+1}
\end{align*}
for $\Curvstar{\honew} = 1 + \frac{1}{2} \, (\Ho-\honew)^{-1}
(1+\log\Ho)^{-1}$.  Furthermore,
inequality~\eqref{eq:fun_diff_BellErr} in
\Cref{sec:fun_diff_BellErr} (Step 3 in the proof of \Cref{thm:deterministic_star}) ensures
\begin{align}
  \distrnorm[\big]{\qfunhath{\honew+1} -
    \qfunstarh{\honew+1}}{\honew+1} & \; \leq \; \sum_{\honewnew =
    \honew+1}^{\Ho-1} \; \Curvstarfun{\honew+1}{\honewnew} \cdot
  \distrnorm[\big]{\BellOpstarh{\honewnew} \qfunhath{\honewnew+1} -
    \qfunhath{\honewnew} }{\honewnew} \; \leq \; \sum_{\honewnew =
    \honew+1}^{\Ho-1} \; \Curvstarfun{\honew + 1}{\honewnew} \cdot
  \BellErrh{\honewnew} \, .
\end{align}
We use the relation $\Curvstar{\honew} \cdot \Curvstarfun{\honew +
  1}{\honewnew} = \Curvstarfun{\honew}{\honewnew}$ and find that
\begin{align*}
  \distrnorm[\big]{\BellOpstarh{\honew} \qfunhath{\honew+1} -
    \BellOpstarh{\honew} \qfunstarh{\honew+1}}{\honew} \; \leq \;
  \sum_{\honewnew = \honew+1}^{\Ho-1} \;
  \Curvstarfun{\honew}{\honewnew} \cdot \BellErrh{\honewnew} \, .
\end{align*}
Under the induction hypothesis that
inequality~\eqref{eq:cor_statement} holds for $\ho = \honew + 1,
\honew + 2, \ldots, \Ho$, \Cref{lemma:lin_cond}(b) guarantees
$\Curvstarfun{\honew}{\honewnew} \leq 3$\,.  It follows that
\begin{align}
  \label{eq:cor_lin1}
  \distrnorm[\big]{\BellOpstarh{\honew} \qfunhath{\honew+1} -
    \BellOpstarh{\honew} \qfunstarh{\honew+1}}{\honew} \; \leq \; 3 \;
  \sum_{\honewnew = \honew+1}^{\Ho-1} \; \BellErrh{\honewnew} \, .
\end{align}
Combining the bound~\eqref{eq:cor_lin1} with
inequality~\eqref{eq:cor_lin0} yields
\begin{align*}
  \distrnorm[\big]{ \qfunhath{\honew} - \qfunstarh{\honew} }{\honew} &
  \; \leq \; \BellErrh{\honew} + 3 \; \sum_{\honewnew =
    \honew+1}^{\Ho-1} \; \BellErrh{\honewnew} \; \leq \; 3 \,
  (\Ho-\honew+1) \, \BellErrh{\honew-1} \, ,
\end{align*}
where the second inequality follows from the regularity condition \eqref{EqnRegular}.

We further apply the definition of metric $\metrich{\ho}$ in
equation~\eqref{eq:def_metric_lin} and obtain
\begin{align*}
  \metrich{\honew} \big( \qfunhath{\honew}, \, \qfunstarh{\honew}
  \big) = \frac{\sqrt{\Dim} \; \CurveNorm{\MyCurve{\honew}}{\honew}}
      {\distrnorm{\qfunstarh{\honew}}{\honew}} \, \cdot \,
      \distrnorm[\big]{ \qfunhath{\honew} - \qfunstarh{\honew}
      }{\honew} \ \leq \ \frac{3 \sqrt{\Dim} \;
        \CurveNorm{\MyCurve{\honew}}{\honew} \, (\Ho-\honew+1) }
                {\distrnorm{\qfunstarh{\honew}}{\honew}} \cdot
                \BellErrh{\honew-1} \, . 
\end{align*} 
Substituting $\BellErrh{\honew-1}$ with its upper bound in
inequality~\eqref{eq:cor_n_cond} then leads to the validity of
inequality~\eqref{eq:cor_statement} with $\ho = \honew$, which
completes the proof of~\Cref{PropLin}.



\section{Discussion \yaqidone}
\label{sec:conclusion}

This paper introduces a novel approach for the analysis of value-based
RL methods for continuous state-action spaces.  Our analysis
highlights two key stability properties of MDPs under which much
sharper bounds on value sub-optimality can be guaranteed.  Studying in
some detail the case of linear approximations to value functions, we
showed that these stability conditions hold for a broad class of
problems.  Our analysis offers fresh perspectives on the commonly used
pessimism and optimism principles, in off-line and on-line settings
respectively, and highlight connections between RL and transfer
learning.

Our study leaves open various questions for future work. First, our
main result (\Cref{thm:deterministic_star}) has consequences for
linear quadratic control, to be described in an upcoming
paper~\cite{X}.  It provides insight into the role of covariate shift
in linear quadratic control, as well as efficient exploration in the
on-line setting.  Second, our current statistical analysis focused on
i.i.d. data with linear function approximation.  It is interesting to
consider the extensions to dependent data and non-parametric function
approximation (e.g. kernels, boosting, and neural networks).  Third,
while this paper has provided upper bounds, it remains to address the
complementary question of lower bounds for policy optimization over
the classes of stable MDPs isolated here.  Last, to better align our
framework with real-world scenarios, we intend to go beyond the
idealized completeness condition used in this paper, and treat the
role of model mis-specification.


\subsection*{Acknowledgements}

This work was partially supported by NSF grant CCF-1955450, ONR grant
N00014-21-1-2842, and NSF DMS-2311072 to MJW.




\bibliographystyle{abbrv} \bibliography{ref}
  

\appendix

\section{General set-up for stability}
\label{AppGeneral}

In this appendix, we describe a general set-up for stability, along
with a precise definition of the local neighborhood $\PlainNeigh$ in
our main theorem.

\subsection{Stability via compatible semi/pseudo-norms}

In the main text, we defined stability conditions in terms of the
$L^2$-norm induced by the occupation measure of the optimal policy.
Here we generalize this definition by allowing for more general pairs
of (semi/pseudo) norms.  In doing so, it is convenient to define the
\emph{Minkowski difference}
\begin{align}
  \label{eq:def_DiffRKHS}
  \DiffRKHS \defn \RKHS - \RKHS = \big\{ \funh{} - \funnewh{} \bigm|
  \funh{}, \funnewh{} \in \RKHS \big \}.
\end{align}

\noindent Given a value function estimate $\Qfunhat$, our more general
framework involves two notions of its closeness to the optimal
$Q$-function, as defined by pseudo-metrics $\{\metrich{\ho}
\}_{\ho=1}^{\Ho}$ and semi-norms $\{ \|\cdot\|_\ho \}_{\ho=1}^\Ho$ on
the difference class $\DiffRKHS$.
\begin{carlist}
\item Our stability conditions are defined on a neighborhood $\Neigh$
  of the optimal $Q$-function, as specified by pseudo-metrics $\{
  \metrich{\ho} \} _{\ho=1}^{\Ho}$.
\item The resulting error bounds are stated in terms of the Bellman
  residuals $\distrnorm[\big]{ \, \BellOpstarh{\ho} \qfunhath{\ho+1} -
    \qfunhath{\ho} }{\ho}$, as measured in the semi-norm
  $\distrnorm[]{\cdot}{\ho}$.
\end{carlist}
\noindent We require that this pair of pseudo-metric and semi-norm are
\emph{compatible} in the sense that
\begin{align}
  \label{EqnPseudoCondition}
  \sup_{\gfunh{} \in \DiffRKHS \, : \, \distrnorm{\gfunh{}}{\ho} > 0}
  \frac{\distrnorm[\big]{ \,\big( \MOppih{\ho-1}{\policy} - \,
      \MOpstarh{\ho-1} \big) \, \gfunh{} \;
    }{\ho-1}}{\distrnorm{\gfunh{}}{\ho}} & \; \leq \;
  \metrich{\ho}\big(\qfunh{\ho}, \, \qfunstarh{\ho}\big),
\end{align}
where $\policy$ denotes the greedy policy associated with function
$\Qfun$. We adopt $\MOpstarh{\ho}$ as a convenient shorthand for the transition
operator~$\TransOph{\ho}^{\policystar}$ defined by an optimal policy
$\policystar$.\footnote{We adopt a complementary definition that
$\MOppih{0}{\policy} = 0$ for any policy $\policy$.}

Defining the multi-step transition operator
\mbox{$\MOpstarhtoh{\ho}{\honew} \defn \MOpstarh{\ho} \,
  \MOpstarh{\ho+1} \cdots \MOpstarh{\honew-1}$,} we require that the
semi-norms satisfy the bound
\begin{align}
  \label{EqnPseudoStable}
  \distrnorm[\big]{\MOpstarhtoh{\ho}{\honew} \, \funh{\,}}{\ho} \;
  \leq \; \Radphistar \; \distrnorm{ \funh{\,}}{\honew} \qquad
  \mbox{for any $\funh{} \in \DiffRKHS$,}
\end{align}
uniformly over all pairs $\ho < \honew$.

As one special case (discussed in the main text), suppose that
$\distrnorm{\cdot}{\ho}$ is the $L^2$-norm induced by the state-action
occupation measure induced by the optimal policy---that is
\begin{align}
  \distrnorm{\funh{\,}}{\ho} \defn \sqrt{\Exp_{\occupstar} \big[
      \funh{}^2(\Stateh{\ho}, \Actionh{\ho}) \big]} \qquad \mbox{for
    any $\funh{} \in \DiffRKHS$}.
    \tag{\ref{EqnDefnSAOcc}}
\end{align}
With this choice, it can be verified (see~\Cref{AppOccCase} for
details) that condition~\eqref{EqnPseudoStable} holds with
$\Radphistar = 1$.

For certain problems, this pair can be chosen to be equivalent,
meaning that \mbox{$\metrich{\ho}(\funh{\,}, \funnewh{}) = \plaincon
  \, \distrnorm{\funh{\,} - \funnewh{}}{\ho}$} for some universal
constant $\plaincon > 0$.  For instance, this choice is valid when
using linear function approximation, as discussed in detail
in~\Cref{sec:lin_fun_star}.  However, it is useful to retain the
flexibility of a general choice of these pseudo-metrics.

\subsection{Stability neighborhood}

Let us now define the neighborhood $\Neigh$ over which the
stability conditions are assumed to hold.  It is specified by a
sequence $\bradius = (\radius_1, \ldots, \radius_\Ho)$ of positive
reals that are small enough to satisfy the bound \mbox{$\radius_{\ho}
  \leq \frac{1}{2 \, \Radphistar} \, (\Ho-\ho+1)^{-1}$.}  Here
$\Radphistar$ is the stability parameter given in the
bound~\eqref{EqnPseudoStable}.  Given any such sequence, we say that
\mbox{a $Q$-function} \mbox{$\Qfun = (\qfun_1, \ldots, \qfun_\Ho)$} is
a \emph{$\bradius$-good approximation} to the optimal $Q$-function
\mbox{$\Qfunstar = (\qfunstar_1, \ldots, \qfunstar_\Ho)$} if
\begin{align} 
  \label{EqnRhoGood}
  \metrich{\ho}\big(\qfunh{\ho}, \, \qfunstarh{\ho} \big) & \; \leq \;
  \radius_{\ho} \qquad \mbox{for $\ho \in [\Ho]$.}
\end{align}
We use $\Neigh$ as a shorthand for the set of all $Q$-functions that
are $\bradius$-good approximations to $\Qfunstar$.  In our statement
of~\Cref{thm:deterministic_star}, the neighborhood $\PlainNeigh$ is
equivalent to $\Neigh$ defined in this way.


\section{Proof of auxiliary lemmas for~\Cref{thm:deterministic_star}}

We now turn to proofs of the two auxiliary results used to establish
our main theorem, with~\Cref{lemma:telescope,lemma:smooth} treated in
in~\Cref{proof:subopt_ub,sec:proof:lemma:smooth}, respectively.

\subsection{Proof of Lemma~\ref{lemma:telescope}}
\label{proof:subopt_ub}

For any integrable vector function~$\funnew = (\funnewh{1}, \ldots,
\funnewh{\Ho}) \in \Real^{\StateSp \times \ActionSp \times \Ho}$, we
define
\begin{subequations}
\begin{align}
  \label{eq:def_Diffun}
  \Diffun(\funnew) & \; = \; \myssum{\ho=1}{\Ho} \big(
  \Exp_{\policy} - \Exp_{\policyhat} \big)
  \big[ \funnewh{\ho} (\Stateh{\ho}, \Actionh{\ho}) \big] \, .
\end{align}
We claim that this functional satisfies the recursive relation
\begin{align}
  \label{eq:Exp_funnew0}
  \Diffun(\funnew) = \myssum{\ho=1}{\Ho} \, \Exp_{
    \policyhat} \big[ \funnewh{\ho}(\Stateh{\ho},
    \policyh{\ho}(\Stateh{\ho})) - \funnewh{\ho}(\Stateh{\ho},
    \policyhath{\ho}(\Stateh{\ho})) \big] +
  \Diffun(\TransOppi{\policy} \funnew),
\end{align}
\end{subequations}
where we have introduced the shorthand $\TransOppi{\policy} \funnew
\defn \big(\TransOppih{1}{\policy} \, \funnewh{2}, \ldots,
\TransOppih{\Ho-1}{\policy} \, \funnewh{\Ho}, 0 \big) \in
\Real^{\StateSp \times \ActionSp \times \Ho}$. \\

\vspace*{0.05in}

Taking this claim as given for the moment, let us prove the
bound~\eqref{eq:subopt_ub} from~\Cref{lemma:telescope}.  First, we set
$\funnew \defn (\TransOppi{\policy})^{\ho} \, \funnew = \big(
\MOppihtoh{1}{1+\ho}{\policy} \, \funnewh{1+\ho}, \ldots,
\MOppihtoh{\Ho-\ho}{\Ho}{\policy} \, \funnewh{\Ho}, 0, \ldots, 0
\big)$ in equation~\eqref{eq:Exp_funnew0} for \mbox{$\ho = 0, 1,
  \ldots, \Ho\!-\!1$}, which yields
\begin{align*}
\Diffun\big( (\TransOppi{\policy})^{\ho} \, \funnew \big) =
\myssum{\begin{subarray}{c} 1 \leq \honew \leq \honewnew \leq \Ho ,
    \\ \honewnew - \honew = \ho \end{subarray}}{} \,
\Exp_{\policyhat} \big[ \{
  \MOppihtoh{\honew}{\honewnew}{\policy} \, \funnewh{\honewnew} \}
  (\Stateh{\honew}, \policyh{\honew}(\Stateh{\honew})) - \{
  \MOppihtoh{\honew}{\honewnew}{\policy} \, \funnewh{\honewnew}
  \}(\Stateh{\honew}, \policyhath{\honew}(\Stateh{\honew})) \big] +
\Diffun\big( (\TransOppi{\policy})^{\ho+1} \, \funnew \big).
\end{align*}
Note that $(\TransOppi{\policy})^{\Ho} \, \funnew = 0$, which implies
$\Diffun\big( (\TransOppi{\policy})^{\Ho} \, \funnew \big) = 0$.  We
then sum the resulting bounds so as to obtain
\begin{align} 
  \Diffun(\funnew)
  \label{eq:diff_distr}  
  & = \sum_{1 \leq \ho \leq \honew \leq \Ho} \Exp_{\policyhat} \big[ \{ \MOppihtoh{\ho}{\honew}{\policy} \,
    \funnewh{\honew} \} (\Stateh{\ho}, \policyh{\ho}(\Stateh{\ho})) -
    \{ \MOppihtoh{\ho}{\honew}{\policy} \, \funnewh{\honew} \}
    (\Stateh{\ho}, \policyhath{\ho}(\Stateh{\ho})) \big].
\end{align}
Setting \mbox{$\funnew = \BellOpstar \qfunhat - \qfunhat$}, or equivalently $\funnewh{\ho} = \BellOpstarh{\ho} \qfunhath{\ho+1} - \qfunhath{\ho}$, in
equation~\eqref{eq:diff_distr}, we find that
\begin{align*}
D \big(\BellOpstar{} \qfunhat{} - \qfunhat{}\big) & =
\myssum{\ho=1}{\Ho-1} \Exp_{ \policyhat} \big[
  \Deltafunh{\ho}(\policy; \Stateh{\ho}, \policyh{\ho}(\Stateh{\ho}))
  - \Deltafunh{\ho}(\policy; \Stateh{\ho},
  \policyhath{\ho}(\Stateh{\ho})) \big] \, ,
\end{align*}
where we have used the fact~\eqref{eq:def_Deltafun} that
\mbox{$\Deltafunh{\ho}(\policy; \, \cdot) = \sum_{\honew=\ho}^{\Ho}
  \MOppihtoh{\ho}{\honew}{\policy} \big(\BellOpstarh{\honew}
  \qfunhath{\honew+1} - \qfunhath{\honew} \big)$.}  Thus, we have
established the bound~\eqref{eq:subopt_ub} stated
in~\Cref{lemma:telescope}. \\
\vspace*{0.1in}

It remains to establish the auxiliary claim~\eqref{eq:Exp_funnew0}.
Note that the functional $\Diffun$ can be decomposed as
\mbox{$\Diffun(\funnew) = \Diffun_1 + \Diffun_2$,} where
\begin{align*}
  \Diffun_1 & \defn \ssum{\ho=1}{\Ho} \Exp _{\policyhat}
  \big[ \funnewh{\ho}(\Stateh{\ho}, \policyh{\ho}(\Stateh{\ho})) -
    \funnewh{\ho}(\Stateh{\ho}, \policyhath{\ho}(\Stateh{\ho})) \big]
  \quad \text{and} \quad \\
  \Diffun_2 & \defn \ssum{\ho=1}{\Ho} \big( \Exp_{
    \policy} - \Exp_{\policyhat} \big) \big[
    \funnewh{\ho} (\Stateh{\ho}, \policyh{\ho}(\Stateh{\ho})) \big] \,
  .
\end{align*}
Applying the tower property of conditional expectation, we find that
\begin{align*}
  \Diffun_2 & = \myssum{\ho=1}{\Ho-1} \big( \Exp_{
    \policy} - \Exp_{\policyhat} \big) \big[ \Exp [
      \funnewh{\ho+1}(\Stateh{\ho+1}, \policyh{\ho+1}(\Stateh{\ho+1}))
      \mid \Stateh{\ho}, \Actionh{\ho} ] \big] \\
& = \myssum{\ho=1}{\Ho-1} \big( \Exp_{\policy} -
  \Exp_{\policyhat} \big) \big[
    (\TransOppih{\ho}{\policy} \, \funnewh{\ho+1}) (\Stateh{\ho},
    \Actionh{\ho}) \big] \\
 & = \Diffun \big(\TransOppi{\policy} \funnew \big).
\end{align*}
Combining the expressions for $\Diffun_1$ and $\Diffun_2$ above yields
the claim~\eqref{eq:Exp_funnew0}.


\subsection{Proof of Lemma~\ref{lemma:smooth}}
\label{sec:proof:lemma:smooth}

The key step in proving~\Cref{lemma:smooth} is establishing that
inequality~\eqref{eq:def_Ttilfast_new} holds when the function
estimation error $\metrich{\ho}\big(\qfunhath{\ho},
\qfunstarh{\ho}\big)$ is sufficiently small.  In order to do so, we
need to establish upper bounds on the term $\DiffOphat(\ho,\honew)$ by
using $\DiffOp(\ho,\honew)$.
%
%
In particular, we will show that for any $1 \leq \ho \leq \honew \leq
\Ho-1$,
\begin{align}
  \label{eq:control_DDiffOp_1_new}
  \DiffOphat(\ho, \, \honew) \; \leq \; \DiffOp(\ho, \, \honew) \; +
  \; \sum_{\honewnew=1}^{\ho-1} \; {\DiffOphat(\honewnew, \, \ho-1)}
  \ \cdot \ \Radphistar \, \cdot \, \metrich{\ho} \big(
  \qfunhath{\ho}, \, \qfunstarh{\ho} \big) \, .
\end{align}
The inequality~\eqref{eq:control_DDiffOp_1_new} is derived based on
the bounds~\eqref{EqnPseudoCondition} and~\eqref{EqnPseudoStable} that
define metric $\metrich{\ho}$ and parameter~$\Radphistar$.  After a
close examination of the right-hand side of this inequality, it
becomes evident that as long as the function estimation error
$\metrich{\ho}\big(\qfunhath{\ho}, \qfunstarh{\ho}\big)$ is
sufficiently small, the terms associated with
$\metrich{\ho}\big(\qfunhath{\ho}, \qfunstarh{\ho}\big)$ are
negligible and are dominated by $\DiffOp(\ho, \, \honew)$.
Consequently, inequality~\eqref{eq:def_Ttilfast_new} within the
arguments in \Cref{sec:proof:thm_outline_StepTwo} is likely to hold
true.

With claim~\eqref{eq:control_DDiffOp_1_new} assumed to be valid at
this point, we now establish a proper upper bound on the estimation error
$\metrich{\ho}\big(\qfunhath{\ho}, \qfunstarh{\ho}\big)$ under which
inequality~\eqref{eq:def_Ttilfast_new} is satisfied.  By taking linear
combinations of inequality~\eqref{eq:control_DDiffOp_1_new} using
weights $\bbellerr = (\BellErrh{1}, \ldots, \BellErrh{\Ho-1},
\BellErrh{\Ho}=0)$, we obtain
\begin{align}
  \sum_{\ho=1}^{\Ho-1} \sum_{\honew=\ho}^{\Ho-1} \, \DiffOphat(\ho, \,
  \honew) \cdot \BellErrh{\honew} & \; \leq \; \sum_{\ho=1}^{\Ho-1}
  \sum_{\honew=\ho}^{\Ho-1} \, \DiffOp(\ho, \, \honew) \cdot
  \BellErrh{\honew} \notag \\ & \quad + \sum_{\ho=2}^{\Ho-1}
  \sum_{\honewnew=1}^{\ho-1} \, \DiffOphat(\honewnew, \, \ho-1) \;
  \cdot \; \Radphistar \; \cdot \; \metrich{\ho} \big(\qfunhath{\ho},
  \, \qfunstarh{\ho} \big) \; \sum_{\honew=\ho}^{\Ho-1} \,
  \BellErrh{\honew} \, .
  \label{eq:control_DDiffOp_1_new_new}
\end{align}
When the sequence $\bbellerr = (\BellErrh{1}, \ldots,
\BellErrh{\Ho-1}, \BellErrh{\Ho}=0)$ is regular in the sense that
inequality~\eqref{EqnRegular} holds, the bound~\eqref{eq:control_DDiffOp_1_new_new}
reduces to
\begin{align*}
  \sum_{1 \leq \ho \leq \honew \leq \Ho} \, \DiffOphat(\ho, \, \honew)
  \cdot \BellErrh{\honew} & \; \leq \; \sum_{1 \leq \ho \leq \honew
    \leq \Ho} \, \DiffOp(\ho, \, \honew) \cdot \BellErrh{\honew} \\ &
  \quad + \sum_{1 \leq \ho \leq \honew \leq \Ho-2} \, \DiffOphat(\ho,
  \, \honew) \cdot \BellErrh{\honew} \; \cdot \; \Radphistar \, (\Ho -
  \honew) \; \cdot \; \metrich{\honew+1} \big(\qfunhath{\honew+1}, \,
  \qfunstarh{\honew+1} \big) \, .
\end{align*}
Under the condition $\metrich{\ho} \big(\qfunhath{\ho}, \,
\qfunstarh{\ho} \big) \leq \frac{1}{2 \, \Radphistar } (\Ho - \ho +
1)^{-1}$ for $2 \leq \ho \leq \Ho-1$, the inequality above implies
bound~\eqref{eq:def_Ttilfast_new}, which further establishes the
bound~\eqref{eq:stable}, as stated in \Cref{lemma:smooth}.  \\


It remains to prove the relation between $\DiffOphat(\ho, \, \honew)$
and $\DiffOp(\ho, \, \honew)$, as shown in
inequality~\eqref{eq:control_DDiffOp_1_new}.

\paragraph{Proof of
bound~\eqref{eq:control_DDiffOp_1_new}:}

It is evident that inequality~\eqref{eq:control_DDiffOp_1_new} holds
for $\ho = 1$, therefore, we focus on its validation for indices $2
\leq \ho \leq \Ho-1$.  Recall the definitions of functions
$\DiffOphat(\ho, \honew)$ and $\DiffOp(\ho, \honew)$, as given by
equations \eqref{eq:def_DiffOphat} and \eqref{eq:def_DiffOp}.  We
apply the triangle inequality and derive that
\begin{align*}
  & \abs[\big]{\DiffOphat(\ho, \, \honew) - \DiffOp(\ho, \, \honew)}
  \notag \\ & \leq \; \sup_{\funh{} \in \DiffRKHS: \,
    \distrnorm{\funh{}}{\honew} > 0} \; \left\{
  \frac{1}{\distrnorm{\funh{}}{\honew}} \abs[\Big]{ \big(
    \Exp_{\policyhat} - \Exp_{\policystar}
    \big) \Big[ \big( \MOpstarhtoh{\ho}{\honew} \, \funh{\,}
      \big)(\Stateh{\ho}, \policystarh{\ho} (\Stateh{\ho})) - \big(
      \MOpstarhtoh{\ho}{\honew} \, \funh{\,} \big) (\Stateh{\ho},
      \policyhath{\ho}(\Stateh{\ho})) \Big]} \right\} \notag \\ & = \;
  \sup_{\funh{} \in \DiffRKHS: \, \distrnorm{\funh{}}{\honew} > 0}
  \left\{ \frac{1}{\distrnorm{\funh{}}{\honew}} \abs[\Big]{ \big(
    \Exp_{\policyhat} - \Exp_{\policystar}
    \big) \Big[ \big\{ \big( \MOpstarh{\ho-1} -
      \MOppih{\ho-1}{\policyhat} \big) \, \MOpstarhtoh{\ho}{\honew} \,
      \funh{\,} \big\} (\Stateh{\ho-1}, \Actionh{\ho-1}) \Big]}
  \right\} \nfed \DDiffOp(\ho, \, \honew) \, .
\end{align*}

The term~$\DDiffOp(\ho, \, \honew)$ involves differences from two
sources: (i) the difference in transition kernels $\MOpstarh{\ho-1} -
\MOppih{\ho-1}{\policyhat}$ that captures the divergence between
policies $\policystarh{\ho}$~and~$\policyhath{\ho}$; (ii) the
discrepancy of occupation measures at the ($\ho-1$)-th step reflected
by the difference in expectations $\Exp_{\policystar} -
\Exp_{\policyhat}$, which is determined by the policies
$(\policystarh{1}, \ldots, \policystarh{\ho-1})$ and $(\policyhath{1},
\ldots, \policyhath{\ho-1})$ until the ($\ho-1$)-th step.  We treat
them separately and write
\begin{align}
  \label{eq:control_DDiffOp_0}
  \DDiffOp(\ho, \, \honew) \; \leq \; \DDiffOptwo(\ho-1, \, \honew) \,
  \cdot \, \DDiffOpone(\ho-1) \, ,
\end{align}
where the functionals $\DDiffOpone$ and $\DDiffOptwo$ are defined as
\begin{align*}
    \DDiffOptwo(\ho-1, \, \honew) & \; \defn \; \sup_{\funh{} \in
      \DiffRKHS: \, \distrnorm{\funh{}}{\honew} > 0} \; \left\{
    \frac{1}{\distrnorm{\funh{}}{\honew}} \distrnorm[\big]{\big(
      \MOpstarh{\ho-1} - \MOppih{\ho-1}{\policyhat} \big) \,
      \MOpstarhtoh{\ho}{\honew} \, \funh{\,}}{\ho-1} \right\} ,
    \\ \DDiffOpone(\ho-1) & \; \defn \; \sup_{\funh{} \in \DiffRKHS:
      \, \distrnorm{ \funh{}}{\ho-1} > 0} \; \left\{
    \frac{1}{\distrnorm{\funh{}}{\ho-1}} \abs[\big]{\big(
      \Exp_{\policyhat} - \Exp_{\policystar}
      \big) \big[ \funh{}(\Stateh{\ho-1}, \Actionh{\ho-1}) \big]}
    \right\} .
\end{align*}

We first consider the term $\DDiffOptwo$.  According to the
definitions of metric~$\metrich{\ho}$ and parameter $\Radphistar$ in
inequalities \eqref{EqnPseudoCondition}~and \eqref{EqnPseudoStable},
we find that
\begin{align*}
  \distrnorm[\big]{\big( \MOpstarh{\ho-1} - \MOppih{\ho-1}{\policyhat}
    \big) \, \MOpstarhtoh{\ho}{\honew} \, \funh{\,}}{\ho-1} &
  \stackrel{\eqref{EqnPseudoCondition}}{\leq} \metrich{\ho}
  \big(\qfunhath{\ho}, \, \qfunstarh{\ho} \big) \cdot
  \distrnorm[\big]{ \MOpstarhtoh{\ho}{\honew} \, \funh{\,}}{\ho}
  \stackrel{\eqref{EqnPseudoStable}}{\leq} \metrich{\ho}
  \big(\qfunhath{\ho}, \, \qfunstarh{\ho} \big) \cdot \Radphistar \,
  \distrnorm{\funh{}}{\honew} \, ,
\end{align*}
which in turn implies
\begin{subequations}
\begin{align}
  \label{eq:control_DDiffOp_1}
  \DDiffOptwo(\ho-1, \, \honew) & \; \leq \; \Radphistar \, \cdot \,
  \metrich{\ho} \big( \qfunhath{\ho}, \, \qfunstarh{\ho} \big) \, .
\end{align}
As for term~$\DDiffOpone$, we claim that
\begin{align}
  \label{eq:control_Diffphi}
  \DDiffOpone(\ho-1) & \, \leq \; \sum_{\honewnew=1}^{\ho-1} \;
             {\DiffOphat(\honewnew, \, \ho-1)} \; .
\end{align}
\end{subequations}
Combining the bound $\DiffOphat(\ho, \, \honew) \leq \DiffOp(\ho, \,
\honew) + \DDiffOp(\ho, \, \honew)$ with
inequalities~\eqref{eq:control_DDiffOp_0},
\eqref{eq:control_DDiffOp_1}~and~\eqref{eq:control_Diffphi}, we
establish the bound~\eqref{eq:control_DDiffOp_1_new}, as claimed.  It
remains to prove the claim~\eqref{eq:control_Diffphi}.

\paragraph{Proof of inequality~\eqref{eq:control_Diffphi}:}

This proof is analogous to that of~\Cref{lemma:telescope}.  We begin
by introducing an analogue of the functional $\Diffun(\funnew)$ from
equation~\eqref{eq:def_Diffun}; in particular, for any index $\ho \in
[\Ho-1]$ and function $\funnewh{} \in \DiffRKHS$, define
\begin{align*}
  \Diffunstar_{\ho}(\funnewh{}) & \defn \big( \Exp_{
    \policystar} - \Exp_{\policyhat} \big) \big[
    \funnewh{} (\Stateh{\ho}, \Actionh{\ho}) \big] \, .
\end{align*}
Using the notation of $\Diffunstar_{\ho}$, we can rewrite the
left-hand side of inequality~\eqref{eq:control_Diffphi} as
$\DDiffOpone(\ho-1) = \sup_{\funh{} \in \DiffRKHS: \, \distrnorm{
    \funh{}}{\ho-1} > 0} \big\{\abs{\Diffunstar_{\ho-1}(\funh{})} /
\distrnorm{\funh{}}{\ho-1} \big\}$.  Following the same arguments as
in the proof of inequality~\eqref{eq:Exp_funnew0}, we can show that
\begin{align}
  \label{eq:recurse_Diffunstar}
  \Diffunstar_{\ho}(\funnewh{}) = \Exp_{\policyhat} \big[
    \funnewh{}(\Stateh{\ho}, \policystarh{\ho}(\Stateh{\ho})) -
    \funnewh{}(\Stateh{\ho}, \policyhath{\ho}(\Stateh{\ho})) \big] +
  \Diffunstar_{\ho-1} (\MOpstarh{\ho-1} \, \funnewh{}) \qquad
  \mbox{for $\ho = 1,2,\ldots,\Ho$},
\end{align}
where we set $\Diffunstar_0 \equiv 0$.

We consider function $\funnewh{} \defn \MOpstarhtoh{\honewnew}{\ho-1}
\, \funh{\,}$ for $1 \leq \honewnew < \ho \leq \Ho-1$.  It follows
from equation~\eqref{eq:recurse_Diffunstar} that
\begin{align*}
  \Diffunstar_{\honewnew} \big(\MOpstarhtoh{\honewnew}{\ho-1} \,
  \funh{\,} \big) & = \Exp_{\policyhat} \big[
    \big(\MOpstarhtoh{\honewnew}{\ho-1} \funh{\,} \big)(
    \Stateh{\honewnew}, \policystarh{\honewnew}(\Stateh {\honewnew}))
    \!  - \! \big(\MOpstarhtoh{\honewnew}{\ho-1} \funh{\,}
    \big)(\Stateh{\honewnew}, \policyhath{\honewnew}(\Stateh{
      \honewnew})) \big] + \Diffunstar_{\honewnew-1}
  \big(\MOpstarhtoh{\honewnew-1}{\ho-1} \, \funh{\,} \big) \, ,
\end{align*}
where we have used the relation $\MOpstarh{\honewnew-1}
\MOpstarhtoh{\honewnew}{\ho-1} = \MOpstarhtoh{\honewnew-1}{\ho-1}$.
Recalling the definition of $\DiffOphat(\honewnew, \, \ho-1)$ in
equation~\eqref{eq:def_DiffOphat}, applying the triangle inequality
yields
\begin{align*}
  \abs[\big]{\Diffunstar_{\honewnew}
    \big(\MOpstarhtoh{\honewnew}{\ho-1} \, \funh{\,} \big)} & \; \leq
  \; \DiffOphat(\honewnew, \, \ho-1) \cdot \distrnorm{\funh{}}{\ho-1}
  \, + \, \abs[\big]{\Diffunstar_{\honewnew-1}
    \big(\MOpstarhtoh{\honewnew-1}{\ho-1} \, \funh{\,} \big)} \; .
\end{align*}
Summing this equation over indices $\honewnew = 1, 2, 3, \ldots,
\ho-1$ yields
\begin{align*}
  \abs{\Diffunstar_{\ho-1}(\funh{})} & \, \leq \;
  \sum_{\honewnew=1}^{\ho-1} \; \DiffOphat(\honewnew, \, \ho-1) \cdot
  \distrnorm{\funh{}}{\ho-1} \, ,
\end{align*}
which establishes inequality~\eqref{eq:control_Diffphi}.



\section{Proof of Lemma~\ref{lemma:lin_cond}}
\label{sec:proof:lemma:lin_cond}

In this section, we prove the main auxiliary result used in the proof
of~\Cref{PropLin}.  We devote a subsection to each of the three claims
in the lemma.


\subsection{Proof of Lemma~\ref{lemma:lin_cond}(a)}

Here we need to show that the bound~\eqref{EqnPseudoCondition} holds
for the distance function that we have chosen.  Consider any function
$\funnewh{}(\cdot) = \inprod{\bweight_{\funnewh{}}}{\Feature(\cdot)}
\in \DiffRKHS$.  For $\ho = 2, 3, \ldots, \Ho-1$, the difference
between transition operators $\MOppih{\ho-1}{\policy}$ and
$\MOpstarh{\ho-1}$ takes the form
\begin{align*}
  \big( (\MOppih{\ho-1}{\policy} - \, \MOpstarh{\ho-1} ) \, \gfunh{}
  \big)(\state, \action) & = \Exp\big[ \funnewh{}(\Stateh{\ho},
    \policyh{\ho}(\Stateh{\ho})) - \funnewh{}(\Stateh{\ho},
    \policystarh{\ho}(\Stateh{\ho})) \bigm| \Stateh{\ho-1} = \state,
    \Actionh{\ho-1} = \action \big] \\ & = \inprod[\Big]{\Exp\big[
      \Feature(\Stateh{\ho}, \policyh{\ho}(\Stateh{\ho})) -
      \Feature(\Stateh{\ho}, \policystarh{\ho}(\Stateh{\ho})) \bigm|
      \Stateh{\ho-1} = \state, \Actionh{\ho-1} = \action \big]}
      {\bweight_{\funnewh{}}} \, .
\end{align*}
Applying the Cauchy--Schwarz inequality yields
\begin{subequations}
\begin{align}
  & \big( (\MOppih{\ho-1}{\policy} - \, \MOpstarh{\ho-1} ) \, \gfunh{}
  \big)^2(\state, \action) \notag \\
  & \leq \distrnorm[\Big]{\Exp\big[ \Feature(\Stateh{\ho},
      \policyh{\ho}(\Stateh{\ho})) - \Feature(\Stateh{\ho},
      \policystarh{\ho}(\Stateh{\ho})) \bigm| \Stateh{\ho-1} = \state,
      \Actionh{\ho-1} = \action \big]}{\CovOp{\ho}^{-1}}^2
  \cdot\distrnorm{\bweight_{\funnewh{}}}{\CovOp{\ho}}^2 \notag \\ &
  \leq \Exp\Big[ \distrnorm[\big]{ \Feature(\Stateh{\ho},
      \policyh{\ho}(\Stateh{\ho})) - \Feature(\Stateh{\ho},
      \policystarh{\ho}(\Stateh{\ho})) }{\CovOp{\ho}^{-1}}^2 \Bigm|
    \Stateh{\ho-1} = \state, \Actionh{\ho-1} = \action \, \Big] \cdot
  \distrnorm{\bweight_{\funnewh{}}}{\CovOp{\ho}}^2 \, .
  \label{eq:Pdiff0}
\end{align}
The definition of the norm $\distrnorm{\,\cdot\,}{\ho}$ ensures
$\distrnorm{\bweight_{\funnewh{}}}{\CovOp{\ho}} =
\distrnorm{\funnewh{}}{\ho}$.  By using the curvature
property~\ref{eq:curv_1}, we find that
\begin{align}
  \label{eq:curv_1_app0}
  \distrnorm[\big]{ \Feature(\Stateh{\ho},
    \policyh{\ho}(\Stateh{\ho})) - \Feature(\Stateh{\ho},
    \policystarh{\ho}(\Stateh{\ho})) }{\CovOp{\ho}^{-1}}^2 \; \leq \;
  \MyCurveTwosq{\ho}{\Stateh{\ho}} \; \Dim \, \cdot \,
  \frac{\distrnorm[\big]{\funh{\ho} - \qfunstarh{\ho}}{\ho}^2}
       {\distrnorm{\qfunstarh{\ho}}{\ho}^2} \; .
\end{align}
\end{subequations}
The combination of inequalities~\eqref{eq:Pdiff0}
and~\eqref{eq:curv_1_app0} leads to the following bound:
\begin{align*}
  \big( (\MOppih{\ho-1}{\policy} - \, \MOpstarh{\ho-1} ) \, \gfunh{}
  \big)^2(\state, \action) & \; \leq \; \Exp\big[
    \MyCurveTwosq{\ho}{\Stateh{\ho}} \bigm| \Stateh{\ho-1} = \state,
    \Actionh{\ho-1} = \action \, \big] \cdot \Dim \, \cdot \,
  \frac{\distrnorm[\big]{\funh{\ho} - \qfunstarh{\ho}}{\ho}^2}
       {\distrnorm{\qfunstarh{\ho}}{\ho}^2} \cdot
       \distrnorm{\funnewh{}}{\ho}^2 \, .
\end{align*}
Taking expectations over the state-action pairs under the occupation
measure {$\distrstarh{\ho-1}$} yields
\begin{align*}
  \distrnorm[\big]{ \,\big( \MOppih{\ho-1}{\policy} - \,
    \MOpstarh{\ho-1} \big) \, \gfunh{} \; }{\ho-1}^2 & \; = \;
  \Exp_{\policystar}\Big[ \big( (\MOppih{\ho-1}{\policy}
    - \, \MOpstarh{\ho-1} ) \, \gfunh{} \big)^2(\Stateh{\ho-1},
    \Actionh{\ho-1}) \, \Big] \\ & \; \leq \; \Exp_{
    \policystar} \big[ \MyCurveTwosq{\ho}{\Stateh{\ho}} \big] \cdot
  \Dim \, \cdot \, \frac{\distrnorm[\big]{\funh{\ho} -
      \qfunstarh{\ho}}{\ho}^2} {\distrnorm{\qfunstarh{\ho}}{\ho}^2}
  \cdot \distrnorm{\funnewh{}}{\ho}^2,
\end{align*}
whence
\begin{align*}
  \frac{\distrnorm[\big]{ \,\big( \MOppih{\ho-1}{\policy} - \,
      \MOpstarh{\ho-1} \big) \, \gfunh{} \; }
    {\ho-1}}{\distrnorm{\funnewh{}}{\ho}} \; \leq \;
  \sqrt{\Exp_{\policystar} \big[
      \MyCurveTwosq{\ho}{\Stateh{\ho}} \big] \cdot \Dim } \; \cdot \,
  \frac{\distrnorm[\big]{\funh{\ho} - \qfunstarh{\ho}}{\ho}}
       {\distrnorm{\qfunstarh{\ho}}{\ho}} \; = \;
       \metrich{\ho}(\funh{\ho}, \, \qfunstarh{\ho}) \, ,
\end{align*}
where the metric $\metrich{\ho}$ is given by
equation~\eqref{eq:def_metric_lin}.  Consequently, the
bound~\eqref{EqnPseudoCondition} holds, as claimed
in~\Cref{lemma:lin_cond}(a).


\subsection{Proof of Lemma~\ref{lemma:lin_cond}(b)}
\label{sec:lin_fun_star_smooth_BellOpstar}

We now show that the condition~\ref{eq:def_smooth_BellOpstar} holds,
as claimed in part (b) of the lemma.  From the definition of the
Bellman (optimal) operator $\BellOpstar$, we have
\begin{align*}
\BellOpstarh{\ho} \funh{\ho+1} = \rewardh{\ho} + \MOppih{\ho}{\policy}
\funh{\ho+1} \qquad \mbox{and} \qquad \BellOpstarh{\ho}
\qfunstarh{\ho+1} = \rewardh{\ho} + \MOpstarh{\ho} \qfunstarh{\ho+1}
\, ,
\end{align*}
where we have adopted the shorthand $\MOpstarh{\ho} =
\MOppih{\ho}{\policystar}$, and exploited the greediness of the
policies $\policy$ and $\policystar$ with respect to the functions
$\fun$ and $\Qfunstar$, respectively. Subtracting these two equations
yields
\begin{align*}
\BellOpstarh{\ho} \funh{\ho+1} - \BellOpstarh{\ho} \qfunstarh{\ho+1} &
\; = \; \MOpstarh{\ho} \, \big(\funh {\ho+1} - \qfunstarh{\ho+1} \big)
+ \, \big(\MOppih{\ho}{\policy} - \MOpstarh{\ho}\big) \, \funh{\ho+1},
\end{align*}
from which an application of the triangle inequality yields
\begin{align}
\label{eq:lin_cond_BellOpstar0}
\distrnorm[\big]{\BellOpstarh{\ho} \funh{\ho+1} - \BellOpstarh{\ho}
  \qfunstarh{\ho+1}}{\ho} & \; \leq \; \underbrace{\distrnorm[\big]{
    \MOpstarh{\ho} \, \big(\funh {\ho+1} - \qfunstarh{\ho+1}
    \big)}{\ho}}_{\Term_{1}} + \, \underbrace{
  \distrnorm[\big]{\big(\MOppih{\ho}{\policy} - \MOpstarh{\ho} \big)
    \, \funh{\ho+1}}{\ho}} _{\Term_{2}} \, .
\end{align}
Consequently, we have reduced the problem to bounding the two terms
$\Term_{1}$ and $\Term_{2}$.

We first focus on term $\Term_{1}$.  As shown in~\Cref{AppOccCase}, the
stability condition~\eqref{EqnPseudoStable} holds with parameter
$\Radphistar = 1$.  As a consequence, the quantity $\Term_{1}$ can be
bounded as
\begin{subequations}
\begin{align}
  \label{eq:lin_cond_BellOpstar1}
  \Term_{1} \; = \; \distrnorm[\big]{ \MOpstarh{\ho} \, \big(\funh
    {\ho+1} - \qfunstarh{\ho+1} \big)}{\ho} \; \leq \;
  \distrnorm[\big]{\funh {\ho+1} - \qfunstarh{\ho+1}}{\ho+1} \, .
\end{align}

As for the term~$\Term_{2}$, we can show that it is second-order with
respect to the function difference $\distrnorm[\big]{\funh{\ho+1} -
  \qfunstarh{\ho+1}}{\ho+1}$, and therefore is negligible when
$\funh{\ho+1}$ is sufficiently close to $\qfunstarh{\ho+1}$.  Note
that for any $(\state, \action) \in \StateSp \times \ActionSp$, we can
write
\begin{align*}
  & \abs[\big]{\big( (\MOppih{\ho}{\policy} - \MOpstarh{\ho} ) \,
    \funh{\ho+1} \big) (\state, \action)} \\ & \; = \; \abs[\Big]{
    \Exp \big[ \funh{\ho+1} (\Stateh{\ho+1},
      \policyh{\ho+1}(\Stateh{\ho+1})) - \funh{\ho+1}(\Stateh{\ho+1},
      \policystarh{\ho+1}(\Stateh{\ho+1})) \bigm| \Stateh{\ho} =
      \state, \Actionh{\ho} = \action \big]} \\ & \; \leq \; \Exp
  \Big[ \abs[\big]{ \funh{\ho+1} (\Stateh{\ho+1},
      \policyh{\ho+1}(\Stateh{\ho+1})) - \funh{\ho+1}(\Stateh{\ho+1},
      \policystarh{\ho+1}(\Stateh{\ho+1}))} \Bigm| \Stateh{\ho} =
    \state, \Actionh{\ho} = \action \Big] \, .
\end{align*}
The curvature property~\ref{eq:curv_2} ensures that
\begin{align*}
 \abs[\big]{\big( (\MOppih{\ho}{\policy} - \MOpstarh{\ho} ) \,
   \funh{\ho+1} \big) (\state, \action)} & \; \leq \; \Exp \big[
   \MyCurveTwo{\ho+1}{\Stateh{\ho+1}} \, \bigm| \Stateh{\ho} = \state,
   \Actionh{\ho} = \action \big] \cdot \sqrt{\Dim} \, \cdot \,
 \frac{\distrnorm[\big]{\funh{\ho+1} - \qfunstarh{\ho+1}}
   {\ho+1}^2}{\distrnorm[\big]{\qfunstarh{\ho+1}}{\ho+1} }
\end{align*}
for each state-action pair $(\state, \action) \in \StateSp \times
\ActionSp$.  Applying the Cauchy--Schwarz inequality yields
\begin{align*}
  \Term_{2} = \distrnorm[\big]{\big(\MOppih{\ho}{\policy} -
    \MOpstarh{\ho} \big) \, \funh{\ho+1}}{\ho} & \; \leq \;
  \sqrt{\Dim} \; \CurveNorm{\MyCurve{\ho+1}}{\ho+1} \, \cdot \,
  \frac{\distrnorm[\big]{\funh{\ho+1} -
      \qfunstarh{\ho+1}}{\ho+1}^2}{\distrnorm[\big]{\qfunstarh
      {\ho+1}}{\ho+1} } \\ & \; = \; \metrich{\ho+1}\big(\funh{\ho+1},
  \, \qfunstarh{\ho+1} \big) \cdot \distrnorm[\big]{\funh{\ho+1} -
    \qfunstarh{\ho+1}}{\ho+1} \, ,
\end{align*}
where we have used the definition of metric
$\metrich{\ho}\big(\funh{\ho+1}, \, \qfunstarh{\ho+1} \big)$ in
equation~\eqref{eq:def_metric_lin}.  Under the condition
$\metrich{\ho+1}\big(\funh{\ho+1}, \qfunstarh{\ho+1}\big) \leq
\radius_{\ho+1} \leq \frac{1}{2} \, (\Ho-\ho)^{-1} (1+\log\Ho)^{-1}$,
we have
\begin{align}
  \label{eq:lin_cond_BellOpstar2}
  \Term_{2} & \; \leq \; \frac{1}{2 (\Ho-\ho) (1+\log\Ho)} \cdot
  \distrnorm[\big]{\funh{\ho+1} - \qfunstarh{\ho+1}}{\ho+1} \, .
\end{align}
\end{subequations}

Combining the bounds~\eqref{eq:lin_cond_BellOpstar1} and
\eqref{eq:lin_cond_BellOpstar2} with inequality
\eqref{eq:lin_cond_BellOpstar0} yields
\begin{align}
  \distrnorm[\big]{\BellOpstarh{\ho} \funh{\ho+1} - \BellOpstarh{\ho}
    \qfunstarh{\ho+1}}{\ho} & \; \leq \; \Curvstar{\ho} \cdot
  \distrnorm[\big]{\funh{\ho+1} - \qfunstarh{\ho+1} }{\ho+1}
  \label{eq:def_smooth_BellOpstar_app}
\end{align}
where $\Curvstar{\ho} = 1 + \frac{1}{2} \, (\Ho-\ho)^{-1}
(1+\log\Ho)^{-1}$.  It then follows that
\begin{align*}
  \Curvstarfun{\ho}{\honew} = \Curvstar{\ho} \Curvstar{\ho+1} \ldots
  \Curvstar{\honew-1} & = \prod_{\honewnew=\ho}^{\honew-1} \Big\{ 1 +
  \frac{1}{2(\Ho-\honewnew)(1+\log\Ho)} \Big\} \\ & \leq \; \exp \bigg\{
  \sum_{\honewnew=\ho}^{\honew-1} \frac{1}{2(\Ho-\honewnew)(1+\log\Ho)}
  \bigg\} \; \leq \; e \; \leq \; 3 \, ,
\end{align*}
which establishes claim~(b) in~\Cref{lemma:lin_cond}.


\subsection{Proof of Lemma~\ref{lemma:lin_cond}(c)}
Finally, we need to show that the smoothness
condition~\ref{eq:smooth_main} holds, as claimed in part (c).  In
order to do so, we make use of the curvature property~\ref{eq:curv_1}.
Consider the left-hand side of inequality~\ref{eq:smooth_main}.  It is
sufficient to show that
\begin{align*}
  \sup_{\begin{subarray}{c} \funh{} \in \DiffRKHS
      \\ \distrnorm{\funh{}}{\honew} > 0 \end{subarray}} \;
  \frac{\abs[\big]{ \, \Exp_{\policystar} \big[
        \big(\MOpstarhtoh{\ho}{\honew} \, \funh{} \, \big)
        (\Stateh{\ho}, \policystarh{\ho}(\Stateh{\ho})) -
        \big(\MOpstarhtoh{\ho}{\honew} \, \funh{} \, \big) (
        \Stateh{\ho}, \policyh{\ho}(\Stateh{\ho})) \big] }}
       {\distrnorm{\funh{\,}}{\honew}} & \; \leq \; \DDiffOpthree(\ho)
       \, \cdot \, \DDiffOpfour(\ho, \honew) \, ,
\end{align*}
where $\DDiffOpfour (\ho, \honew) \defn \; \sup_{\begin{subarray}{c}
    \funh{} \in \DiffRKHS \\ \distrnorm{\funh{}}{\honew} >
    0 \end{subarray}} \big\{ \distrnorm[\big]{
    \MOpstarhtoh{\ho}{\honew} \,
    \funh{\,}}{\ho} \, / \, \distrnorm{\funh{\,}}{\honew} \big\}$, and
\begin{align*}
  \DDiffOpthree (\ho) \defn \; \sup_{\begin{subarray}{c} \funnewh{}
      \in \DiffRKHS \\ \distrnorm{\funnewh{}}{\ho} > 0
  \end{subarray}} 
  \left\{ \frac{1}{\distrnorm{\funnewh{}}{\ho}} \abs[\big]{ \,
    \Exp_{\policystar} \big[ \funnewh{} (\Stateh{\ho},
      \policystarh{\ho}(\Stateh{\ho})) - \funnewh{}( \Stateh{\ho},
      \policyh{\ho}(\Stateh{\ho})) \big] } \right\}.
\end{align*}
Recall that the bound~\eqref{EqnPseudoStable} holds with radius
$\Radphistar = 1$, i.e.  $\distrnorm[\big]{ \MOpstarhtoh{\ho}{\honew}
  \, \funh{\,}}{\ho} \leq \distrnorm{\funh{}\,}{\honew}$.  Therefore,
we have $\DDiffOpfour(\ho, \honew) \leq 1$.  It remains to bound the
term $\DDiffOpthree (\ho)$.

Any $\funnewh{} \in \DiffRKHS$ has the representation
$\funnewh{}(\cdot) = \inprod{\Feature(\cdot)}{\bweight_{\funnewh{}}}$
for some vector $\bweight_{\funnewh{}} \in \Real^{\Dim}$, whence
\begin{align*}
  \abs[\big]{ \, \Exp_{\policystar} \big[ \funnewh{}
      (\Stateh{\ho}, \policystarh{\ho}(\Stateh{\ho})) - \funnewh{}(
      \Stateh{\ho}, \policyh{\ho}(\Stateh{\ho})) \big] } & \; = \;
  \abs[\Big]{ \, \inprod[\Big]{\Exp_{\policystar} \big[ \Feature
      (\Stateh{\ho}, \policystarh{\ho}(\Stateh{\ho})) - \Feature(
      \Stateh{\ho}, \policyh{\ho}(\Stateh{\ho})) \big]}
    {\bweight_{\funnewh{}}} \, } \\ & \; \leq \; \distrnorm[\big] {
    \Exp_{\policystar} \big[ \Feature (\Stateh{\ho},
      \policystarh{\ho} (\Stateh{\ho})) - \Feature ( \Stateh{\ho},
      \policyh{\ho}(\Stateh{\ho})) \big]}{\CovOp{\ho}^{-1}} \, \cdot
  \, \distrnorm{\bweight_{\funnewh{}}}{\CovOp{\ho}} \, .
\end{align*}
From the relation $\distrnorm{\bweight_{\funnewh{}}}{\CovOp{\ho}} =
\distrnorm{\funnewh{}}{\ho}$, we have
\begin{align*}
  \DDiffOpthree(\ho) \; \leq \; \distrnorm[\big] { \Exp_{
      \policystar} \big[ \Feature (\Stateh{\ho}, \policystarh{\ho}
      (\Stateh{\ho})) - \Feature ( \Stateh{\ho},
      \policyh{\ho}(\Stateh{\ho})) \big]}{\CovOp{\ho}^{-1}} \, .
\end{align*}
The curvature property~\ref{eq:curv_1} ensures that
\begin{align*}
  \distrnorm[\big]{ \Feature(\Stateh{\ho},
    \policyh{\ho}(\Stateh{\ho})) - \Feature(\Stateh{\ho},
    \policystarh{\ho}(\Stateh{\ho})) }{\CovOp{\ho}^{-1}} \; \leq \;
  \MyCurveTwo{\ho}{\Stateh{\ho}} \; \sqrt{\Dim} \, \cdot \,
  \frac{\distrnorm[\big]{\funh{\ho} - \qfunstarh{\ho}}{\ho}}
       {\distrnorm{\qfunstarh{\ho}}{\ho}} \; .
\end{align*}
Applying the Cauchy--Schwarz inequality yields
\begin{align*}
  \DDiffOpthree(\ho) \; \leq \; \Exp_{\policystar} \Big[
    \distrnorm[\big] { \Feature (\Stateh{\ho}, \policystarh{\ho}
      (\Stateh{\ho})) - \Feature ( \Stateh{\ho},
      \policyh{\ho}(\Stateh{\ho}))}{\CovOp{\ho}^{-1}}^2
    \Big]^{\frac{1}{2}}
  & \; \leq \; \sqrt{\Dim \cdot \Exp_{\policystar} \big[
      \MyCurve{\ho}^2(\Stateh{\ho}) \big]} \, \cdot \,
  \frac{\distrnorm[\big]{\funh{\ho} -
      \qfunstarh{\ho}}{\ho}}{\distrnorm{\qfunstarh{\ho}}{\ho}} \; .
\end{align*}
Putting together the pieces, we conclude that the stability
condition~\ref{eq:smooth_main} holds with parameter
$\Curvhtoh{\ho}{\honew} \leq \sqrt{\Dim} \;
\CurveNorm{\MyCurve{\ho}}{\ho}$, as claimed in~\Cref{lemma:lin_cond}.


\section{Proof of corollaries}

In this appendix, we prove our two corollaries about ridge-based FQI
in both off-line (\Cref{CorOffRidge}, proved
in~\Cref{SecProofCorOffRidge}) and on-line settings (\Cref{CorOnRidge}
proved in~\Cref{SecProofCorOnRidge}).

  
\subsection{Proof of Corollary~\ref{CorOffRidge}}
\label{SecProofCorOffRidge}

We begin with our result on ridge-based FQI in the off-line setting.

\subsubsection{Main argument}

At a high-level, we prove~\Cref{CorOffRidge} by specifying choices of
regularization parameters $\{\ridgeh{\ho} \}_{\ho=1}^{\Ho-1}$, along
with lower bounds on the sample size $\numobs$ at each stage $\ho$,
such that the ridge regression estimates $(\qfunhath{1}, \ldots,
\qfunhath{\Ho})$ satisfy the bounds
\begin{align*}
  \distrnorm[\big]{ \BellOpstarh{\ho} \qfunhath {\ho+1} -
    \qfunhath{\ho}}{\ho} \; \leq \BellErrh{\ho}
\end{align*}
where $\BellErrh{\ho}$ was defined in equation~\eqref{EqnOffRidge}.

Moving recursively backwards from the terminal stage $\Ho$, we can
define the radii
\begin{align*}
  \newradhath{\ho} = \norm{\bweight_{\rewardh{\ho}}}_2 + \norm
             {\bweighthat_{\ho+1}}_2,
\end{align*}
where the vectors $\bweight_{\rewardh{\ho}}$ and $\bweighthat_{\ho+1}
\in \Real^{\Dim}$ represent the linear coefficients associated with
the reward function $\rewardh{\ho}(\state) =
\inprod{\Feature(\state)}{\bweight_{\rewardh{\ho}}}$ and stage $\ho +
1$ value function estimate $\qfunhath{\ho+1}(\state) =
\inprod{\Feature(\state)}{\bweighthat_{\ho+1}}$.  We also recall the
definition~\eqref{eq:def_stderrdata} of the conditional variances
$\stderrhatdatah{\ho}^2(\funh{\,})$.  Throughout the following, we use
$c, c', C$ etc. to denote universal constants.

With this set-up, we claim that if, for some failure probability
$\errprob \in (0,1)$, if the sample sizes satisfy the lower bounds
\begin{subequations}
  \begin{align}
\label{EqnHitchLower}    
\numobs \geq \plaincon \, \big\{ \newradhath{\ho} \, \big/
\stderrhatdatah{\ho}(\qfunhath{\ho+1}) \big\}^2 \log(\Dim/\errprob),
  \end{align}
and we take the regularization parameters
\begin{align}
\label{EqnHitchReg}  
  \ridgeh{\ho} = c' \big\{\stderrhatdatah{\ho}(\qfunhath{\ho+1}) \, /
  \, \newradhath{\ho} \big\}^2 (\Dim/\numobs) \log(\Dim/\errprob),
  \end{align}
\end{subequations}
then the Bellman residuals satisfy the upper bounds
\begin{align}
  \label{eq:lin_BellErrtilh}
  \distrnorm[\big]{ \BellOpstarh{\ho} \qfunhath {\ho+1} -
    \qfunhath{\ho}}{\ho} \; \leq \; \PlainCon \; \norm[\big]{
    \CovOp{\ho} ^{\frac{1}{2}} \, ( \CovOpdata{\ho} + \ridgeh{\ho}
    \IdMt )^{-\frac{1}{2}}}_2 \;
  \stderrhatdatah{\ho}\big(\qfunhath{\ho+1} \big) \sqrt{\frac{\Dim \;
      \log(\Dim/\errprob)}{\numobs}} \; ,
\end{align}
with probability at least $1 - \delta$.

We now turn to the proof of this claim.  For notational convenience,
we introduce the (squared) norm
\begin{align*}
  \distrnorm{\funh{\,}}{\ho, \, \Data}^2 \defn
  \frac{1}{\abs{\Data_{\ho}}} \sum_{\Data_{\ho}}
  \funh{}^2(\stateh{\ho, \, i}, \actionh{\ho, \, i}) + \ridgeh{\ho}
  \distrnorm{\bweight}{2}^2 \qquad \mbox{for any function $\funh{} =
    \inprod{\Feature(\cdot)}{\bweight} \in \RKHS$}\, .
\end{align*}
By construction, we have $\distrnorm{\funh{\,}}{\ho, \, \Data}^2 =
\bweight^{\top} \big(\CovOpdata{\ho} + \ridgeh{\ho} \IdMt \big) \,
\bweight$, where the empirical covariance matrix~$\CovOpdata{\ho}$ was
previously defined~\eqref{eq:def_CovOpdata}.  Now consider the
inequality
\begin{align}
  \label{eq:lin_criteq}
  \sqrt{\frac{\log(\Dim/\errprob)}{ \eighat_{\min} + \delcrith{}^2}}
  \; \leq \; \frac{\sqrt{ \numobs } \;
    \newradhath{\ho}}{\stderrdatah{\ho} \big(\qfunhath{\ho+1}\big)} \,
  \delcrith{}, \tag{CI}
\end{align}
in terms of the scalar $\delcrith{} > 0$, where $\eighat_{\min}$
denotes the minimum eigenvalue of the empirical covariance matrix.  We
let $\delcrith{\ho} > 0$ be the smallest positive solution to
inequality~\eqref{eq:lin_criteq}; as we discuss in the sequel, this
solution always exists.  With this set-up, let us state the key
auxiliary result in our proof:
\begin{lemma}
\label{lemma:ridge}
Given a sample size $\numobs$ satisfying the lower
bound~\eqref{EqnHitchLower}, suppose that we implement FQI using ridge
regression with penalties $\ridgeh{\ho} \geq \delcrith{\ho}^2$.  Then
we have the bound
\begin{align}
  \label{eq:lemma_ridge}
  \distrnorm[\big]{\BellOpstarh{\ho} \qfunhath {\ho+1} -
    \qfunhath{\ho}}{\ho, \, \Data}^2 \; \leq \; \PlainCon \,
  \newradhath{\ho}^2 \; \big\{ \delcrith{\ho}^2 \, + \, \ridgeh{\ho}
  \big\}
\end{align}
with probability at least $1 - \errprob$.
\end{lemma}

Suppose that the smallest eigenvalue $\eighat_{\min}$ is at least the
order of $\Dim^{-1}$. Then the critical radius $\delcrith{\ho}$
satisfies an upper bound of the form
\begin{align*}
  \delcrith{\ho} \; \leq \; c \, \{\stderrdatah{\ho}(\qfunhath{\ho+1})
  \, / \, \newradhath{\ho} \} \sqrt{(\Dim/\numobs)
    \log(\Dim/\errprob)} \; \nfed \; \delcrittildeh{\ho} \, .
\end{align*}
By properly tuning the regularization parameter $\ridgeh{\ho}$ so that
$\ridgeh{\ho} \asymp \delcrittildeh{\ho}^2$, we can ensure that
\begin{align*}
  \distrnorm[\big]{\BellOpstarh{\ho} \qfunhath {\ho+1} -
    \qfunhath{\ho}}{\ho, \, \Data} \; \leq \; c'
  \stderrdatah{\ho}(\qfunhath{\ho+1}) \sqrt{(\Dim/\numobs)
    \log(\Dim/\errprob)} \, .
\end{align*}
Combining this bound with the relation
\begin{align*}
  \distrnorm[\big]{\BellOpstarh{\ho} \qfunhath{\ho+1} -
    \qfunhath{\ho}}{\ho} \; \leq \; \norm[\big]{ \CovOp{\ho}
    ^{\frac{1}{2}} \, ( \CovOpdata{\ho} + \ridgeh{\ho} \IdMt
    )^{-\frac{1}{2}}}_2 \cdot \distrnorm[\big]{\BellOpstarh{\ho}
    \qfunhath {\ho+1} - \qfunhath{\ho}}{\ho, \, \Data}\,,
\end{align*}
yields the claimed inequality~\eqref{eq:lin_BellErrtilh}. \\
  
\noindent It remains to prove~\Cref{lemma:ridge}.


\subsubsection{Proof of Lemma~\ref{lemma:ridge}}

Our proof is based on two auxiliary results, which we begin
by stating.

\paragraph{Step 1:}
Define the random variable
\begin{align*}
  \supZ(\ridgeh{\ho}) & \; \defn \sup_{\funh{} \in \RKHS \, : \;
    \distrnorm{\funh{\,}}{\ho, \Data} \leq 1} ~
  \frac{1}{\abs{\Data_{\ho}}} \, \sum_{\Data _{\ho}} \,
  \funh{}(\stateh{\ho, \, i}, \, \actionh {\ho, \, i}) \; \Diff_{\ho,
    \, i},
\end{align*}
where $\Diff_{\ho, \, i} \defn \reward_{\ho, \, i} + \max_{\action \in
  \ActionSp} \, \qfunhath{\ho+1} (\statenewh{\ho, \, i}, \action) -
\big(\BellOpstarh{\ho} \qfunhath{\ho+1} \big) (\stateh{\ho, \, i}, \,
\actionh{\ho, \, i})$.  Our first step is to show that
\begin{subequations}
\begin{align}
\label{eq:lin_ridge0}
  \distrnorm[\big]{\BellOpstarh{\ho} \qfunhath {\ho+1} -
    \qfunhath{\ho}}{\ho, \, \Data}^2 & \; \leq \; 2 \;
  \distrnorm[\big]{\BellOpstarh{\ho} \qfunhath{\ho+1} -
    \qfunhath{\ho}}{\ho, \, \Data} \, \cdot \, Z(\ridgeh{\ho}) \, + \,
  \ridgeh{\ho} \, \newradhath{\ho}^2 \, ,
\end{align}

\paragraph{Step 2:}
We then apply a matrix-form Bernstein
inequality to derive a concentration bound on
$\supZ(\ridgeh{\ho})$. In particular, we claim that
\begin{align}
  \label{eq:lin_supZ}
  \supZ(\ridgeh{\ho}) & \; \leq \; \PlainCon' \, \newradhath{\ho} \,
  \delcrith{\ho}
\end{align}
\end{subequations}
with probability exceeding $1 - \errprob$.

The bound~\eqref{eq:lemma_ridge} on Bellman residual then follows from
combining inequalities~\eqref{eq:lin_ridge0} and~\eqref{eq:lin_supZ}
and solving the quadratic inequality with respect to
$\distrnorm[\big]{ \BellOpstarh{\ho} \qfunhath{\ho+1} -
  \qfunhath{\ho}}{\ho, \, \Data}$\,. \\

\noindent Let us now turn to the proofs of
inequalities~\eqref{eq:lin_ridge0} and~\eqref{eq:lin_supZ}.


\paragraph{Proof of inequality~\eqref{eq:lin_ridge0}:}

We can write $\qfunhath{\ho}(\cdot) = \inprod{\Feature(\cdot)}{
  \bweighthat_{\ho}}$ and $\BellOpstarh{\ho} \qfunhath{\ho+1}(\cdot) =
\inprod[\big]{\Feature(\cdot)} {\bweight_{\BellOpstarh{\ho}
    \qfunhath{\ho+1}}}$ for vectors $\bweighthat_{\ho}$ and
$\bweight_{\BellOpstarh{\ho} \qfunhath{\ho+1}}$.  Introducing the
shorthand $\Deltaweight \defn \bweighthat_{\ho} - \bweight
_{\BellOpstarh{\ho} \qfunhath{\ho+1}}$, and using the
definition~\eqref{eq:ridge} of the ridge estimate, we have
\begin{align}
\label{eq:lin_ridge2}
\Deltaweight = ( \CovOpdata{\ho} + \ridgeh{\ho} \, \IdMt )^{-1}
\bigg\{ \frac{1}{\abs{ \Data_{\ho}}} \sum_{\Data_{\ho}}
\Feature(\stateh{\ho, \, i}, \, \actionh{\ho, \, i}) \, \Diff_{\ho, \,
  i} \bigg\} - \ridgeh{\ho} \, ( \CovOpdata{\ho} + \ridgeh{\ho} \,
\IdMt )^{-1} \, \bweight_{ \BellOpstarh{\ho} \qfunhath{\ho+1}} \, .
\end{align}
Multiplying both sides of equation~\eqref{eq:lin_ridge2} by
$\Deltaweight^{\top} \, ( \CovOpdata{\ho} + \ridgeh{\ho} \, \IdMt )$
from the left yields
\begin{align}
\distrnorm[\big]{\BellOpstarh{\ho} \qfunhath {\ho+1} -
  \qfunhath{\ho}}{\ho, \, \Data}^2 & = \Deltaweight^{\top} (
\CovOpdata{\ho} + \ridgeh{\ho} \, \IdMt ) \, \Deltaweight \notag \\
& = \frac{1}{\abs{ \Data_{\ho}}} \sum_{\Data_{\ho}} \,
\big(\qfunhath{\ho} - \BellOpstarh{\ho}
\qfunhath{\ho+1}\big)(\stateh{\ho, \, i}, \, \actionh{\ho, \, i}) \,
\cdot \, \Diff_{\ho, \, i} \; - \; \ridgeh{\ho} \,
\inprod[\big]{\Deltaweight} {\bweight_{\BellOpstarh{\ho}
    \qfunhath{\ho+1}} } \label{eq:lin_ridge3} \, .
\end{align}
Using the definition of random variable $\supZ(\ridgeh{\ho})$, 
equation~\eqref{eq:lin_ridge3} implies that
\begin{align}
  \distrnorm[\big]{\BellOpstarh{\ho} \qfunhath {\ho+1} -
    \qfunhath{\ho}}{\ho, \, \Data}^2 \; & \leq \;
  \distrnorm[\big]{\BellOpstarh{\ho} \qfunhath{\ho+1} -
    \qfunhath{\ho}}{\ho, \, \Data} \, \cdot \, \supZ(\ridgeh{\ho}) +
  \frac{\ridgeh{\ho}}{2} \, \Big\{ \norm{\Deltaweight}_2^2 + \norm[\big]
       {\bweight_{\BellOpstarh{\ho} \qfunhath {\ho+1}}}_2^2 \Big\}
       \notag \\
       \label{eq:lin_ridge4}       
& \leq \; \distrnorm[\big]{\BellOpstarh{\ho} \qfunhath{\ho+1} -
         \qfunhath{\ho}}{\ho, \, \Data} \, \cdot \,
       \supZ(\ridgeh{\ho}) + \frac{1}{2} \,
       \distrnorm[\big]{\BellOpstarh{\ho} \qfunhath{\ho+1} -
         \qfunhath{\ho}}{\ho, \, \Data}^2 + \frac{1}{2} \,
       \ridgeh{\ho} \, \newradhath{\ho}^2 \, .
\end{align}
The last inequality follows from the bounds \mbox{ $\ridgeh{\ho}
  \distrnorm{\Deltaweight}{2}^2 \leq \distrnorm[\big]{\BellOpstarh{\ho}
    \qfunhath{\ho+1} - \qfunhath{\ho}}{\ho, \, \Data}^2$} and
{$\norm[\big] {\bweight_{\BellOpstarh{\ho} \qfunhath {\ho+1}}}_2
  \leq \norm{\bweight_{\rewardh{\ho}}}_2 +
  \norm{\bweighthat_{\ho+1}}_2 \leq \newradhath{\ho}$.}  Simplifying
inequality~\eqref{eq:lin_ridge4} yields
inequality~\eqref{eq:lin_ridge0}.


\paragraph{Proof of inequality~\eqref{eq:lin_supZ}:}

We prove this claim via matrix-form Bernstein inequality.  Introducing
the shorthand $\Featurenew (\state, \action) \defn (\CovOpdata{\ho} +
\ridgeh{\ho} \, \IdMt )^{- \frac{1}{2}} \, \Feature(\state, \action)$,
the property $\distrnorm{\Feature(\state, \action)}{2} \leq 1$ then
implies $\distrnorm{\Featurenew( \state, \action)}{2} \leq
(\eighat_{\min} + \ridgeh{\ho})^{-\frac{1}{2}}$ for any $( \state,
\action) \in \StateSp \times \ActionSp$.  Notice that
\begin{align*}
\supZ(\ridgeh{\ho}) = \distrnorm[\bigg]{\, \frac{1}{\abs{\Data_{\ho}}}
  \sum_{\Data_{\ho}} \Featurenew (\stateh{\ho,\,i}, \actionh{\ho,\,i})
  \, \Diff_{\ho, \, i} \, }{2} \, .
\end{align*}
It then follows from the bound $\abs{\Diff_{\ho, \, i}} \leq 2 \,
\newradhath{\ho}$ that
\begin{align*}
\distrnorm[\big]{\Featurenew(\stateh{\ho, \, i}, \, \actionh{\ho, \,
    i}) \, \Diff_{\ho, \, i}} {2} \leq 2 \, \newradhath{\ho} \, /
\sqrt{\eighat_{\min} + \ridgeh{\ho}}.
\end{align*}
Moreover, since $\Exp\big[ \Diff_{\ho, \, i}^2 \bigm| \stateh{\ho, \,
    i}, \actionh{\ho, \, i} \big] = \Var\big[ \max_{\action \in
    \ActionSp} \qfunhath{\ho+1} (\statenewh{\ho, \, i}, \action)
  \bigm| \stateh{\ho, \, i}, \actionh{\ho, \, i} \big]$, we find that
the second order moment satisfies
\begin{align*}
  \frac{1}{\abs{\Data_{\ho}}} \sum_{\Data_{\ho}} \, \Exp\Big[
    \distrnorm[\big]{\Featurenew(\stateh{\ho, \, i}, \, \actionh{\ho,
        \, i}) \, \Diff_{\ho, \, i}} {2}^2 \Bigm| \stateh{\ho, \, i},
    \, \actionh{\ho, \, i} \Big] \; \leq \;
  \frac{\stderrdatah{\ho}^2(\qfunhath {\ho+1})}{\eighat_{\min} +
    \ridgeh{\ho}} \, ,
\end{align*}
where the conditional variance $\stderrdatah{\ho}^2(\qfunhath{\ho+1})$
is given by definition~\eqref{eq:def_stderrdata}.  A standard matrix
Bernstein inequality (see Theorem~6.17 in the
book~\cite{wainwright2019high}) then implies that
\begin{align*}
  \supZ(\ridgeh{\ho}) & \; \leq \; \Const{1} \Bigg\{
  \stderrdatah{\ho}(\qfunhath {\ho+1})
  \sqrt{\frac{\log(\Dim/\errprob)} { \numobs \, (\eighat_{\min} +
      \ridgeh{\ho} )}} \, + \, \newradhath{\ho} \,
  \frac{\log(\Dim/\errprob)} { \numobs \sqrt{\eighat_{\min} +
      \ridgeh{\ho}}} \Bigg\} \; \leq \; \Const{2} \, \newradhath{\ho}
  \, \delcrith{\ho}
\end{align*}
with probability exceeding $1 - \errprob$, which establishes
inequality~\eqref{eq:lin_supZ}.  Here the last inequality follows from
the critical inequality~\eqref{eq:lin_criteq} and the sample size
condition $\numobs \, \geq \, \const{2} \, \big\{ \newradhath{\ho}^2
\, \big/ \stderrdatah{\ho}^2(\qfunhath {\ho+1}) \big\} \,
\log(\Dim/\errprob)$\,.


\subsection{Proof of Corollary~\ref{CorOnRidge}}
\label{SecProofCorOnRidge}

In \PhaseOne~of pure exploration, the cumulative regret is always
bounded from above by $\Tsafe \cdot \Ho$. During \PhaseTwo~of
fine-tuning, we let $\policyhat^k$ be the policy employed in the
rounds \mbox{ $\!\Tsafe 2^k\!+\!1$, $\Tsafe 2^k\!+\!2$, \!$\ldots$,}
$\Tsafe 2^{k+1}$, which is determined by the estimate
$\qfunhat^{(\Tsafe 2^k)}$ calculated at the end of the $(\Tsafe
2^k)$-th round.  To estimate the regret, we consider the decomposition
\begin{align*}
\sum_{t=\Tsafe + 1}^T \big\{ \valuescalar(\policystar) -
\valuescalar(\policyhat^{(t)}) \big\} \leq \sum_{k=0}^{K-1} \sum_{t =
  \Tsafe 2^k}^{\Tsafe 2^{k+1}} \big\{ \valuescalar(\policystar) -
\valuescalar(\policyhat^{(t)}) \big\} = \sum_{k=0}^{K-1} \Tsafe \, 2^k
\big\{ \valuescalar(\policystar) - \valuescalar(\policyhat^{k}) \big\}
\, .
\end{align*}

We leverage our bound~\eqref{eq:sub_opt_offline_order} for off-line RL
in~\Cref{SecFQIFast} to control the value sub-optimality
$\valuescalar(\policystar) - \valuescalar(\policyhat^{k})$.  Recall
that the policy $\policyhat^{k}$ is derived from i.i.d. trajectories
collected from the rounds $\Tsafe \, 2^{k-1} + 1, \Tsafe \, 2^{k-1} +
2, \ldots, \Tsafe \, 2^{k}$. We divide those $\Tsafe \, 2^{k-1}$
trajectories into $\Ho-1$ equal shares and use each share to conduct
estimation in one iteration of the FQI procedure.  This subsampling
technique ensures the independence of samples used in different
iterations. It is primarily adopted for the sake of convenience (to
keep the explanations concise) and is not essential in general.  It
follows from inequality~\eqref{eq:sub_opt_offline_order} that the
bound
\begin{align*}
  \valuescalar(\policystar) - \valuescalar(\policyhat^{k}) \; \leq \;
  \PlainCon \; \frac{\Dim\sqrt{\Dim} \; \Ho^4}{\Tsafe \, 2^k} \,
  \log(\Dim \Ho K / \errprob)
\end{align*}
holds uniformly for indices $k = 0, 1,\ldots, K-1$ with a probability
exceeding $1 - \errprob$.

Putting together the pieces, we arrive at
\begin{align*}
\regret(T) \leq \Tsafe \cdot \Ho + \PlainCon \; \Dim\sqrt{\Dim} \;
\Ho^4 \, K \, \log(\Dim \Ho K / \errprob) \, .
\end{align*}
We then derive the regret bound~\eqref{eq:regret} by noticing that $K
= \bigO(\log T)$.


\subsection{Scaling of bounds in off-line and on-line RL}
\label{sec:lin_offline_order}

In this part, we provide detailed explanations regarding the scaling,
in terms of dimension $d$ and horizon $\Ho$, of the bounds that arise
in the discussion of off-line RL from~\Cref{sec:thm_imp_offline}.

\subsubsection{Effect of mild covariate shift \yaqidone}
\label{sec:lin_offline_order_1}
  
In this section, we justify the bound~\eqref{eq:sub_opt_offline_order}
stated following~\Cref{CorOffRidge} in~\Cref{SecFQIFast}.  For rewards
taking values in $[0,1]$, it is reasonable to assume that the
$Q$-functions~$\qfunstarh{\ho}$ satisfy the bounds
$\distrnorm{\qfunstarh{\ho}}{\ho} \asymp \Ho - \ho + 1$, and moreover
that the conditional variance
$\stderrhatdatah{\ho}\big(\qfunhath{\ho+1} \big) \asymp
\sqrt{\Ho-\ho}$. (We will provide a detailed justification for this
argument later.)  Furthermore, suppose the covariate shift is mild
such that $\norm[\big]{ \CovOp{\ho}^{\frac{1}{2}} \, ( \CovOpdata{\ho}
  + \ridgeh{\ho} \IdMt)^{-\frac{1}{2}}}_2 \leq \PlainCon'$, and the
curvature parameter $\MyCurveTwo{\ho}{\state}$ in
inequalities~\ref{eq:curv_1}~and~\ref{eq:curv_2} satisfies
$\CurveNorm{\MyCurve{\ho}}{\ho} \leq \PlainCon'$.

Provided with these conditions, we apply inequality
\eqref{eq:lin_BellErrtilh} from our preceding analysis, and then set
the Bellman residual parameter
\begin{align}
  \label{eq:def_lin_BellErrh}
  \BellErrh{\ho} \; \defn \; \PlainCon'' \; \sqrt{\frac{\Dim \,
      (\Ho-\ho) \log(\Dim \Ho/\errprob)}{\numobs}}
\end{align}
for a suitably chosen constant $\PlainCon'' > 0$.  Then the sequence
$\bbellerr = (\BellErrh{1}, \ldots, \BellErrh{\Ho-1},
\BellErrh{\Ho}=0)$ is regular since $\sqrt{x+1} \geq \frac{1}{x}
\sum_{i=1}^x \sqrt{i}$ for any integer $x \geq 1$.

Given the Bellman residual $\BellErrh{\ho}$ defined in
equation~\eqref{eq:def_lin_BellErrh}, the
condition~\eqref{eq:cor_n_cond} from~\Cref{PropLin} becomes
\mbox{$\numobs \; \geq \; \plaincon \, \cdot \, \Dim^2 \, \Ho^3 \, (1
  + \log \Ho)^2 \, \log(\Dim \Ho / \errprob)$,} and the
bound~\eqref{EqnOffRidge} reduces to
\begin{align*}
  \valuescalar(\policystar) - \valuescalar(\policyhat) \; \leq \;
  \frac{\PlainCon \; \Dim^{\frac{3}{2}} \Ho^3
    \log(\Dim\Ho/\errprob)}{\numobs} \, ,
\end{align*}
as claimed. 


\paragraph{Justification of the bound $\stderrhatdatah{\ho}
\big(\qfunhath{\ho+1} \big) \asymp \sqrt{\Ho-\ho}$\,:}

When the estimate~$\qfunhat$ is relatively close to the optimal
$Q$-function $\Qfunstar$, the conditional
variance~$\stderrdatah{\ho}^2(\qfunhath{\ho+1})$ can be bounded as
follows:
\begin{align*}
  \stderrdatah{\ho}^2(\qfunhath{\ho+1}) \, \asymp \, \stderrh{\ho}^2
  \qquad \mbox{with~} \stderrh{\ho}^2 \defn \Exp_{
    \policystar}\Big[ \max_{\action \in \ActionSp} \big[
      \qfunstarh{\ho+1}(\Stateh{\ho+1}, \action) \bigm| \Stateh{\ho},
      \Actionh{\ho} \big] \Big] \, .
\end{align*}
From the law of total variance, we have $\sum_{\honew=\ho} ^{\Ho-1}
\stderrh{\honew}^2 \; \leq \; \PlainCon \; (\Ho-\ho)^2$.  Therefore,
it is reasonable to consider $\stderrh{\ho} \asymp \sqrt{\Ho-\ho}$\,,
which further leads to the scaling~$\stderrhatdatah{\ho}
\big(\qfunhath{\ho+1} \big) \asymp \sqrt{\Ho-\ho}$.


\subsubsection{Comparing to known off-line bounds
\yaqidone}
\label{sec:lin_offline_order_2}

In this section, we derive inequality~\eqref{EqnZanette} based on the
results of Zanette et al.~\cite{zanette2021provable}; it gives the
conventional $1/\sqrt{\numobs}$ slow rate to which we compare.
Zanette et al.~\cite{zanette2021provable} proved upper bounds on a
pessimistic actor-critic scheme based on $d$-dimensional linear
function approximation.  Using our notation, Theorem 1 in their
paper~\cite{zanette2021provable} can be expressed as
\begin{align}
  \label{eq:lin_old}
  \valuescalar(\policystar) - \valuescalar(\policyhat) & \; \leq \; c
  \, \bigg\{ \frac{1}{\Ho} \sum_{\ho=1}^{\Ho-1} \,
  \sqrt{\featurebarh{\ho}\!\!\,^{\top} ( \CovOpdata{\ho} +
    \ridgeh{\ho} \IdMt )^{-1} \, \featurebarh{\ho}} \bigg\} \,
  \sqrt{\frac{\Dim \Ho^4}{\numobs}} \, ,
\end{align}
where the vector $\featurebarh{\ho}$ is given by $\featurebarh{\ho}
\defn \Exp_{\policystar}\big[ \Feature(\Stateh{\ho},
  \Actionh{\ho}) \big]$.

 We now consider the explicit dependence of this upper bound on
 dimension $\Dim$, horizon~$\Ho$ and sample size~$\numobs$.  The
 divergence term $\featurebarh{\ho}\!\!\,^{ \top} ( \CovOpdata{\ho} +
 \ridgeh{\ho} \IdMt )^{-1} \, \featurebarh{\ho}$ measures the
 conditioning of the regularized covariance matrix $( \CovOpdata{\ho}
 + \ridgeh{\ho} \IdMt )$ along a specific direction of
 $\featurebarh{\ho}$. When the feature mapping $\Feature$ operates
 within a $\Dim$-dimensional space, it is reasonable to assume that
 \begin{align*}
   \featurebarh{\ho}\!\!\,^{ \top} ( \CovOpdata{\ho} + \ridgeh{\ho}
   \IdMt )^{-1} \, \featurebarh{\ho} \; \leq \; \PlainCon' \; \Dim \,
   .
 \end{align*}
 The bound~\eqref{eq:lin_old} then reduces to $
 \valuescalar(\policystar) - \valuescalar(\policyhat) \, \leq \,
 \PlainCon \, {\Dim \Ho^2}/{\sqrt{\numobs}} $\,.  Regarding the
 dependence on horizon~$\Ho$, we conjecture that by incorporating the
 law of total variance in a more refined manner, it may be possible to
 further reduce the dependence by a factor of $\sqrt{\Ho}$.  Under
 these conditions, the bound takes the form \mbox{$\valuescalar(
   \policystar) - \valuescalar(\policyhat)$} \mbox{$\leq \, \PlainCon
   \, \Dim \sqrt{\Ho^3/ \numobs}$}.


\section{General guarantee for linear curvature}
\label{SecGenLinear}

In this section, we state and prove a general result under which the
curvature conditions~\ref{eq:curv_1} and \ref{eq:curv_2} hold.


\subsection{A general curvature guarantee}
\label{SecPropCurve}

For any state $\state \in \StateSp$, suppose that the feature set
$\Featureset(\state) = \{ \Feature(\state, \action) \mid \action \in
\ActionSp \} \subseteq \Real^{\Dim}$ can be described in the form
\begin{align*}
  \Featureset(\state) = \big\{ \bx \in \Real^{\Dim} \mid
  \constrain_{\state}(\bx) \leq \zerovec \big\} \, .
\end{align*}
where $\constrain_\state: \real^\Dim \rightarrow \real^{m_\state}$ is
an $m_\state$-vector of constraints, with $m_\state \leq \Dim$.

Given any $Q$-function estimate $\funh{\ho}: \StateSp \times \ActionSp
\rightarrow \Real$, the associated greedy policy $\policyh{\ho}$ is
characterized by
\begin{align}
  \label{eq:lin_opt0}
  \policyh{\ho}(\state) \in \argmax \big\{ \funh{\ho}(\state, \action)
  \bigm| \action \in \ActionSp ~~\mbox{with }
  \constrain_{\state}(\Feature(\state, \action)) \leq \zerovec \,
  \big\} \, .
\end{align}
We assume all the constraint functions $\constraini{\state, 1},
\ldots, \constraini{\state, \numcon}$ are strongly convex and twice
differentiable\footnote{To be precise, strong convexity and twice
differentiability are only required within a neighborhood around
$\Feature(\state, \policystarh{\ho}(\state))$ for the following
arguments to hold.}, so that the solution to the optimization problem
is unique and the greedy policy $\policyh{\ho}(\state)$ is
well-defined and deterministic.
For the sake of simplicity in our subsequent discussion, we omit the
dependence on state $\state$ and time index $\ho$, and instead use
$\Featureset$, $\constrain$, $\numcon$ in place of
$\Featureset(\state)$, $\constrain_{\state}$, $\numcon_{\state}$ and
use $\qfunstarh{}$, $\policystarh{}$, $\funh{}$, $\policyh{}$,
$\CovOp{}$ to represent $\qfunstarh{\ho}$, $\policystarh{\ho}$,
$\funh{\ho}$, $\policyh{\ho}$, $\CovOp{\ho}$ when the context is
clear.

It is worth noting that while the optimization formulation
\eqref{eq:lin_opt0} is originally defined within the action space
$\ActionSp$, we can transform it into a problem that operates in the
vector space $\Real^{\Dim}$.  Consider the vector representation
$\bweight \in \Real^{\Dim}$ of function $\funh{}(\cdot) =
\Feature(\cdot)^{\top} \bweight$.  By introducing a feature vector
$\Feature \defn \Feature(\state, \policyh{}(\state)) \in
\Real^{\Dim}$, we can equivalently reformulate problem
\eqref{eq:lin_opt0} as follows
\begin{align}
  \Feature \equiv \Feature(\state, \policyh{}(\state)) \quad = \quad
  \argmax_{\bx \in \Real^{\Dim}} & \quad \bx^{\top}
  \bweight \label{eq:def_lin_opt} \\ \mbox{subject to} & \quad
  \constrain(\bx) \leq \zerovec \, .  \notag
\end{align}
We define the vector $\Featurestar \defn \Feature(\state,
\policystarh{}(\state)) \in \Real^{\Dim}$ as the optimizer
corresponding to the optimal \mbox{$Q$-function} $\qfunstarh{}(\cdot)
= \Feature(\cdot)^{\top} \bweightstar$.  For simplicity, we assume all
the constraints are active, i.e.  the maxima are achieved at the
boundary of $\Featureset$ so that $\constrain(\Feature) =
\constrain(\Featurestar) = \zerovec$.

In our framework, we capture the ``curvature'' by using the following
two key ingredients: a local Hessian matrix $\Hess \in \Real^{\Dim
  \times \Dim}$ and a tangent space at the point $\Featurestar$.
\paragraph{Local Hessian matrix:}
Defining the Lagrangian $\Lag(\bx, \blambda) \defn
\inprod{\bx}{\bweightstar} - \blambda^{\top} \constrain(\bx)$, we let
       {$(\Featurestar, \blambdastar) \in \Real^{\Dim} \times
         \Real^{\numcon}$} be the saddle point of the problem
\begin{align*}
  \max_{\substack{\bx \in \Real^{\Dim} \\ \constrain(\bx) \leq
      \zerovec}} \min_{\substack{ \blambda \in \Real^{\numcon}
      \\ \blambda \geq \zerovec}} \quad \Lag(\bx, \blambda).
\end{align*}
We use the Lagrange multiplier $\blambdastar = (\lambdastari{1},
\lambdastari{2}, \ldots, \lambdastari{\numcon})^{\top} \in
\Real^{\numcon}$ to define the weighted sum
\begin{align}
  \label{eq:def_Hess}
  \Hess & \,\defn\; \sum_{i=1}^{\numcon} \; \lambdastari{i} \;
  \nabla^2 \constraini{i}(\Featurestar) \; \in \; \Real^{\Dim \times
    \Dim} \, ,
\end{align}
which is a positive definite matrix whenever $\blambdastar$ is
non-zero, given our strong convexity conditions on the constraint
functions.

\paragraph{Normal vectors and tangent space:}
Let $\subspace \subseteq \Real^{\Dim}$ be a linear subspace defined as
\begin{align*}
  \subspace \defn \Span[\Big]{\Hess^{-\frac{1}{2}} \nabla
    \constraini{1}(\Featurestar), \; \Hess^{-\frac{1}{2}} \nabla
    \constraini{2}(\Featurestar), \; \ldots, \; \Hess^{-\frac{1}{2}}
    \nabla \constraini{\numcon}(\Featurestar)} \, .
\end{align*}
The elements within the space $\subspace$ can be interpreted as normal
vectors that are perpendicular to the boundary (after a specific
linear transformation).  Let $\projG$ represent the projection onto
space~$\subspace$ under the Euclidean norm $\distrnorm{\cdot}{2}$.
More explicitly, we define $\projG$ as follows:
\begin{align*}
  \projG = \Hess^{-\frac{1}{2}} \,
  \nabla\constrain(\Featurestar)^{\top} \big[ \nabla
    \constrain(\Featurestar) \, \Hess^{-1} \nabla
    \constrain(\Featurestar)^{\top} \big]^{-1} \, \nabla
  \constrain(\Featurestar) \, \Hess^{-\frac{1}{2}} \, .
\end{align*}
We use the operator $(\IdMt - \projG)$ to denote the projection onto
the orthogonal complement of linear space $\subspace$, which can be
viewed as the projection onto the \emph{tangent space} of the boundary
at point~$\Featurestar$.  Intuitively, the tangent space contains all
possible directions in which one can tangentially pass through
$\Featurestar$ when moving along the boundary of the feature set
$\Featureset$.

\vspace{1.5em}

In addition to the Hessian and tangent space, we also need
characterizations of the smoothness of the boundary as shown below,
which are in general direct consequences of twice differentiability.
\vspace{-2em}
\paragraph{Smoothness condition of the boundary:}
We introduce a compact notation of the gradients
\begin{align*}
  \nabla \constrain(\Featurestar) & \; \defn \; \big[ \, \nabla
    \constraini{1}(\Featurestar), \; \nabla
    \constraini{2}(\Featurestar), \; \ldots, \; \nabla
    \constraini{\numcon}(\Featurestar) \, \big]^{\top} \in
  \Real^{\numcon \times \Dim} \, .
\end{align*}
When the constraint functions $\{\constraini{i}\}_{i=1}^{\numcon}$ are
twice differentiable, it follows from the definition of the Hessian
matrix $\Hess$ that $\big\{ \nabla \constrain(\bx) - \nabla
\constrain(\Featurestar) \big\}^{\top} \blambdastar = \Hess (\bx -
\Featurestar) + \smallo( \distrnorm{\bx - \Featurestar}{\Hess} )$.
\begin{subequations}
Therefore, there exists a neighborhood around the vector
$\Featurestar$ such that any point $\bx$ within it satisfies
\begin{align}
\label{eq:curv_smooth1}
\distrnorm[\Big]{\big\{ \nabla \constrain(\bx) - \nabla
  \constrain(\Featurestar) \big\}^{\top} \blambdastar - \Hess (\bx -
  \Featurestar) }{\Hess^{-1}} \leq \, \frac{1}{4} \, \distrnorm{\bx -
  \Featurestar}{\Hess} \, .
\end{align}
Furthermore, we use a parameter $\Lip > 0$ to characterize the
Lipschitz continuity of the gradient~$\nabla \constrain$. This means
that
\begin{align}
\label{eq:curv_smooth2}
\distrnorm[\Big]{\WMt^{-\frac{1}{2}} \big\{ \nabla \constrain(\bx) -
  \nabla \constrain(\Featurestar) \big\} \Hess^{-\frac{1}{2}}}{2} \leq
\; \Lip \cdot \distrnorm{\bx - \Featurestar}{\Hess} \, ,
\end{align}
where the matrix $\WMt \in \Real^{\numcon \times \numcon}$ is defined
as $\WMt \defn \nabla \constrain(\Featurestar) \, \Hess^{-1} \nabla
\constrain(\Featurestar)^{\top}$.  Note that the gradient Lipschitz
property is in general less restrictive than being twice
differentiable.
\end{subequations}

\vspace{1.5em}

We are now ready to present the exact formulations of
inequalities~\ref{eq:curv_1} and~\ref{eq:curv_2} within the specific
context we have established earlier.

\begin{proposition}
\label{PropCurve}
Suppose that $\distrnorm{\Feature - \Featurestar}{\Hess} \leq
\frac{\distrnorm{(\IdMt - \projG) \, \Hess^{-\frac{1}{2}}
    \CovOp{}^{-\frac{1}{2}}}{2}} {3 \Lip \,
  \distrnorm{\Hess^{-\frac{1}{2}} \CovOp{}^{-\frac{1}{2}}}{2}}$ and
$\distrnorm[\big]{\projG \, \Hess^{-\frac{1}{2}} (\bweight -
  \bweightstar)}{2} \leq \frac{1}{12 \Lip}$.  Then the
bounds~\ref{eq:curv_1}~and~\ref{eq:curv_2} hold with parameter
\begin{align}
  \label{eq:MyCurveTwo_lin}
  \MyCurveTwo{\ho}{\state} \; \defn \; \frac{5}{\sqrt{\Dim}} \;
  \distrnorm{\bweightstar}{\CovOp{}} \, \distrnorm[\big]{(\IdMt -
    \projG) \, \Hess^{-\frac{1}{2}} \CovOp{}^{-\frac{1}{2}}}{2}^2 \, .
\end{align}
Equivalently, it means that
\begin{subequations}
\begin{align}
  \distrnorm{\Feature - \Featurestar}{\CovOp{}^{-1}} \ & \leq
  \ \MyCurveTwo{\ho}{\state} \, \sqrt{\Dim} \, \cdot \,
  \frac{\distrnorm{\bweight - \bweightstar}{\CovOp{}}}
       {\distrnorm{\bweightstar}{\CovOp{}}} \qquad \mbox{and}
  \label{eq:curv_1_app} \\
  \bweight^{\top} (\Feature - \Featurestar) \ & \leq
  \ \MyCurveTwo{\ho}{\state} \, \sqrt{\Dim} \, \cdot \,
  \distrnorm{\bweightstar}{\CovOp{}} \, \cdot \, \bigg\{
  \frac{\distrnorm{\bweight - \bweightstar}{\CovOp{}}}
       {\distrnorm{\bweightstar}{\CovOp{}}} \bigg\}^2 \, .
  \label{eq:curv_2_app}
\end{align}
\end{subequations}
\end{proposition}
\noindent See~\Cref{SecProofPropCurve} for the proof. \\

Let us make some comments about~\Cref{PropCurve}.  We first observe
that the conditions under which the curvature conditions hold are
relatively mild, requiring only: (i) strong convexity constraint
functions $\{ \constraini{i} \}_{i=1}^{\numcon}$ around the point
$\Featurestar$, and (ii) twice differentiability of the constraint
function $\constrain$.  Thus, the guarantee of~\Cref{PropCurve}
applies to a fairly broad class of problems.

Second, in stating our result, we have defined the parameter
$\MyCurveTwo{\ho}{\state}$ from equation~\eqref{eq:MyCurveTwo_lin}
such that it is independent of the scaling of vector $\bweight$, and
typically independent of the dimension $\Dim$.
  \begin{itemize}
  \item First, suppose that we rescale the parameter vector
    $\bweight$. We redefine the objective function in
    optimization problem~\eqref{eq:lin_opt0} by doubling the vector
    $\bweightstar$ and setting $\bweighttil \defn 2 \bweightstar$.
    The new Lagrangian multiplier $\blambdatil$ then undergoes the
    rescaling $\blambdatil = 2 \blambdastar$, and leads to the a new
    Hessian matrix \mbox{$\widetilde{\Hess} = 2 \Hess$,} as in
    equation~\eqref{eq:def_Hess}.  Overall, the parameter
    $\MyCurveTwo{\ho}{\state}$ remains unchanged, as claimed.
    \item Regarding the dependence of parameter
      $\MyCurveTwo{\ho}{\state}$ on the dimension $\Dim$ of feature
      vector~$\Feature$, recall our boundedness condition
      $\sup_{(\state, \action) \in \StateSp \times \ActionSp}
      \distrnorm{\Feature(\state, \action)}{2} \leq 1$.  Under this
      bound, the eigenvalues of the covariance matrix
      \mbox{$\CovOp{\ho} = \Exp_{ \policystar}\big[
          \Feature(\Stateh{\ho}, \Actionh{\ho}) \,
          \Feature(\Stateh{\ho}, \Actionh{\ho})^{\top} \big]$} are of
      the order of $1/\Dim$, so that the norm
      $\distrnorm{\bweightstar}{\CovOp{}} =
      \sqrt{(\bweightstar)^{\top}\CovOp{}\,\bweightstar}$ scales as
      $1/\sqrt{\Dim}$, while the norm $\distrnorm[\big]{(\IdMt -
        \projG) \, \Hess^{-\frac{1}{2}} \CovOp{}^{-\frac{1}{2}}}{2}$
      scales as $\sqrt{\Dim}$. After rescaling by the
      factor~$1/\sqrt{\Dim}$ in the
      definition~\eqref{eq:MyCurveTwo_lin}, we see that the parameter
      $\MyCurveTwo{\ho}{\state}$ becomes dimension-free.
  \end{itemize}


\subsection{Proof of Proposition~\ref{PropCurve}}
\label{SecProofPropCurve}

This section is devoted to the proof of~\Cref{PropCurve}, with two
subsections corresponding to each of the claims.  Our proof relies on
an auxiliary result derived from the smoothness
conditions~\eqref{eq:curv_smooth1} and~\eqref{eq:curv_smooth2}, and
exploiting the twice differentiability of the constraint
functions~$\constrain$.
\begin{lemma}
\label{lemma:curv}
\begin{subequations}
The perturbation terms $\bweight - \bweightstar $, $\Feature -
\Featurestar$ and $\blambda - \blambdastar$ satisfy
\begin{align}
  \bweight - \bweightstar & \; = \; \Hess \, (\Feature - \Featurestar)
  + \nabla \constrain(\Featurestar)^{\top} (\blambda - \blambdastar) +
  \Deltafun
  \label{eq:KKT1} 
\end{align}
with $\distrnorm{\Deltafun}{\Hess^{-1}} \leq \frac{1}{4} \,
\distrnorm{\Feature - \Featurestar}{\Hess} + \Lip \,
\distrnorm{\Feature - \Featurestar}{\Hess} \, \distrnorm{\blambda -
  \blambdastar}{\WMt}$, and
\begin{align}
  \distrnorm[\big]{\projG \, \Hess^{\frac{1}{2}} (\Feature -
    \Featurestar)}{2} & \; \leq \; \frac{\Lip}{2} \,
  \distrnorm{\Feature - \Featurestar}{\Hess}^2 \, .
  \label{eq:KKT2}  
\end{align}
\end{subequations}
\end{lemma}
\noindent See~\Cref{sec:proof:lemma:curv} for the proof.


\subsubsection{Proof of the bound~\eqref{eq:curv_1_app}}
\label{sec:curv_just_0_proof}

Our proof consists of three steps: (i) We first provide an upper bound
on the difference in Lagrangian multipliers $\blambda - \blambdastar$,
and (ii) Then use this bound to control the difference in optimizers
$\Feature - \Featurestar$; (iii) Finally, we perform a norm
transformation to include the weighted norm
$\distrnorm{\cdot}{\CovOp{}}$, thereby establishing a connection with
the $\distrnorm{\cdot}{\ho}$ norm. 
  
\paragraph{Step 1 \!\!\!\!} (\emph{Bounding $\blambda -
\blambdastar$}) {\bf :}
We first derive an upper bound on the difference in Lagrangian
multipliers $\blambda - \blambdastar$.  We multiply
equation~\eqref{eq:KKT1} by $\projG \, \Hess^{-\frac{1}{2}}$ from the
left and find that
  \begin{align}
    \label{eq:KKT_diff1}
    \projG \, \Hess^{-\frac{1}{2}} (\bweight - \bweightstar) & =
    \projG \, \Hess^{\frac{1}{2}} (\Feature - \Featurestar) +
    \Hess^{-\frac{1}{2}} \nabla \constrain(\Featurestar)^{\top}
    (\blambda - \blambdastar) + \projG \, \Hess^{-\frac{1}{2}}
    \Deltafun \, .
  \end{align}
Recall that the matrix $\WMt$ is defined as $\WMt = \nabla
\constrain(\Featurestar) \, \Hess^{-1} \nabla
\constrain(\Featurestar)^{\top}$.  Therefore, we have
$\distrnorm{\blambda - \blambdastar}{\WMt}$ $=
\distrnorm[\big]{\Hess^{-\frac{1}{2}} \nabla
  \constrain(\Featurestar)^{\top} (\blambda - \blambdastar) }{2}$.  By
the triangle inequality, it follows from \eqref{eq:KKT_diff1} that
\begin{align}
\label{eq:blambda_diff0}
\distrnorm{\blambda - \blambdastar}{\WMt} \leq \distrnorm[\big]{\projG
  \, \Hess^{-\frac{1}{2}} (\bweight - \bweightstar) }{2} +
\distrnorm[\big]{\projG \, \Hess^{\frac{1}{2}} (\Feature -
  \Featurestar)}{2} + \distrnorm[\big]{\projG \, \Hess^{-\frac{1}{2}}
  \Deltafun}{2} \, .
\end{align}
On the right-hand side of this inequality, the first term
$\distrnorm[\big]{\projG \, \Hess^{-\frac{1}{2}} (\bweight -
  \bweightstar) }{2}$ is the one we consider dominant, as suggested
by~\Cref{lemma:curv}.  Intuitively, we expect to show that
$\distrnorm{\blambda - \blambdastar}{\WMt} \leq \PlainCon \,
\distrnorm[\big]{\projG \, \Hess^{-\frac{1}{2}}(\bweight -
  \bweightstar) }{2} \leq \frac{\PlainCon'}{\Lip}$.  In the following,
we present a rigorous argument to validate this intuition.
  
We use the inequality $\distrnorm[\big]{\projG \, \Hess^{-\frac{1}{2}}
  \Deltafun}{2} \leq \distrnorm{\Deltafun}{\Hess^{-1}}$ and invoke the
upper bounds for $\distrnorm{\Deltafun}{\Hess^{-1}}$ and
$\distrnorm[\big]{\projG \, \Hess^{\frac{1}{2}} (\Feature -
  \Featurestar)}{2}$ in~\Cref{lemma:curv}.  By further leveraging the
bound~$\distrnorm{\Feature - \Featurestar}{\Hess} \leq \frac{1}{3
  \Lip}$, we can deduce from inequality~\eqref{eq:blambda_diff0} that
  \begin{align*}
    \distrnorm{\blambda - \blambdastar}{\WMt} \leq
    \distrnorm[\big]{\projG \, \Hess^{-\frac{1}{2}} (\bweight -
      \bweightstar) }{2} + \frac{5}{12} \, \distrnorm{\Feature -
      \Featurestar}{\Hess} + \frac{1}{3} \, \distrnorm{\blambda -
      \blambdastar}{\WMt} \, .
  \end{align*}
We solve this inequality and apply the bound $\distrnorm[\big]{\projG
  \, \Hess^{-\frac{1}{2}} (\bweight - \bweightstar)}{2} \leq
\frac{1}{12 \Lip}$.  It follows that
\begin{align}
  \label{eq:blambda_diff}
  \distrnorm{\blambda - \blambdastar}{\WMt} \leq \frac{1}{3 \Lip} \, .
  \end{align}

\paragraph{Step 2 \!\!\!\!} (\emph{Bounding $\Feature - \Featurestar$}) {\bf :}
We now turn to bound the difference in optimizers $\Feature -
\Featurestar$ using the difference in vectors $\bweight -
\bweightstar$.  Let us multiply equation~\eqref{eq:KKT1} by
$\Hess^{-\frac{1}{2}}$ from the left, which yields
\begin{align}
  \label{eq:KKT_diff2}
  \Hess^{-\frac{1}{2}} (\bweight - \bweightstar) & =
  \Hess^{\frac{1}{2}} \, (\Feature - \Featurestar) +
  \Hess^{-\frac{1}{2}} \nabla \constrain(\Featurestar)^{\top}
  (\blambda - \blambdastar) + \Hess^{-\frac{1}{2}} \Deltafun \, .
\end{align}
Subtracting equations \eqref{eq:KKT_diff1}~and~\eqref{eq:KKT_diff2}
yields
\begin{align}
  \label{eq:KKT_diff}
  \Hess^{\frac{1}{2}} (\Feature - \Featurestar) = (\IdMt - \projG) \,
  \Hess^{-\frac{1}{2}} (\bweight - \bweightstar) + \projG \,
  \Hess^{\frac{1}{2}} (\Feature - \Featurestar) - (\IdMt - \projG) \,
  \Hess^{-\frac{1}{2}} \Deltafun \, .
  \end{align}
In a manner similar to Step~1, we employ the triangle inequality and
the bounds for $\distrnorm{\Deltafun}{\Hess^{-1}}$ and
\mbox{$\distrnorm[\big]{\projG \, \Hess^{\frac{1}{2}} (\Feature -
    \Featurestar)}{2}$} in \Cref{lemma:curv}.  This enables us to show
that, under the condition \mbox{$\distrnorm{\Feature -
    \Featurestar}{\Hess} \leq \frac{1}{3 \Lip}$}, the following holds:
\begin{align*}
  \distrnorm{\Feature - \Featurestar}{\Hess} \leq
  \distrnorm[\big]{(\IdMt - \projG) \, \Hess^{-\frac{1}{2}} (\bweight
    - \bweightstar)}{2} + \frac{5}{12} \, \distrnorm{\Feature -
    \Featurestar}{\Hess} + \Lip \, \distrnorm{\Feature -
    \Featurestar}{\Hess} \, \distrnorm{\blambda - \blambdastar}{\WMt}
  \, .
\end{align*}
We now substitute the term $\distrnorm{\blambda - \blambdastar}{\WMt}$
by its upper bound $\frac{1}{3 \Lip}$ in
inequality~\eqref{eq:blambda_diff} and solve the inequality.  It
follows that
\begin{align}
  \label{eq:feature_diff}
  \distrnorm{\Feature - \Featurestar}{\Hess} \leq 4 \,
  \distrnorm[\big]{(\IdMt - \projG) \, \Hess^{-\frac{1}{2}} (\bweight
    - \bweightstar)} {2} \, .
\end{align}
  
\paragraph{Step 3 \!\!\!\!} (\emph{Norm transformation}) {\bf :}
Finally, we perform a change of norm to transform
inequality~\eqref{eq:feature_diff} into the format of
bound~\eqref{eq:curv_1_app}.  We first decompose the deviation
$\Feature - \Featurestar$ into two components: one along the linear
space $\subspace$ and the other within the tangent space. It follows
from the triangle inequality and the Cauchy--Schwarz inequality that
\begin{align*}
  \distrnorm{\Feature - \Featurestar}{\CovOp{}^{-1}} & \leq
  \distrnorm[\big]{\CovOp{}^{-\frac{1}{2}} \Hess^{-\frac{1}{2}} (\IdMt
    - \projG) \, \Hess^{\frac{1}{2}} (\Feature - \Featurestar)}{2} +
  \distrnorm[\big]{\CovOp{}^{-\frac{1}{2}} \Hess^{-\frac{1}{2}} \,
    \projG \, \Hess^{\frac{1}{2}} (\Feature - \Featurestar)}{2} \\ &
  \leq \distrnorm[\big]{(\IdMt - \projG) \, \Hess^{-\frac{1}{2}}
    \CovOp{}^{-\frac{1}{2}}}{2} \, \distrnorm{\Feature -
    \Featurestar}{\Hess} + \distrnorm[\big]{\Hess^{-\frac{1}{2}}
    \CovOp{}^{-\frac{1}{2}}}{2} \, \distrnorm[\big]{\projG \,
    \Hess^{\frac{1}{2}} (\Feature - \Featurestar)}{2} \; .
  \end{align*}
As suggested by inequality~\eqref{eq:KKT2} in~\Cref{lemma:curv}, the
second term on the right-hand side is ``high-order'' and negligible.
Specifically, under the condition $\distrnorm{\Feature -
  \Featurestar}{\Hess} \leq \frac{\distrnorm{(\IdMt - \projG) \,
    \Hess^{-\frac{1}{2}} \CovOp{}^{-\frac{1}{2}}}{2}} {3 \Lip \,
  \distrnorm{\Hess^{-\frac{1}{2}} \CovOp{}^{-\frac{1}{2}}}{2}}$, we
have
\begin{align*}
\distrnorm[\big]{\Hess^{-\frac{1}{2}} \CovOp{}^{-\frac{1}{2}}}{2} \,
\distrnorm[\big]{\projG \, \Hess^{\frac{1}{2}} (\Feature -
  \Featurestar)}{2}
\; \leq \; \frac{1}{6} \, \distrnorm[\big]{(\IdMt - \projG) \,
  \Hess^{-\frac{1}{2}} \CovOp{}^{-\frac{1}{2}}}{2} \,
\distrnorm{\Feature - \Featurestar}{\Hess} \, ,
\end{align*}
which further implies that
\begin{align*}
\distrnorm{\Feature - \Featurestar}{\CovOp{}^{-1}} & \leq \,
\frac{7}{6} \, \distrnorm[\big]{(\IdMt - \projG) \,
  \Hess^{-\frac{1}{2}} \CovOp{}^{-\frac{1}{2}}}{2} \,
\distrnorm{\Feature - \Featurestar}{\Hess} \, .
\end{align*}
  
We proceed by substituting $\distrnorm{\Feature -
  \Featurestar}{\Hess}$ with its bound from
inequality~\eqref{eq:feature_diff}, which yields
\begin{align*}
\distrnorm{\Feature - \Featurestar}{\CovOp{}^{-1}} & \leq \, 5 \,
\distrnorm[\big]{(\IdMt - \projG) \, \Hess^{-\frac{1}{2}}
  \CovOp{}^{-\frac{1}{2}}}{2} \, \distrnorm[\big]{(\IdMt - \projG) \,
  \Hess^{-\frac{1}{2}} (\bweight - \bweightstar)}{2} \, .
\end{align*}
Applying the Cauchy--Schwarz inequality yields
\begin{align*}
\distrnorm{\Feature - \Featurestar}{\CovOp{}^{-1}} & \leq \, 5 \,
\distrnorm[\big]{(\IdMt - \projG) \, \Hess^{-\frac{1}{2}}
  \CovOp{}^{-\frac{1}{2}}}{2}^2 \cdot \distrnorm{\bweight -
  \bweightstar}{\CovOp{}} \, ,
\end{align*}
which establishes the bound~\eqref{eq:curv_1_app}, as stated
in~\Cref{PropCurve}.


\subsubsection{Proof of the 
bound~\eqref{eq:curv_2_app} \yaqidone}
\label{sec:curv_just_1_proof}

We observe that the vector $\Featurestar$ maximizes the linear
function $\bx \mapsto \inprod{\bx}{\bweightstar}$ over the constraint
set~$\Featureset$, whence $\inprod{\Feature}{\bweightstar} \leq
\inprod{\Featurestar}{\bweightstar}$, or equivalently
\begin{align}
  \label{eq:lin_fun_diff0}
  \inprod{\bweight}{\Feature - \Featurestar} \leq \inprod{\bweight -
    \bweightstar}{\Feature - \Featurestar}\,.
\end{align}
Our next step is to upper the right-hand side.
  
Multiplying inequality~\eqref{eq:KKT_diff} by $(\bweight -
\bweightstar)^{\top} \Hess^{-\frac{1}{2}}$ on the left yields
that
\begin{align*}
(\bweight - \bweightstar)^{\top} (\Feature - \Featurestar) & =
  (\bweight - \bweightstar)^{\top} \Hess^{-\frac{1}{2}} (\IdMt -
  \projG) \, \Hess^{-\frac{1}{2}} (\bweight - \bweightstar) \\ & +
  (\bweight - \bweightstar)^{\top} \Hess^{-\frac{1}{2}} \projG \,
  \Hess^{\frac{1}{2}} (\Feature - \Featurestar) - (\bweight -
  \bweightstar)^{\top} \Hess^{-\frac{1}{2}}(\IdMt - \projG) \,
  \Hess^{-\frac{1}{2}} \Deltafun \, .
  \end{align*}
It follows that
\begin{multline}
\label{eq:lin_fun_diff1}
(\bweight \! - \! \bweightstar)^{\top} (\Feature \! - \!
\Featurestar) \leq \distrnorm[\big]{(\IdMt - \projG) \,
  \Hess^{-\frac{1}{2}} (\bweight - \bweightstar)}{2}^2 +
\underbrace{\distrnorm[\big]{\projG \, \Hess^{-\frac{1}{2}} (\bweight
    - \bweightstar)}{2} \, \distrnorm[\big]{\projG \,
    \Hess^{\frac{1}{2}} (\Feature - \Featurestar)}{2}}_{\Term_{3}} \\ +
\underbrace{\distrnorm[\big]{(\IdMt - \projG) \, \Hess^{-\frac{1}{2}}
    (\bweight - \bweightstar)}{\Term_{4}} \,
  \distrnorm{\Deltafun}{\Hess^{-1}}}_{\Term_{4}} \, .
  \end{multline}
On the right-hand side of inequality~\eqref{eq:lin_fun_diff1}, the
terms $\Term_{3}$ and $\Term_{4}$ that involve the factor
$\distrnorm{\Deltafun}{\Hess^{-1}}$ or~\mbox{$\distrnorm[\big]{\projG
    \, \Hess^{\frac{1}{2}} (\Feature - \Featurestar)}{2}$} are
considered ``high-order'' according to \Cref{lemma:curv}.  We upper
bound term~$\Term_{3}$ using inequalities~\eqref{eq:KKT2},
\eqref{eq:feature_diff} and the
condition~\mbox{$\distrnorm[\big]{\projG \, \Hess^{-\frac{1}{2}}
    (\bweight - \bweightstar)}{2} \leq \frac{1}{12 \Lip}$}.
Furthermore, we control term~$\Term_{4}$ using the bound on
$\distrnorm{\Deltafun}{\Hess^{-1}}$ in \Cref{lemma:curv}, along with
inequalities~\eqref{eq:blambda_diff} and~\eqref{eq:feature_diff}.
This leads to the following result
\begin{align}
 (\bweight - \bweightstar)^{\top} (\Feature - \Featurestar) & \; \leq
\; 4 \, \distrnorm[\big]{(\IdMt - \projG) \, \Hess^{-\frac{1}{2}}
  (\bweight - \bweightstar)}{2}^2 \notag \\
\label{eq:lin_fun_diff2}  
& \leq \; 4 \, \distrnorm[\big]{(\IdMt - \projG) \,
  \Hess^{-\frac{1}{2}} \CovOp{}^{-\frac{1}{2}}}{2}^2 \cdot
\distrnorm{\bweight - \bweightstar}{\CovOp{}}^2.
\end{align}
Combining
inequalities~\eqref{eq:lin_fun_diff0}~and~\eqref{eq:lin_fun_diff2}
yields inequality~\eqref{eq:curv_2_app} as claimed.


\subsubsection{Proof of Lemma~\ref{lemma:curv}
\yaqidone} \label{sec:proof:lemma:curv}

\paragraph{Proof of equation~\eqref{eq:KKT1}:}
From the KKT conditions of the optimization
problem~\eqref{eq:def_lin_opt}, there are Lagrange multipliers
$\blambda$ and $\blambdastar \in \Real^{\numcon}$ such that
\begin{align*}
  \bweight = \nabla \constrain (\Feature)^{\top} \blambda \qquad
  \mbox{and} \qquad \bweightstar = \nabla \constrain
  (\Featurestar)^{\top} \blambdastar \, .
\end{align*}
Subtracting these two equations yields
\begin{align*}
  \bweight = \bweightstar + \Hess \, (\Feature - \Featurestar) +
  \nabla \constrain(\Featurestar) \, (\blambda - \blambdastar) +
  \Deltafun \, ,
\end{align*}
where the vector $\Deltafun \in \Real^{\Dim}$ is defined as
\begin{align}
  \label{eq:def_Deltafun0}
  \Deltafun \defn \big\{ \nabla \constrain(\Feature)^{\top}
  \blambdastar - \nabla \constrain(\Featurestar)^{\top} \blambdastar -
  \Hess \, (\Feature - \Featurestar) \big\} + \big\{ \nabla
  \constrain(\Feature) - \nabla \constrain(\Featurestar) \big\}^{\top}
  (\blambda - \blambdastar) \, .
\end{align}
We control the two terms on the right-hand side of
\eqref{eq:def_Deltafun0} separately.

By using the smoothness condition~\eqref{eq:curv_smooth1}, we derive
that
\begin{align*}
  \distrnorm[\Big]{ \nabla \constrain (\Feature)^{\top} \blambdastar -
    \nabla \constrain(\Featurestar)^{\top} \blambdastar - \Hess \,
    (\Feature - \Featurestar) }{\Hess^{-1}} \leq \, \frac{1}{4} \,
  \distrnorm{\Feature - \Featurestar}{\Hess} \, .
\end{align*}
Moreover, the smoothness condition~\eqref{eq:curv_smooth2} implies
that the second term satisfies
\begin{align*}
  \distrnorm[\Big]{\big\{ \nabla \constrain(\Feature) - \nabla
    \constrain(\Featurestar) \big\}^{\top} (\blambda -
    \blambdastar)}{\Hess^{-1}} \leq \; \Lip \cdot \distrnorm{\Feature
    - \Featurestar}{\Hess} \cdot \distrnorm{\blambda -
    \blambdastar}{\WMt} \, .
\end{align*}
By combining the components using the triangle inequality, we arrive
at an upper bound for the norm~$\distrnorm{\Deltafun}{\Hess^{-1}}$ as
stated in \Cref{lemma:curv}.

\paragraph{Proof of equation~\eqref{eq:KKT2}:}
Recall that the linear space $\subspace$ is defined as the span of the
rows of matrix $\nabla \constrain (\Featurestar) \,
\Hess^{-\frac{1}{2}}$\,. Therefore, there exists a vector $\by \in
\Real^{\numcon}$ such that
\begin{align}
  \label{eq:def_by}
  \frac{\projG \, \Hess^{\frac{1}{2}} (\Feature -
    \Featurestar)}{\distrnorm[\big]{\projG \,
      \Hess^{\frac{1}{2}}(\Feature - \Featurestar)}{2}} =
  \Hess^{-\frac{1}{2}} \, \nabla \constrain(\Featurestar)^{\top} \by
  \, .
\end{align}
Given the vector $\by$, we define a function $\constraintil(\cdot)
\defn \constrain(\cdot)^{\top} \by : \Real^{\Dim} \rightarrow \Real$.
The function $\constraintil$ exhibits some desired properties:
\begin{itemize}
  \item By definition of the matrix $\WMt = \nabla
    \constrain(\Featurestar) \, \Hess^{-1} \nabla
    \constrain(\Featurestar)^{\top}$, we have
  \begin{subequations}
  \begin{align}
    \label{eq:by_1}
    \distrnorm{\by}{\WMt} = \distrnorm[\big]{\Hess^{-\frac{1}{2}} \,
      \nabla \constrain(\Featurestar)^{\top} \by}{2} = 1\,.
  \end{align}
  \item Multiplying equation~\eqref{eq:def_by} by $(\Feature -
    \Featurestar)^{\top} \Hess^{\frac{1}{2}}$ from the left, we find
    that
  \begin{align}
    \label{eq:by_2}
    \distrnorm[\big]{\projG \, \Hess^{\frac{1}{2}} (\Feature -
      \Featurestar)}{2} \; = \; (\Feature - \Featurestar)^{\top} \,
    \nabla \constrain(\Featurestar)^{\top} \by = \; \nabla
    \constraintil(\Featurestar)^{\top} (\Feature - \Featurestar) \; .
  \end{align}
  \end{subequations}
\end{itemize}

Furthermore, the smoothness condition~\eqref{eq:curv_smooth2}
guarantees the Lipschitz continuity of the gradient of function
$\constraintil$. Specifically, we have:
\begin{align*}
  \distrnorm[\big]{\nabla \constraintil(\bx) - \nabla
    \constraintil(\Featurestar)}{\Hess^{-1}} & = \;
  \distrnorm[\Big]{\Hess^{-\frac{1}{2}} \big\{ \nabla \constrain(\bx)
    - \nabla \constrain(\Featurestar) \big\}^{\top} \by}{2} \\ & \,
  \leq \; \Lip \cdot \distrnorm{\bx - \Featurestar}{\Hess} \,
  \distrnorm{\by}{\WMt} \, \stackrel{\eqref{eq:by_1}}{=} \, \Lip \cdot
  \distrnorm{\bx - \Featurestar}{\Hess} \, .
\end{align*}
The property of gradient Lipschitz (see e.g. Lemma~1.2.3 in
textbook~\cite{nesterov2003introductory}) implies that
\begin{align}
  \label{eq:constraintil_Lip}
  \abs[\Big]{\, \constraintil(\Feature) - \constraintil(\Featurestar)
    - \nabla \constraintil(\Featurestar)^{\top} (\Feature -
    \Featurestar) \,} \; \leq \; \frac{\Lip}{2} \, \distrnorm{\Feature
    - \Featurestar}{\Hess}^2 \, .
\end{align}
Applying the equality relation~\eqref{eq:by_2} along with
$\constraintil(\Feature) = \constraintil(\Featurestar) = 0$, we see
that the claimed inequality~\eqref{eq:KKT2} follows from
inequality~\eqref{eq:constraintil_Lip}.



\section{Details of the mountain car experiment \yaqidone}
\label{sec:mountain_car}

In this experiment, a car is situated in a valley between two
hills. The car's objective is to overcome the gravitational pull and
reach the top of the right hill by efficiently controlling its
acceleration.

\subsection{Structure of the Markov decision process}

The Markov decision process underlying the mountain car problem has a
state space $\StateSp \subset \Real^2$ and an action space $\ActionSp
\subset \Real$.  The state $\state = (\pos, \vel)$ consists of the
current position $\pos$ and velocity $\vel$, whereas the scalar action
$\action = \for$ corresponds to the applied input force. The state
variables $(\pos, \vel)$ and action $\for$ are restricted as
\begin{align*}
  \pos \in [\posmin, \posmax] = [-1.2, 0.6], \quad \vel \in [\velmin,
    \velmax] = [-0.07, 0.07] \quad \text{and} \quad \for \in [\formin,
    \formax] = [-1, 1] \, .
\end{align*}
The mountain is described by the function
\begin{align*}
  \mou(\pos) = \tfrac{1}{3} \sin(3\pos) + \frac{0.025}{(\posmax -
    \pos)(\pos - \posmin)},
\end{align*}
over the interval $\pos \in [\posmin, \posmax]$.

Let $\moudiff$ be the derivative of the mountain shape function
$\mou$, which represents the instantaneous slope, and let $(\noisevel,
\noisepos) = (0.01, 0.0025)$ be a pair of standard deviations that
dictate the amount of randomness in the updates.  For an interval
$[a,b]$, we define the truncation function
\begin{align*}
  \truncate_{[a,b]}(u) & \defn \begin{cases} u & \mbox{if $u \in
      [a,b]$} , \\ b & \mbox{if $u > b$} , \\ a & \mbox{if $u < a$}
    .\\
  \end{cases}
\end{align*}
With this notation, at each discrete time step $\ho = 0, 1, 2,
\ldots$, the position and velocity of the car evolve as
\begin{align*}
  \vel_{\ho+1} & = \truncate_{[\velmin, \velmax]} \Big( \, \vel_{\ho}
  + 0.0015 \, \for_{\ho} - 0.0025 \, \moudiff(\pos_{\ho}) + \noisevel
  Z_\ho \Big) \\
  \pos_{\ho+1} & = \truncate_{[\posmin, \posmax]} \Big( \, \pos_{\ho}
  + \vel_{\ho+1} + \noisepos Z'_\ho \Big)
\end{align*}
where $(Z_\ho, Z'_\ho)$ are a pair of independent standard normal
variables.  Note that the system dynamics are non-linear due to both
the presence of the derivative $\moudiff$ and the truncation function
$\truncate$.

The objective of the car is to reach the peak of the mountain,
designated by the position $\posgoal = 0.45$. The reward at
state-action pair $(\state, \action)$ is given by
\begin{align*}
  \reward(\state, \action) \defn - \tfrac{1}{10} \for^2 + 100 \big [
    \max \{ 0, \, \pos - \posgoal \} \big]^2.
\end{align*}
For any policy $\policyh{}$, we define the $\discount$-discounted
value function
\begin{align*}
  \valuescalar(\policyh{}) \defn \Exp_{\policyh{}} \Big[
    \sum_{\ho=0}^{\infty} \, \discount^{\ho} \, \reward(\Stateh{\ho},
    \Actionh{\ho}) \Big],
\end{align*}
using $\discount = 0.97$.  The initial state $\stateh{0} = (\pos_0,
\vel_0)$ is generated with $\pos_0$ following a uniform distribution
over the interval $[-0.6,-0.4]$, and we initialize with velocity
$\vel_0 = 0$.


\subsection{Fitted Q-iteration (FQI) with
linear function approximation}

Here we describe the use of fitted Q-iteration (FQI) with linear
function approximation to estimate the optimal $Q$-function, along
with the corresponding greedy policy $\policyhath{}$.

\paragraph{Linear function approximation}

We approximate the the optimal $Q$-function $(\state, \action) \mapsto
\qfunstarh{}(\state, \action)$ using a $d$-dimensional linear function
class with $d = 3000$ features.  We begin by defining the \emph{base
feature maps} \mbox{$\featurepos: [\,\posmin, \posmax\,] \rightarrow
  \Real^{50}$} for position, and \mbox{$\featurevel: [\,\velmin,
    \velmax\,] \rightarrow \Real^{15}$} for velocity, with components
given by
\begin{align*}
  \begin{cases}
    \featureposj{2j+1}(\pos) \defn \cos(j\pos), \!\!  & \mbox{for
      $j=0, 1, \ldots, 24$, and} \\ \featureposj{2j}(\pos) \defn
    \sin(j\pos), & \mbox{for $j = 1, 2, \ldots, 25$} \, ;
  \end{cases}
  ~~
  \begin{cases}
    \featurevelj{2j+1}(\vel) \defn \cos(j\vel), \!\!  & \mbox{for
      $j=0, 1, \ldots, 7$, and} \\ \featurevelj{2j}(\vel) \defn
    \sin(j\vel), & \mbox{for $j = 1, 2,\ldots, 7$.}
  \end{cases}
\end{align*}
To represent the action $\action \equiv \for$, we define the
\emph{base action feature map}
\begin{align*}
  \featurefor(\for) \defn \big(1, \for, \for^2, \for^3\big) \in
  \real^4.
\end{align*}
The overall feature map $\Feature: \StateSp \times \ActionSp
\rightarrow \Real^{3000}$ is constructed by taking the outer product
of the three base feature maps $\featurepos$, $\featurevel$, and
$\featurefor$ as follows:
\begin{align}
  \Feature(\state, \action) \defn \vectorize\big\{ \featurepos(\pos)
  \otimes \featurevel(\vel) \otimes \featurefor(\for) \big\} \in
  \Real^{3000} \, .
\end{align}
Taking all possible triples of the three base features in the outer
product leads to the overall dimension $d = 3000 = 50 \times 15 \times
4$.  Given a weight vector $\bweight \in \real^{3000}$, we define the
function \mbox{$f_\bweight(\state, \action) \defn
  \inprod{\bweight}{\Feature(\state, \action)}$,} and we approximate
the optimal $Q$-function using the function class \mbox{$\RKHS \defn
  \big\{ f_\bweight \mid \bweight \in \Real^{3000} \big\}$.}


\paragraph{Fitted Q-iteration (FQI)}

We employed fitted Q-iteration with the linear feature
\mbox{$\Feature: \StateSp \!\times\! \ActionSp \!\rightarrow\!
  \Real^{3000}$} to estimate an optimal policy $\policyhath{}$.  The
FQI process begins by initializing the weight vector as $\bweight_0
\defn \zerovec \in \Real^{3000}$. In each iteration, we first use the
dataset $\Data = \big\{ (\stateh{i}, \actionh{i}, \rewardh{i},
\statenewh{i}) \big\}_{i=1}^{\numobs} \subset \StateSp \times
\ActionSp \times \Real \times \StateSp$ to construct the
pseudo-responses
\begin{align}
\label{EqnPseudoResponse}
y_i & \defn \rewardh{i} + \discount \, \max_{\action \in \ActionSp}
\underbrace{\inprod{\bweight_t}{\Feature(\statenewh{i},
    \action)}}_{f_{\bweight_t}(\statenewh{i}, \action)} \qquad
\mbox{for $i = 1, \ldots, \numobs$,}
\end{align}
corresponding to a stochastic estimate of the Bellman update applied
to our current $Q$-function estimate $f_{\bweight_t}$.  The polynomial
form of the force feature $\featurefor$ allows for a closed-form
solution to the maximum operation required in
equation~\eqref{EqnPseudoResponse}.  Given these pseudo-responses, we
then update the weight vector $\bweight_t \rightarrow \bweight_{t +
  1}$ via the ridge regression
\begin{align}
  \label{eq:exp_FQI}
  \bweight_{t+1} \defn \arg \min_{\bweight \in \real^{3000}} \Big \{
  \frac{1}{\numobs} \sum_{i=1}^{\numobs} \big\{ y_i -
  \inprod{\bweight}{\Feature(\stateh{i}, \actionh{i})} \big\}^2 +
  \ridge \norm{\bweight}_2^2 \Big \},
\end{align}
where $\ridge = \tfrac{0.01}{\numobs}$ in all experiments reported
here.

We terminate the procedure after at most $500$ iterations, or when
there have been $5$ consecutive iterations with insignificant
improvements in weights, where insignificant means that
$\norm{\bweight_{t+1} - \bweight_{t}}_2 \, / \sqrt{3000} < 0.005$.
Letting $\bweighthat$ represent the weight vector obtained from this
procedure, the resulting policy $\policyhath{}$ is given by selecting
the greedy action based on the $Q$-function estimate
\mbox{$\qfunhath{}(\state, \action) \defn
  \inprod{\bweighthat}{\Feature(\state, \action)}$.}


\subsection{Experimental configurations}

Our experiments were based on an off-line dataset consisting
of~$\numobs$ i.i.d. tuples
\begin{align*}
  \Data = \big\{ (\stateh{i}, \actionh{i}, \rewardh{i}, \statenewh{i})
  \big\}_{i=1}^{\numobs} \subset \StateSp \times \ActionSp \times
  \Real \times \StateSp,
\end{align*}
where the state-action pairs $\big\{ (\state_i, \action_i) = (\pos_i,
\vel_i, \for_i) \big\}_{i=1}^{\numobs}$ were generated from a uniform
distribution over the cube $[\posmin, \posmax] \times [\velmin,
  \velmax] \times [\formin, \formax]$.  We performed independent
experiments with the sample size $\numobs$ varying over the range
\begin{align*}
  \numobs & \in \big\{ \lfloor e^{k} \rfloor \bigm| k = 10.5, 10.75,
  11, \ldots, 13 \big\} \\ & = \{ 36315 , 46630, 59874, 76879, 98715,
  126753, 162754, 208981, 268337, 344551, 442413 \} \, .
\end{align*}
In each experiment, we generated a dataset $\Data$, estimated an
optimal policy $\policyhath{}$ based on the data, and evaluated the
return $\valuescalar(\policyhath{})$.  For each sample size, we
conducted $80$ independent trials.

In order to evaluate the return $\valuescalar(\policyhath{})$, for
each initial position $\pos_0 = -0.5 + 0.2 \, j/1000$ with $j = -500,
-499,$ $-498, \ldots, 499$, we simulated $30$ independent $1000$-step
trajectories by executing the estimated policy~$\policyhath{}$. The
average return over the $30 \times 1000$ trajectories is used as the
estimate of $\valuescalar(\policyhath{})$.

In order to approximate the policy\footnote{In general, it is not
guaranteed that $\policyrefh{}$ is equal to the optimal policy
$\policystarh{}$, due to approximation error that might arise from
using the linear function class defined here.}  $\policyrefh{}$ that
represents ``ground truth'', we conducted a single experiment with
sample size $\numobs = 6.4 \times 10^6$ to obtain $\policyrefh{}$. We
simulated $1000$ trajectories for each initial position $\pos_0$ and
calculated the average return, which serves as the reference value
$\valuescalar(\policyrefh{})$.  The value sub-optimality is then
computed as the difference \mbox{$\valuescalar(\policyrefh{}) -
  \valuescalar(\policyhath{})$.}

\section{Verification of auxiliary claims}
\label{AppAuxClaims}

In this appendix, we collect the verification of various auxiliary
claims made in the main text.


\subsection{Condition~\eqref{EqnPseudoStable} for occupation measures}
\label{AppOccCase}

In this appendix, we verify that condition~\eqref{EqnPseudoStable}
holds for the state-action occupation measures~\eqref{EqnDefnSAOcc}.
By definition, we have
    \begin{align*}
      \distrnorm{\MOpstarh{\ho} \, \funh{\,}}{\ho}^2 =
      \Exp_{\occupstar} \big[ (\MOpstarh{\ho} \,
        \funh{\,})^2(\Stateh{\ho}, \Actionh{\ho}) \big] =
      \Exp_{\occupstar} \Big[ \Exp_{\ho} \big[ \funh{}(\Stateh{\ho+1},
          \policystarh{\ho+1}(\Stateh{ \ho + 1})) \bigm| \Stateh{\ho},
          \Actionh{\ho} \big]^2 \Big] \, .
    \end{align*}
According to the property of variance, we can deduce
    \begin{align*}
      \Exp_{\occupstar} \Big[ \Exp_{\ho} \big[
          \funh{}\big(\Stateh{\ho+1}, \policystarh{\ho+1}(\Stateh{ \ho
            + 1})\big) \bigm| \Stateh{\ho}, \Actionh{\ho} \big]^2
        \Big] \leq \Exp_{\occupstar} \Big[
        \funh{}^2\big(\Stateh{\ho+1}, \policystarh{\ho+1}( \Stateh{\ho
          + 1}) \big) \Big] = \distrnorm{\funh{\,}}{\ho+1}^2 \, .
    \end{align*}
As a consequence, we find that $\distrnorm{\MOpstarh{\ho} \,
  \funh{\,}}{\ho} \leq \distrnorm{\funh{\,}}{\ho+1}$.  Applying this
inequality recursively leads to the conclusion that for any indices $1
\leq \ho \leq \honew \leq \Ho$, we have
    \begin{align*}
      \distrnorm[\big]{\MOpstarhtoh{\ho}{\honew} \, \funh{\,}}{\ho} =
      \distrnorm[\big]{\MOpstarh{\ho} \, \MOpstarhtoh{\ho+1}{\honew}
        \, \funh{\,}}{\ho} \leq
      \distrnorm[\big]{\MOpstarhtoh{\ho+1}{\honew} \,
        \funh{\,}}{\ho+1} \leq
      \distrnorm[\big]{\MOpstarhtoh{\ho+2}{\honew} \,
        \funh{\,}}{\ho+2} \leq \cdots \leq
      \distrnorm{\funh{\,}}{\honew} \, .
    \end{align*}
    This establishes the bound~\eqref{EqnPseudoStable} with
    $\Radphistar = 1$.

    
\subsection{Details of Example~\ref{ExaCurve}}
\label{AppExample}

In this appendix, we complete the argument outlined
in~\Cref{ExaCurve}.  In particular, our goal is to show that
conditions~\ref{eq:curv_1} and~\ref{eq:curv_2} hold with parameter
$\MyCurveTwo{\ho}{\state} \defn 16\sqrt{2} \; \exprad$.

We begin by connecting the Euclidean norm $\distrnorm{\cdot}{2}$ with
the $\distrnorm{\cdot}{\ho}$ norm that is defined by the occupation
measure.\footnote{In the argument given here, we consider a general
dimension $\Dim$ so as to convey the general idea, but the example
itself has $\Dim = 2$.}  Let us assume that the occupation measure
under the optimal policy is sufficiently exploratory so as to ensure
that the covariance matrix \mbox{$\CovOp{\ho} = \Exp_{\occupstar}
  \big[ \, \Feature(\Stateh{\ho}, \Actionh{\ho}) \,
    \Feature(\Stateh{\ho}, \Actionh{\ho})^{\top} \big] \in \Real^{\Dim
    \times \Dim}$} is well-conditioned in the sense that $\tfrac{1}{2
  \Dim} \IdMt \preceq \CovOp{\ho} \preceq \tfrac{2}{\Dim} \IdMt$.
This sandwich relation implies that
  \begin{align}
\label{EqnSimple}    
  \tfrac{1}{\sqrt{2\Dim}} \, \distrnorm{\bx}{2} \leq
  \distrnorm{\bx}{\CovOp{\ho}} \leq \sqrt{\tfrac{2}{\Dim}} \,
  \distrnorm{\bx}{2} \quad \mbox{and} \quad
  \distrnorm{\bx}{\CovOp{\ho}^{-1}} \leq \sqrt{2\Dim} \,
  \distrnorm{\bx}{2}.
\end{align}
For linear functions $\funh{\ho}(\state) =
\inprod{\bweight}{\Feature(\state)}$ and $\qfunstarh{\ho}(\state) =
\inprod{\bweightstar}{\Feature(\state)}$, we have $\distrnorm{\bweight
  - \bweightstar} {\CovOp{\ho}} = \distrnorm{\funh{\ho} -
  \qfunstarh{\ho}}{\ho}$ and
\mbox{$\distrnorm{\bweightstar}{\CovOp{\ho}} =
  \distrnorm{\qfunstarh{\ho}}{\ho}$}.  Using our
inequalities~\eqref{EqnSimple}, we find that
\begin{align}
\frac{\distrnorm{\bweight -
    \bweightstar}{2}}{\distrnorm{\bweightstar}{2}} \; \leq
\frac{\sqrt{2 \Dim} \distrnorm{\bweight -
    \bweightstar}{\CovOp{\ho}}}{\sqrt{\frac{\Dim}{2}}
  \distrnorm{\bweightstar}{\CovOp{\ho}}} = \frac{2 \,
  \distrnorm{\funh{\ho} - \qfunstarh{\ho}}{\ho}}
     {\distrnorm{\qfunstarh{\ho}}{\ho}}.
\end{align}
Furthermore, when $\distrnorm{\bweight - \bweightstar}{2} \leq
\distrnorm{\bweightstar}{2}$, we have the bound
\begin{align*}
   \distrnorm{\bweight}{2} \leq 2 \, \distrnorm{\bweightstar}{2} \leq
   2\sqrt{2\Dim} \, \distrnorm{\bweightstar}{\CovOp{\ho}} =
   2\sqrt{2\Dim} \, \distrnorm{\qfunstarh{\ho}}{\ho}.
\end{align*}

Now recall from the main text our two inequalities
\eqref{eq:exp_curv0_1} and~\eqref{eq:exp_curv0_2}, as well as the
inequality
\begin{align*}
\angle(\bweight, \, \bweightstar) \; \leq \frac{2 \,
  \distrnorm{\bweight - \bweightstar}{2}}
      {\distrnorm{\bweightstar}{2}}.
\end{align*}
Combining these bounds with the inequalities above, we find that
\begin{align*}
\distrnorm[\big]{\Feature(\state, \policyh{\ho} (\state)) -
  \Feature(\state, \policystarh {\ho}(\state))}{\CovOp{\ho}^{-1}} & \;
\leq \; 4 \sqrt{2 \Dim} \, \exprad \, \cdot \,
\frac{\distrnorm{\funh{\ho} - \qfunstarh{\ho}}{\ho}}
     {\distrnorm{\qfunstarh{\ho}}{\ho}} \; , \\
\abs[\big]{\funh{\ho}(\state, \policyh {\ho}(\state)) -
  \funh{\ho}(\state, \policystarh{\ho}(\state))} & \; \leq \; 16
\sqrt{2 \Dim} \, \exprad \, \distrnorm{\qfunstarh{\ho}}{\ho} \;
\bigg\{ \frac{\distrnorm{\funh{\ho} - \qfunstarh{\ho}}{\ho}}
        {\distrnorm{\qfunstarh{\ho}}{\ho}} \bigg\}^2 \, .
\end{align*}
Consequently, we have established the claim---namely, that
conditions~\ref{eq:curv_1} and~\ref{eq:curv_2} hold with parameter
$\MyCurveTwo{\ho}{\state} \defn 16\sqrt{2} \; \exprad$.


\subsection{Proof of the telescope inequality~\eqref{eq:subopt_ub_old}}
\label{proof:subopt_ub_old}

For completeness of this paper,\footnote{We are not claiming novelty
here; see Theorem 2 of the paper~\cite{xie2020q}; or Lemma 3.2 in the
paper~\cite{duan2021risk} for analogous results.} let us prove the
telescope relation~\eqref{eq:subopt_ub_old} stated
in~\Cref{SecFastIntuition}.
For any policy $\policy = (\policyh{1},
\ldots, \policyh{\Ho})$ and sequence of functions \mbox{$\fun =
  (\funh{1}, \ldots, \funh{\Ho})$} with $\funh{\Ho} = \rewardh{\Ho}$,
we have the ``telescope'' relation
\begin{align}
 \label{eq:telescope}
 \Valuepolicyh{1}{\policy}(\state) =
 \funh{1}(\state,\policyh{1}(\state)) + \myssum{\ho=1}{\Ho-1}
 \,\Exp_{\policy} \big[ \big(\BellOph{\ho}{\policy} \funh {\ho+1} -
   \funh{\ho}\big)(\Stateh{\ho}, \Actionh{\ho}) \bigm| \Stateh{1} =
   \state \, \big] \quad \text{for any state $\state \in \StateSp$.}
\end{align}
Here the value function $\Valuepolicyh{1}{\policy}$ is given by
$\Valuepolicyh{1}{\policy}(\state) \defn \qfunh{1}^{\policy}(\state,
\policyh{1}(\state))$ for the $Q$-function $\qfunh{1}^{\policy}$
defined in equation \eqref{EqnDefnQfunPol}.  Taking $\fun = \qfunhat$
in equation~\eqref{eq:telescope} yields
\begin{subequations}
  \begin{align}
    \label{eq:telescopestar}
    \Valuepolicyh{1}{\policy}(\state) & = \qfunhath{1} ( \state,
    \policyh{1}(\state) ) + \myssum{\ho=1}{\Ho-1} \, \Exp_{\policy}
    \big[ \big( \BellOph{\ho}{\policy} \qfunhath{\ho+1} -
      \qfunhath{\ho} \big)(\Stateh{\ho}, \Actionh{\ho}) \bigm|
      \Stateh{1} = \state \, \big] \, .
  \end{align}
Letting $\policy = \policyhat$ in equation~\eqref{eq:telescopestar}
yields
\begin{align}
\label{eq:telescopehat}  
\Valuepolicyh{1}{\policyhat}(\state) & = \qfunhath{1} (\state,
\policyhath{1}(\state)) + \myssum{\ho=1}{\Ho-1} \, \Exp_{\policyhat}
\big[ \big( \BellOph{\ho}{\policyhat} \qfunhath{\ho+1} -\qfunhath{\ho}
  \big)(\Stateh{\ho}, \Actionh{\ho}) \bigm| \Stateh{1} = \state \,
  \big] \, .
\end{align}
\end{subequations}
Since $\policyhat{}$ is a greedy policy with respect to function
$\qfunhat$, we have
\begin{align*}
\qfunhath{1}(\state, \policyhath{1}(\state)) \geq \qfunhath{1}(\state,
\policyh{1}(\state)), \quad \mbox{and} \quad
\mbox{$\BellOph{\ho}{\policyhat} \qfunhath{\ho+1} = \BellOpstarh{\ho}
  \! \qfunhath{\ho+1} \geq \BellOph{\ho}{\policy} \qfunhath{\ho+1}
  \quad$ for any policy~$\policy$.}
\end{align*}
Using this fact and subtracting equations~\eqref{eq:telescopestar}
and~\eqref{eq:telescopehat}, we obtain
\begin{align*}
\Valuepolicyh{1}{\policy}(\state) - \Valuepolicyh{1}{\policyhat}
(\state) \leq \sum_{\ho=1}^{\Ho-1} \, \big( \Exp_{\policy} -
\Exp_{\policyhat} \big) \big[ \big( \BellOpstarh{\ho} \qfunhath{\ho+1}
  - \qfunhath{\ho} \big)(\Stateh{\ho}, \Actionh{\ho}) \bigm|
  \Stateh{1} = \state \, \big] \, .
\end{align*}
Finally, taking the expectation over the initial distribution
$\sdistrinit$ yields the claimed inequality~\eqref{eq:subopt_ub_old}.







\end{document}